\pdfoutput=1
\documentclass{article}

\usepackage{arxiv}
\usepackage[T1]{fontenc}
\usepackage[numbers,sort&compress]{natbib}
\usepackage{amsmath,amssymb,bm}
\usepackage{booktabs}
\usepackage{longtable}
\usepackage{multirow}
\usepackage{multicol}
\usepackage[dvipsnames]{xcolor}
\usepackage{siunitx}
\usepackage{microtype}
\usepackage{placeins}
\usepackage{capt-of}
\usepackage[inline]{enumitem}
\usepackage{graphicx}
\usepackage{threeparttable}
\usepackage{subcaption}
\usepackage{authblk}
\usepackage{hyperref}
\usepackage{doi}
\DeclareGraphicsExtensions{.pdf,.png,.jpg,.jpeg}

\hypersetup{
  colorlinks=true,
  linkcolor=blue!60!black,
  citecolor=blue!60!black,
  urlcolor=blue!60!black,
  pdftitle={Physics-Informed Identification of Tensegrity Prestress across Taut–Slack Transitions},
  pdfauthor={Sandesh Lamsal, Sumit Kumar, Rubi Bhandari, Subhamoy Sen, Andrzej Rutkiewicz, Leszek Malyszko, Landolf Rhode-Barbarigos}
}

\renewcommand{\headeright}{}
\renewcommand{\undertitle}{}
\renewcommand{\shorttitle}{}

\title{Physics-Informed Identification of Tensegrity Prestress across Taut--Slack Transitions}

\author[1,*]{Sandesh~Lamsal}
\author[1,*]{Sumit~Kumar}
\author[2]{Rubi~Bhandari}
\author[3]{Subhamoy~Sen}
\author[4]{Andrzej~Rutkiewicz}
\author[5]{Leszek~Ma{\l}yszko}
\author[1]{Landolf~Rhode-Barbarigos}
\affil[1]{Department of Civil and Architectural Engineering, University of Miami, Coral Gables, FL 33146, USA}
\affil[2]{Department of Civil and Environmental Engineering, Florida International University, Miami, FL 33174, USA}
\affil[3]{i4S Laboratory, Indian Institute of Technology Mandi, Mandi, Himachal Pradesh, India}
\affil[4]{Faculty of Geoengineering, University of Warmia and Mazury in Olsztyn, Prawoche\'nskiego 15, 10-720 Olsztyn, Poland}
\affil[5]{Institute of Technology, University of Applied Sciences in Elbl\k{a}g, Al. Grunwaldzka 137, 82-300 Elbl\k{a}g, Poland}
\affil[*]{These authors contributed equally to this work.}
\affil[ ]{Corresponding author: Sandesh Lamsal, \texttt{sandeshlamsal@miami.edu}}

\date{}

\begin{document}
\maketitle

\begin{abstract}
Prestress governs the stability, stiffness, and dynamic behavior of tensegrity
structures, but it is not directly observable from sparse structural
measurements, and it may decrease during service life. Under strong
excitation, the cables may slacken and the dynamics become nonsmooth. Existing
identification methods are limited to the taut regime, and physics-informed
networks for tensegrities solve only the forward problem of predicting the
response from known member parameters.

This study formulates the inverse problem. Neural networks are trained on the
equation-of-motion residual in constrained Cartesian coordinates, where the
mass matrix is constant and the force densities enter linearly. The force
densities, damping, and axial stiffness are trainable parameters, and the
tension-only cable law is kept. A Fisher-information analysis shows, before
inversion, which member parameters a record can determine.

The method is tested on simulated planar tensegrity frames with up to five modules and a three-dimensional prism, and on measured records
of a biomimetic leg and a full-scale tensegrity simplex. On the one-module
frame, the identifiable force densities are recovered within $1.5$\,\% from
noisy accelerations at half the free coordinates, on records with and without
taut--slack transitions. Records with transitions determine members that taut
records leave undetermined. On the five-module frame, a 16-cable two-span
bridge, the median force-density error rises moderately, to $4.2$\,\%. On
measured records, an identified spring stiffness is within $0.1$\,\% of its
independently measured value on average, and the identified simplex prestress
at 13 levels agrees with its cable force transducers to $2.9$\,\%
root-mean-square.
\end{abstract}


\section*{Nomenclature}
\begingroup\footnotesize\linespread{0.94}\selectfont
\setlength{\columnsep}{16pt}
\newlength{\nomw}\setlength{\nomw}{5.0em}
\newcommand{\nomhead}[1]{\par\vspace{2.5pt}\noindent\textit{#1}\par\nopagebreak}
\newlength{\nomtw}
\newcommand{\nomen}[2]{\par\noindent\hangindent=\nomw\hangafter=1
  \settowidth{\nomtw}{#1}%
  \ifdim\nomtw>\nomw #1\\\hspace*{\nomw}#2\else\makebox[\nomw][l]{#1}#2\fi}
\begin{multicols}{2}\raggedright
\nomhead{Structure and dynamics}
\nomen{$q,\ \dot q,\ \ddot q$}{nodal coordinates, velocities, accelerations ($\mathbb{R}^{3n}$, $n$ nodes)}
\nomen{$n_{\mathrm{free}},\ n_{\mathrm{mem}}$}{free coordinates, members}
\nomen{$M,\ G$}{constant mass matrix, gravity vector}
\nomen{$X_k,\ Y_k$}{bar and cable member operators}
\nomen{$R(q),\ \nabla R$}{constraint functions and their Jacobian}
\nomen{$\lambda$}{constraint multipliers}
\nomen{$\kappa$}{quadratic velocity terms, bar constraints}
\nomen{$f_s,\ f_d,\ f_{\mathrm{ext}}$}{cable elastic, damping, external forces (N)}
\nomen{$p$}{forcing amplitude (N)}
\nomen{$\bm A$}{equilibrium matrix}
\nomen{$s,\ m$}{self-stress states, mechanisms}
\nomen{$T,\ V_g,\ V_s$}{kinetic, gravitational, strain energy (J)}
\nomen{$E,\ W$}{mechanical energy, work (J)}

\nomhead{Members}
\nomen{$s_k$}{member vector of cable $k$}
\nomen{$\bar L_k$}{length of bar $k$ (m)}
\nomen{$L_k^{\mathrm{ref}},\ L_{0k}$}{length of cable $k$ in the reference configuration, and its unstressed rest length (m)}
\nomen{$\sigma_k,\ K_k$}{force density (N/m) and stiffness (N/m) of cable $k$}
\nomen{$EA,\ c_k$}{cable axial stiffness (N) and damper coefficient (N\,s/m)}
\nomen{$T_k,\ g_k$}{cable tension (N) and slackness (m)}

\nomhead{Learning}
\nomen{$\tau,\ T_{\mathrm{ref}}$}{scaled time $t/T_{\mathrm{ref}}$ and reference time (s)}
\nomen{$q_\theta,\ \lambda_\phi$}{displacement network, multiplier head}
\nomen{$\mathbf{c},\ s(\tau)$}{record descriptors; hard-IC window}
\nomen{$r_{\mathrm{EOM}}$}{equation-of-motion residual}
\nomen{$r_{\mathrm{bar}}$}{bar-length constraint residual}
\nomen{$\mathcal{L}_{\mathrm{data}}$}{sensor-data loss}
\nomen{$\mathcal{L}_{\mathrm{EOM}},\ \mathcal{L}_{\mathrm{bar}}$}{physics, constraint losses}
\nomen{$w_b,\ w_d$}{loss weights}
\nomen{$\beta$}{sharpness of the smoothed cable law (--)}

\nomhead{Identifiability and assessment}
\nomen{$S(t)$}{sensitivities of the sensor accelerations}
\nomen{$J$}{Fisher information matrix}
\nomen{$s_{n,i}$}{noise standard deviation, channel $i$}
\nomen{$\bm S,\ \bm\Gamma$}{stress matrix, member grouping matrix}
\nomen{$\lambda_{\min}$}{smallest eigenvalue (stability margin)}
\nomen{$U_c$}{cable utilization (--)}
\nomen{$\hat{\boldsymbol\sigma},\ \hat c,\ \widehat{EA}$}{identified prestress, damping, stiffness}

\nomhead{Abbreviations}
\par\noindent CRLB: Cram\'er--Rao lower bound; DAE: differential-algebraic
equations; DOF: degree of freedom; EOM: equations of motion; LP: linear
program; NCP: nonlinear complementarity problem; PINN: physics-informed
neural network; RMSE: root-mean-square error.
\end{multicols}
\endgroup

\section{Introduction}
\label{sec:intro}

Tensegrities are self-stressed networks of bars in compression and cables in
tension, in which self-stress, instead of moment-resisting joints, provides the
structural integrity \citep{Snelson2012, Motro2003, Skelton2009book}. This
self-stress governs the tangent stiffness, the natural frequencies, and the
load paths, so a single geometry behaves differently at different prestress
levels \citep{Calladine1978, PellegrinoCalladine1986}. This dependence makes tensegrities tunable and collapsible, since a tensegrity collapses when its prestress is removed. High prestress levels, however, may reduce the capacity of the system to sustain external loads. Prestress cannot be inferred from the geometry or measured without embedded sensors, and it may decrease during service through creep, anchorage slip, and thermal cycling; its loss is a critical damage mode \citep{AshwearEriksson2014,
SychterzSmith2018}. Because stiffness and frequency scale with the
self-stress, this loss changes the structural behavior while the geometry
stays visibly unchanged.

Forward analysis of tensegrities is well developed. Form-finding determines the shapes that sustain a self-stress \citep{TibertPellegrino2003}, and dynamic formulations predict motion about a prestressed configuration
\citep{Murakami2001Part1, GoyalSkelton2019}. These analyses take the member parameters as input. Structural identification solves the opposite problem: it
infers the model, or the state, that produced an observed response, for model
updating, health monitoring, virtual sensing, and control \citep{Ljung1999,
FriswellMottershead1995, FarrarWorden2013}.

For a tensegrity, identification is both more critical and more difficult than for a conventional frame. It is more critical because the governing quantity, the prestress, may decrease during service and is not directly observable from
sparse structural measurements. It is more difficult for three reasons rooted
in the mechanics. The prestress is distributed: once several independent
self-stress states exist, it is no longer fixed by the geometry, and a change
in the rest length of one cable redistributes force through the whole network.
The cables are unilateral: they carry tension only and may slacken under strong excitation, which makes the dynamics nonsmooth. Finally, the instrumentation
is sparse: measuring every member force is impractical, and the few
measurements available must be combined with equilibrium, compatibility, the
constitutive law, and member-wise admissibility.

The recovery of the prestress state and the member properties of a tensegrity
from sparse, noisy dynamic records, with the tension-only cable law
$T_k \geq 0$, $g_k \geq 0$, $T_k g_k = 0$ kept in the model, has received
little attention. Here $T_k$ is the tension in cable $k$ and $g_k$ its
slackness. It is also unclear which member parameters a given record can
determine, and whether a taut--slack transition supplies information that a
taut record lacks.

Earlier identification studies address parts of this problem.
Prestress-dependent dynamic properties have been tracked on a grid
\citep{DubeAngellier2013}, on a simplex across 13 measured prestress levels
\citep{Malyszko2020Simplex}, and on a five-module active tensegrity across 11 levels \citep{BelHadjAliSmith2010IJSS}. Self-diagnosis locates unknown loads
and damage \citep{AdamSmith2007, SychterzSmith2018}, cellular decomposition
determines the minimal sensor configurations
\citep{Aloui2019SensorPlacement}, and global equilibrium recovers several
cable forces from measurements on a few members
\citep{Tian2024GlobalCableForce}. Modal updating and Bayesian filtering
track prestress-sensitive frequencies and parameter drift on linearized
models \citep{Prism2025JSV, Aswal2021, ChatziSmyth2009UKF}. These studies are limited to the linearized taut regime. They rely on modal extraction, which
loses sensitivity under sparse instrumentation and on structures whose
modes are stiffened elastically rather than geometrically
\citep{BelHadjAli2010}. They also fix the damping beforehand or absorb it into the identified parameters, although measured tensegrity dissipation
comes mainly from joint friction \citep{OppenheimWilliams2001}.

Physics-informed neural networks can avoid the linearization. A network
approximates the response while the governing equations enter the training
objective as a residual, and unknown physical parameters enter as trainable
variables, so that response reconstruction and parameter identification
become one inverse problem \citep{Lagaris1998, Raissi2019, Karniadakis2021}.
For structural systems the approach is established: it has been applied to
inversion in solid mechanics \citep{Haghighat2021SolidPINN}, to structural
parameter identification, to nonlinear and thermally damped systems, and to
health assessment \citep{Guo2023PINNStructural, LiuMeidani2023JEM,
AlAdlyKripakaran2024}. Bayesian variants quantify the uncertainty in the
recovered quantities, and recent work estimates state and parameters jointly
under incomplete measurement \citep{Yang2021BPINN, Mahar2025MSSP}. All of these studies treat lumped systems divided by algebraic block or by degree of freedom; a tensegrity requires a
division by physical member.

Physics-informed learning has been applied to tensegrities for
force-density-informed prestress design \citep{Trinh2022}, for dynamic
response prediction \citep{Xu2025NAECON}, and in geometrically nonlinear,
symmetry-reduced, and energy-informed forms \citep{WangZhuMiao2025,
QinChen2026SymPINN, Song2025NLDyn}. All of these studies solve the forward
problem on taut structures. The learning-based work closest to
identification fits the parameters of the physics engine used to simulate
tensegrity robots to their locomotion trajectories \citep{Wang2023Real2Sim,
Chen2024GNN}, without a structural interpretation of the identified
quantities or a statement of what a record can determine. Related fields
have addressed parts of the nonsmooth problem. An active elastoplastic
constraint has been used as an additional measurement equation
\citep{Chatzis2017DUKF}, and the sensitivity to an event has been included
in the Fisher information through saltation matrices, which map a small
change in the state just before an event to the change just after it
\citep{Kong2024Saltation}. Complementarity conditions have been included in
physics-informed losses for static contact \citep{Sahin2024Contact}, and
networks for nonsmooth dynamics exist for low-dimensional friction
oscillators \citep{Li2024Friction}. 

The motivation for this study follows from this review. First, the
identification methods applied to tensegrities are limited to the linearized
taut regime and do not keep the tension-only cable law when cables slacken.
Second, physics-informed methods have identified the parameters of lumped
structural models, but for tensegrities they have been used only for forward
analysis of taut structures. Third, the studies of nonsmooth identification
reviewed above do not treat cable structures, and they do not quantify how
much information a taut--slack transition contributes.

This study aims to address these points with a physics-informed network for
the inverse problem. The equations of motion are written in constrained
Cartesian coordinates \citep{GoyalSkelton2019}. In this form the mass matrix
is constant, which stabilizes training, and the cable force densities enter
the residual linearly, which gives closed-form residual sensitivities. A
displacement network and a constraint-multiplier network are trained on that
residual, with the force densities, the axial stiffness, and the damping
coefficients treated as trainable parameters.

The study makes three contributions: an inverse formulation for tensegrity
dynamics that keeps the tension-only cable law through taut--slack
transitions; an identifiability analysis that shows, before inversion, which
member parameters a record can determine and to what precision; and a
demonstration of both on benchmark structures and on measured records. In the
formulation, the tension-only condition is imposed through a smoothed
(softplus) cable law, which is compared on identical records with two
alternatives: a Fischer--Burmeister complementarity residual (a function that
is zero exactly when the tension-only conditions hold) and collocation points
concentrated near the transitions. The identifiability analysis uses Fisher
information and Cram\'er--Rao bounds, and the linear entry of the force
densities shows which members a record cannot determine. Where the parameter
set is ill-conditioned, tying members into groups can restore a well-posed
estimate, and those groups are the clustered cables already used for
actuation \citep{Friswell1998Subset, Moored2009Clustered}. The demonstration uses simulated records of planar tensegrity frames built from one, two, three, and five Snelson X-modules, the largest of which forms a 16-cable two-span bridge, and of a three-dimensional triplex prism, with gravity and in a zero-gravity prestress series. It then uses measured records from three laboratory tests: the modal frequencies of a hexagonal tensegrity prism, chirp records of a biomimetic leg whose spring rates were measured independently, and impact records of a full-scale tensegrity simplex with a force transducer in each cable, at 13 prestress levels. No
measured record in this study contains a slack cable. The treatment of
taut--slack transitions is therefore validated on simulated records only, and
the measured experiments validate the rebuilt models, the small-oscillation
identification, and the time-domain residual.

The remainder of this study is organized as follows.
Section~\ref{sec:data} presents the dynamics formulation, the benchmark
structures, and the simulated datasets. Section~\ref{sec:methodology}
develops the inverse formulation, the transition treatments, and the
identifiability analysis. Section~\ref{sec:results} reports the
demonstrations, Section~\ref{sec:discussion} discusses their implications
and limits, and Section~\ref{sec:conclusions} summarizes the findings and
the scope.

\FloatBarrier
\section{Tensegrity Dynamics and Simulated Datasets}
\label{sec:data}

\subsection{Constrained Lagrangian dynamics in Cartesian coordinates}
\label{sec:formulation}

The dynamics are written in redundant Cartesian nodal coordinates
$q \in \mathbb{R}^{3n}$, following \citet{GoyalSkelton2019} as implemented in the STEDY code \citep{Tadikonda2019STEDY}. Each
rigid bar contributes consistent translational and rotational
inertia to a mass matrix $M$ \emph{constant} in $q$. Gravity
enters as a constant generalized force $G$. Support conditions are
linear constraints $Aq = b$ selecting the fixed coordinates. Each bar
$k$ imposes the quadratic length constraint
$q^{\top}X_k^{\top}X_k\,q = \bar L_k^2$, with $\bar L_k$ the bar
length and $X_k$ the constant operator extracting the bar vector.
The position-level constraints are collected as $R(q) = 0$. The motion then satisfies the index-3 differential-algebraic system, which is integrated in its index-1 (acceleration-level) saddle-point form,
\begin{equation}
  M\ddot q \;=\; \nabla R(q)^{\top}\lambda \;+\; f_s(q) \;+\;
  f_d(q,\dot q) \;+\; G \;+\; f_{\mathrm{ext}}(t),
  \label{eq:eom}
\end{equation}
with $\lambda$ the constraint multipliers and $f_s$, $f_d$ the cable
elastic and damping forces defined below. The kinetic energy is
$T = \tfrac{1}{2}\dot q^{\top} M \dot q$. Differentiating the
position-level constraints twice gives their velocity- and
acceleration-level forms,
\begin{equation}
  \nabla R(q)\,\dot q = 0,
  \qquad
  \nabla R(q)\,\ddot q + \kappa(q,\dot q) = 0,
  \label{eq:constraint-levels}
\end{equation}
in which $\kappa$ collects the quadratic velocity terms of the bar
constraints. Combining Eq.~\eqref{eq:eom} with the acceleration-level
condition gives the saddle-point system solved at every
right-hand-side evaluation,
\begin{equation}
  \begin{bmatrix} M & -\nabla R(q)^{\top}\\[2pt]
                  -\nabla R(q) & 0 \end{bmatrix}
  \begin{bmatrix} \ddot q\\[2pt] \lambda \end{bmatrix}
  \;=\;
  \begin{bmatrix} f_s + f_d + G + f_{\mathrm{ext}}\\[2pt]
                  \kappa \end{bmatrix},
  \label{eq:saddle}
\end{equation}
which returns the accelerations and the bar constraint forces
together. An admissible initial state satisfies $R(q_0) = 0$ and
$\nabla R(q_0)\,\dot q_0 = 0$. This study uses the constant mass matrix and the linear entry of the cable force densities in $f_s$ throughout. Parameter
sensitivities of the residual are therefore closed-form
(Section~\ref{sec:identifiability}).

The simulated records are obtained by integrating Eq.~\eqref{eq:eom} through Eq.~\eqref{eq:saddle} with tight tolerances. The simulator reproduces the state derivatives of the STEDY code to a relative accuracy of $10^{-14}$, and its small-oscillation frequencies agree with those of an independently assembled prestressed truss tangent stiffness (Table~\ref{tab:verification}). Its energy and constraint checks are described in Section~\ref{sec:audits}.

\subsection{Tension-only cable law and taut--slack transitions}
\label{sec:cablelaw}

Each cable $k$ is linear elastic and works only in tension. With $s_k = Y_k q$ the member vector, $L_{0k}$ the rest length, that is, the length of the unstressed cable, and $K_k$ the stiffness, the force density is
\begin{equation}
  \sigma_k(q) \;=\;
  \begin{cases}
    K_k\left(1 - L_{0k}/\lVert s_k\rVert\right), & \lVert s_k\rVert > L_{0k},\\[2pt]
    0, & \text{otherwise},
  \end{cases}
  \label{eq:cablelaw}
\end{equation}
and the nodal cable force is $f_s = -\sum_k \sigma_k Y_k^{\top}Y_k q$.
Axial viscous dampers with coefficients $c_k$ act only while the cable
is taut. Viscous cable damping is a modeling choice made for
differentiability; measured tensegrity dissipation is dominated by
joint friction \citep{OppenheimWilliams2001}, and since the damping
coefficients are among the identified parameters, the consequence of that mismatch is shown on measured data in Section~\ref{sec:res-olsztyn} and discussed in Section~\ref{sec:discussion}. Equation~\eqref{eq:cablelaw} is the piecewise-smooth form of
the complementarity law $T_k \geq 0$, $g_k \geq 0$, $T_k g_k = 0$, in
which $T_k = \sigma_k \lVert s_k\rVert$ is the tension in cable $k$ and $g_k = L_{0k} - \lVert s_k\rVert + T_k/K_k$ is its elastic slackness.
That slackness vanishes exactly while the cable is taut and reduces
to the geometric slack $L_{0k} - \lVert s_k\rVert$ while force-free.
Classical forward solvers treat the taut--slack transitions with complementarity time-stepping \citep{OlivetoSivaselvan2011, Kan2018}.

The prestress state is described by the cable force densities $\sigma_k$ at the reference configuration, and errors in the identified prestress are quoted throughout as force-density errors. Here $\sigma_k$ denotes a force density, the tension per unit length of the cable, and not a stress; the symbol $q$, often used for force densities in form-finding, is kept for the nodal coordinates of the dynamic formulation in Section~\ref{sec:formulation}. Member properties follow the convention $K_k = EA/L_k^{\mathrm{ref}} + \sigma_k$ and $L_{0k} = L_k^{\mathrm{ref}}(1 - \sigma_k/K_k)$, in which $L_k^{\mathrm{ref}}$ is the length of cable $k$ in the prestressed reference configuration, longer than its rest length $L_{0k}$. A given $\boldsymbol\sigma$ then makes the reference configuration an exact static equilibrium. Admissible prestress states come from a linear program
over force densities that balances gravity with all
$\sigma_k \geq \sigma_{\min}$. The self-stress null-space modes of the
equilibrium operator are computed for the identifiability analysis.

\subsection{Benchmark structures and prestress series}
\label{sec:benchmarks}

\begin{figure*}[!tbp]
  \centering
  \includegraphics[width=\textwidth]{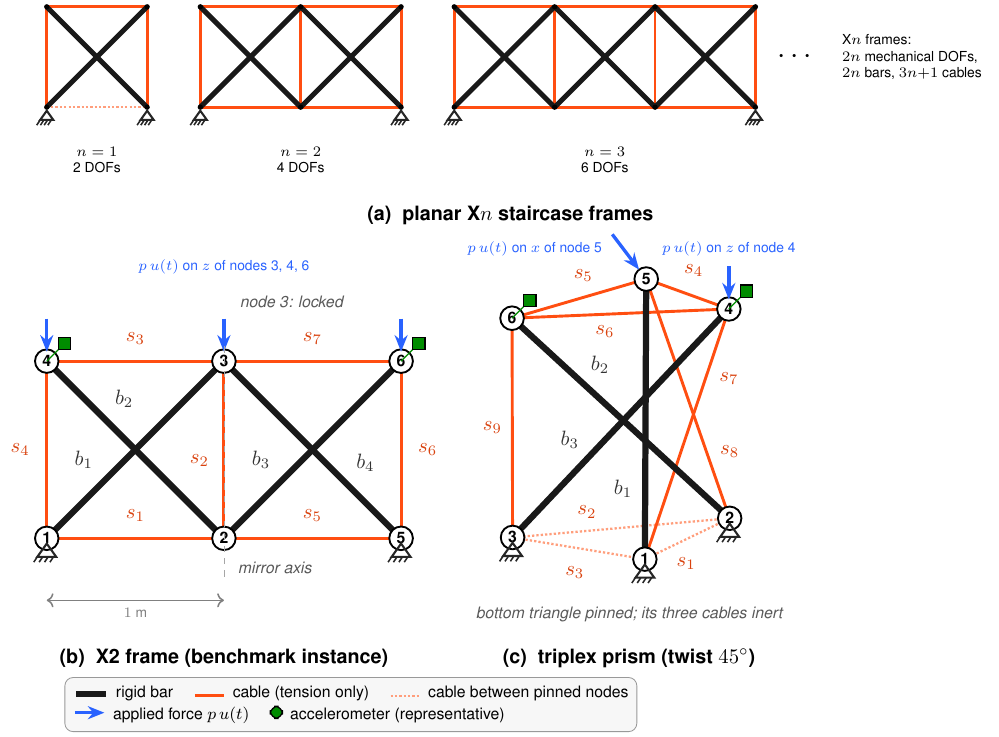}
  \caption{The benchmark structures, drawn from the simulator geometry:
  \textbf{(a)}~the planar X$n$ staircase frames, \textbf{(b)}~the X2
  instance, \textbf{(c)}~the triplex prism. Blue arrows mark the loaded
  coordinates and green squares a representative accelerometer
  placement; properties are given in Table~\ref{tab:structures}.}
  \label{fig:benchmarks}
\end{figure*}

Figure~\ref{fig:benchmarks} presents the benchmark structures. The planar X$n$ frames (Figure~\ref{fig:benchmarks}a) are staircases of Snelson X-modules, each the classical assembly of two
crossed struts and four perimeter cables \citep{Snelson2012, Motro2003}.
Chaining $n$ modules gives $2(n{+}1)$ nodes, $2n$ bars (the module
diagonals), and $3n{+}1$ cables. With the two end base nodes pinned and motion confined to the plane, the frame has $2n$ mechanical degrees of freedom (DOFs): its $4n$ free nodal coordinates less one length constraint for each of its $2n$ bars. The X1 frame ($n{=}1$, two DOFs) is the entry point for the identification experiments. Its base cable joins the two pinned nodes and therefore transmits no information to the dynamics. That
member is provably unidentifiable, a built-in control for the
identifiability analysis. The X2 frame (Figure~\ref{fig:benchmarks}b) has the geometry of the earlier simulations on which this study builds. Its node~3 is kinematically
locked by its two bars to the pinned supports, and the structure and
default forcing are mirror-symmetric. The evaluation metrics treat
both properties explicitly. The code that generates the frames is verified against the earlier X2 implementation (Table~\ref{tab:verification}). Scale-up
studies use $n$ as the size parameter.

The triplex prism (Figure~\ref{fig:benchmarks}c) is the
three-dimensional benchmark: bottom triangle pinned, top triangle free, three bars, and nine cables, which leave six DOFs from nine free coordinates. Its
construction reproduces two classical results, independent checks of
the mechanics. The free-floating pure self-stress exists only at the twist angle of $30^\circ$, the value $90^\circ - 180^\circ/n_b$ for a prism with $n_b = 3$ bars. In that state, the force densities of the bars, the cross cables, and the triangle cables are in the ratio $-t : t : t/\sqrt{3}$, with $t$ the prestress magnitude. Under gravity with the base pinned, the members must also carry their own weight. The linear program of Section~\ref{sec:cablelaw}, with every cable force density at least $1$\,N/m, then finds a prestress with all bars in compression for twist angles from just above $30^\circ$ to about $58^\circ$, for the bars of this study (radius $5$\,cm, density $500$\,kg/m$^3$). The lower bound is the self-stress twist, and the upper bound grows with the self-weight relative to the minimum force density. The benchmark is built at $45^\circ$, inside this range. The prestress of the gravity-loaded prism is fixed uniquely by equilibrium; the only free self-stress states involve the bottom cables, which do not affect the dynamics. The prism therefore poses an inverse problem with
a unique answer. The X$n$ frames, by contrast, have free self-stress states and are used for the studies at several prestress levels. The prestress series of the prism is generated on the zero-gravity
$30^\circ$ configuration, where the force densities scale linearly
with the prestress level.

\subsection{Excitation suite, sensor protocols, and noise model}
\label{sec:excitation}

Each record is generated by forward simulation from rest in the
prestressed equilibrium fixed by the linear program of
Section~\ref{sec:benchmarks} ($q(0) = q_0$, $\dot q(0) = 0$). A time-dependent force $f_{\mathrm{ext}}(t)$ is then applied at one or more coordinates while Eq.~\eqref{eq:eom} is integrated forward. A harmonic record applies at coordinate $d$ the force
\begin{equation}
  f_{\mathrm{ext},d}(t) \;=\; p\,\sin(2\pi f_0 t),
  \label{eq:forcing}
\end{equation}
with amplitude $p$ (N) and frequency $f_0$ (Hz). Three further classes
replace this scalar signal at the same driven coordinate: a half-sine
impulse, a band-limited earthquake-like signal, and stationary Gaussian
noise. The earthquake-like class is white noise, low-pass filtered,
shaped by a trapezoidal envelope, and normalized to its peak. The repeated runs with multi-point excitation in Section~\ref{sec:res-family} superpose signals of
the form of Eq.~\eqref{eq:forcing} at two to four coordinates with
distinct amplitudes and frequencies.

The two random classes are generated from fixed seeds, purely so that
every dataset can be regenerated exactly; the seeds play no
methodological role. These dataset seeds are also distinct from the
training seeds used later: the records are fixed once, while the
identification is repeated from several random network initializations
and noise realizations to quantify the scatter of the identified values
(Section~\ref{sec:res-baseline}).

The simulated tests cover the X2 frame at three prestress levels
under harmonic and impulse forcing, with earthquake-like forcing added
at the lowest level, together with the gravity-loaded prism and the
zero-gravity prism at three prestress levels. At each
combination the forcing amplitude is set on both sides of the
slackening threshold, the smallest amplitude at which the forward
simulation's event tracker reports a cable leaving tension. Below that threshold a record stays strictly taut; above it the records contain many taut--slack transitions.

Every record spans 5\,s at a 0.01\,s output step, integrated at
relative and absolute tolerances of $10^{-8}$ with continuous tracking of taut--slack transitions. The stored outputs of each record are the
positions, velocities, accelerations, cable tensions, per-cable taut
flags, and the times of the taut--slack transitions.

The sensor data are the instrumented-coordinate accelerations of the
integrated trajectory, with Gaussian noise added at 2\,\% of each
channel's root-mean-square. The sensor protocols of the identification experiments subsample the free coordinates under the same 2\,\% noise.
The retained fraction ranges from four-ninths to two-thirds across the
demonstrations of Section~\ref{sec:results}; the X1 frame uses half.
The remaining coordinates are held out for the posterior-predictive
check of Section~\ref{sec:res-1cell}.

\subsection{Constraint and energy checks}
\label{sec:audits}

Two verification checks are evaluated along every stored record.
The first is the relative violation of the squared bar length; since a
relative length error $\epsilon$ violates the squared length by
$2\epsilon$ to leading order, a reported violation $v$ corresponds to
a length error of $v/2$ and converts directly to a physical length.
The second is the energy-balance error $|E(t) - W(t) - E(0)|$, with
$E$ the total mechanical energy and $W$ the work integral of the
external and damping forces; the balance holds across taut--slack
transitions because a slack cable stores no strain energy.

On the records used for identification the worst figures, at the end
of the record, are a length drift of $0.33$\,\textmu m on the
$1.414$\,m bar diagonal and a relative energy error of
$5\times10^{-6}$. These bound the numerical error the integration
carries into the inverse problem; they establish that the simulated
records are far more accurate than the $2$\,\% sensor noise added to
them. The same checks apply unchanged to trained networks, where they measure the physical consistency of reconstructed
motions.

\FloatBarrier
\section{Methodology}
\label{sec:methodology}

\subsection{Physics-informed inverse formulation}
\label{sec:pinn}

\begin{figure*}[!tbp]
  \centering
  \includegraphics[width=\linewidth]{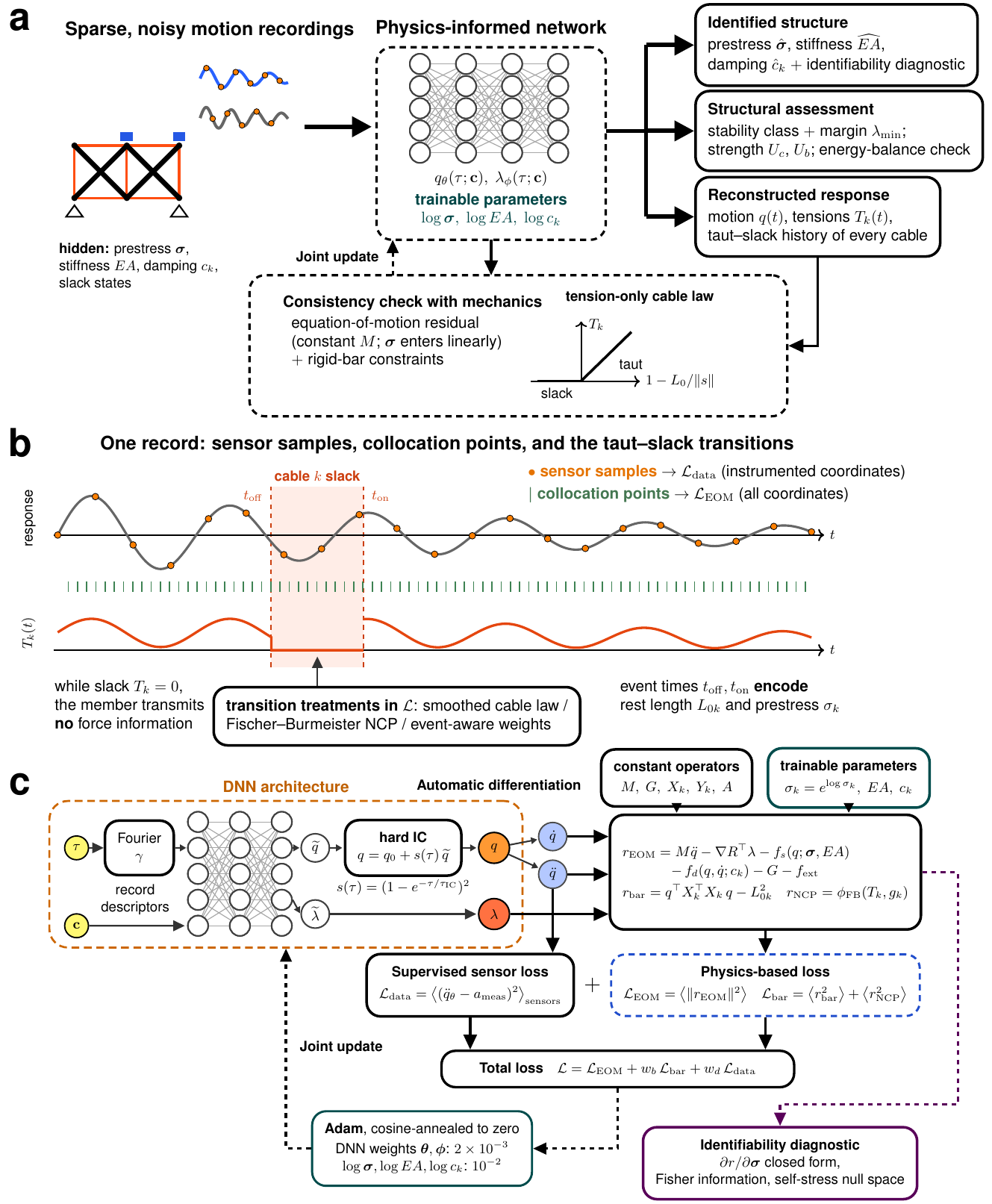}
  \caption{Physics-informed identification framework:
  \textbf{(a)}~pipeline from sparse, noisy accelerometer records to
  identified member parameters and the structural assessment;
  \textbf{(b)}~a single record, with sparse sensor samples feeding the data
  loss and dense collocation points enforcing the physics residual through the taut--slack transitions; \textbf{(c)}~the computational
  graph, with the member parameters entering the residual as trainable
  inputs.}
  \label{fig:arch}
\end{figure*}

\subsubsection{Network parameterization and hard constraints}
A displacement network and a multiplier head share a trunk, drawn as
the computational graph of Figure~\ref{fig:arch}c. The trunk maps the
scaled time $\tau = t/T_{\mathrm{ref}}$, embedded in a fixed Fourier
feature basis, through a $\tanh$ multilayer perceptron; velocities and
accelerations follow by automatic differentiation of the outputs in
time. Outputting the constraint multipliers alongside the
displacements places the formulation in the mixed-variable,
hard-constraint PINN class \citep{RaoSunLiu2021Elastodynamics,
Lu2021HardConstraintPINN}.

The displacement head outputs only the free coordinates, and
$q_\theta(\tau) = q_0 + s(\tau)\,B\,\widetilde u_\theta(\tau)$ imposes
the fixed ones by construction: $B$ selects the free coordinates and
the window $s(\tau) = (1 - e^{-\tau/\tau_{\mathrm{IC}}})^2$
enforces $q(0) = q_0$ and $\dot q(0) = 0$ exactly, so no
initial-condition loss is needed.

The multiplier head outputs the full vector $\lambda_\phi(\tau)$ and
is initialized at static equilibrium multipliers. In the synthetic
benchmarks these are computed from the exact prestress state, a value
available only because the benchmark is synthetic. That value is used only for the initialization and is never a trainable parameter. In an application, the same initialization would be computed from the nominal or design prestress, and an error in it would change only the starting point of the head. Every transition treatment uses this initialization. For multi-record training the trunk is additionally conditioned on record descriptors $\mathbf{c}$.

\subsubsection{Equation-of-motion residual with trainable member parameters}
The physics residual evaluates Eq.~\eqref{eq:eom} on the network
outputs,
\begin{equation}
  r_{\mathrm{EOM}} = M\ddot q_\theta - \nabla R(q_\theta)^{\top}\lambda_\phi
  - f_s(q_\theta;\boldsymbol\sigma, EA) - f_d(q_\theta,\dot q_\theta;\{c_k\})
  - G - f_{\mathrm{ext}},
  \label{eq:residual}
\end{equation}
in which the force densities $\boldsymbol\sigma$, the axial stiffness,
and the damper coefficients are trainable parameters, represented in logarithmic form to enforce positivity. The external forcing
$f_{\mathrm{ext}}$ is a known input, as in a controlled excitation test
with prescribed, recorded load. An unknown or ambient forcing would add
its amplitude and phase as unknowns; that is left to future work. All
structural operators in Eq.~\eqref{eq:residual} are constant tensors
precomputed from the incidence matrices, so the residual applies
unchanged to any tensegrity topology. Because the residual is linear in $\boldsymbol\sigma$, the sensitivities $\partial r_{\mathrm{EOM}}/\partial\sigma_k$ are closed-form, and Section~\ref{sec:identifiability} uses this property in the identifiability analysis. The residual implementation is verified by the checks collected in Table~\ref{tab:verification}; training itself uses the smoothed cable law described below.

Throughout this study the exact parameters of a synthetic benchmark
are used in three places only: to verify the residual implementation,
to evaluate the reference Fisher analysis of
Section~\ref{sec:identifiability}, and to score the recovered values
after a run. The identification itself uses only the noisy sensor records, the known forcing, the geometry and supports, and the stated initial guesses.

The training objective combines three terms: a sensor-data loss on
measured accelerations at instrumented coordinates, the mean squared
residual of Eq.~\eqref{eq:residual} on fresh collocation points
resampled each step, and the squared bar-length constraint residual,
\begin{equation}
  \mathcal{L} \;=\; \mathcal{L}_{\mathrm{data}}
  \;+\; \mathcal{L}_{\mathrm{EOM}}
  \;+\; w_b\,\mathcal{L}_{\mathrm{bar}},
  \label{eq:loss}
\end{equation}
each normalized by its natural scale: the pooled mean-square measured
acceleration over the instrumented channels, the squared forcing
amplitude, and the squared bar length. The data normalizer, a single
global response power, stays bounded away from zero for a forced
record, so no denominator regularization is needed.

Because each term of Eq.~\eqref{eq:loss} is divided by its own scale, the remaining weights are dimensionless, and data and physics are weighted equally. A
sensitivity study in Section~S2 of the Supplementary Material varies
both weights on the X1 taut record. Recovery is flat to within about 2 percentage points across three decades of $w_b$ around the value
used throughout ($w_b = 100$), and worsens sharply only when the bar constraint is effectively unenforced. Changing the data-to-physics
ratio between $0.3$ and $3$ changes the mean force-density error by about 1 percentage point; at $10$, where the sensor data are effectively ignored, the error increases sharply. Under model-form error,
down-weighting the suspect physics makes the identification slightly worse ($6.0$ against $5.8$\,\% at a ratio of
$0.1$), because the residual's error is structured by the geometry and
still informs the parameters.

A single Adam optimizer \citep{KingmaBa2015} updates the network
weights and the physical parameters jointly, in two groups with
learning rates of $2\times10^{-3}$ and $10^{-2}$, both cosine-annealed
to zero over the run \citep{LoshchilovHutter2017}. The physical parameters are represented in logarithmic form, which makes their updates multiplicative. One relative step size therefore serves force densities, an
axial stiffness, and damper coefficients whose values differ by orders
of magnitude (details in Section~S2 of the Supplementary Material). No
gradient clipping is applied and no per-structure tuning is performed:
the same two rates are used for every result reported here.

Table~\ref{tab:structures} lists the benchmark structures; the
full table of simulation, sensing, and training parameters, with the
origin of each value, and the archival scheme that reproduces every
run exactly are given in Section~S2 of the Supplementary Material. The X1 demonstrations of
Section~\ref{sec:res-1cell} use a four-layer trunk of width 64, 256
collocation points per step, and 25{,}000 steps; the scale-up grid of Section~\ref{sec:res-family} uses the same architecture at 12{,}000 steps.

The values in these tables come from different sources. The member properties of the planar benchmarks, cables of radius $1$\,mm and modulus $70$\,MPa and rigid bars of radius $5$\,cm and density $500$\,kg/m$^3$, and their geometry are those of the earlier simulations on which this study builds; they were kept so that the frame generator could be verified against those simulations (Table~\ref{tab:verification}). The excitation amplitudes are set on both sides of the slackening threshold of each structure (Section~\ref{sec:excitation}), and the sensor noise and the sensor count are varied in Section~\ref{sec:res-robust}. The method settings, namely the network size, the learning rates, the loss weights, and the smoothing sharpness $\beta$, were fixed once and used unchanged from the two-degree-of-freedom frame to the 16-cable bridge and the three-dimensional prism, over a 15-fold range of prestress and sensor noise from $2$ to $10$\,\%. The sensitivity of the results to the loss weights is reported in Section~S2 of the Supplementary Material, and the bias introduced by $\beta$ is bounded in Section~\ref{sec:switch}.

\begin{table}[!tbp]
  \centering
  \caption{Benchmark structures used for identification. The classic tensegrity forms assessed but not identified are listed in Section~S3 of the
  Supplementary Material.}
  \label{tab:structures}
  \footnotesize
  \setlength{\tabcolsep}{5pt}
  \begin{tabular}{@{}lccccl@{}}
\toprule
Structure & Nodes & Bars & Cables & Mech.\ DOFs & Prestress source\\
\midrule
X1 frame                       & 4  & 2 & 4  & 2         & LP, 25\,N/m min.\ force density\\
X2 frame                       & 6  & 4 & 7  & 4         & LP, 25\,N/m min.\ force density\\
X3 frame                       & 8  & 6 & 10 & 6         & LP, 25\,N/m min.\ force density\\
X5 bridge (two-span)             & 12 & 10 & 16 & 8        & LP, 25\,N/m; intermediate support\\
Triplex prism ($45^\circ$, gravity)  & 6  & 3 & 9  & 6         & uniquely fixed by equilibrium\\
Prism series ($30^\circ$, zero-$g$)  & 6  & 3 & 9  & 6         & self-stress scale 1--15\,N/m\\
Hexagonal prism (laboratory)         & 12 & 6 & 18 & free--free & single self-stress mode\\
\bottomrule
\end{tabular}

\end{table}

\subsubsection{Treatments of the taut--slack transition}
\label{sec:switch}
The tension-only law of Eq.~\eqref{eq:cablelaw} is a complementarity
condition, and a gradient-based optimizer must cross the taut--slack
boundary repeatedly while training. A force law whose derivative jumps at that boundary produces discontinuous gradients there.
The primary treatment therefore replaces the ramp in the force density with a smoothed (softplus) cable law,
\begin{equation}
  \sigma_k(q) \;=\; K_k\,
  \operatorname{softplus}_\beta\!\bigl(1 - L_{0k}/\lVert s_k\rVert\bigr),
  \qquad
  \operatorname{softplus}_\beta(x) \;=\; \tfrac{1}{\beta}\log\bigl(1 + e^{\beta x}\bigr),
  \label{eq:smoothed}
\end{equation}
and smooths the on--off switching of the damping by the matching sigmoid. Because the smoothed cable law is differentiable, the gradients that reach the optimizer remain continuous across the taut--slack boundary, where the exact law has a kink.

The smoothing introduces a bias, and the sharpness $\beta$ is chosen to keep it small. The bias is confined to a boundary layer of width $1/\beta$ in the strain coordinate, where the softplus exceeds the exact ramp by at most $\ln 2/\beta$. With $\beta = 800$, the layer width is $1.3\times10^{-3}$, the largest overshoot is $9\times10^{-4}$, and the peak bias is below $1$\,\% of the recovered force densities. The same value is used for all structures and records. Outside the boundary layer, the bias decays rapidly to zero.

Two alternatives are compared with the smoothed cable law on identical records in Section~\ref{sec:res-treatments}. The complementarity treatment keeps the exact law. It removes the nonsmoothness from the residual by adding an output head for the non-negative cable tensions, and it links each tension to the elastic law through a Fischer--Burmeister complementarity residual, a function that is zero exactly when the tension-only conditions hold, adapting the static-contact construction of
\citet{Sahin2024Contact} to the dynamic setting. The event-aware treatment keeps the smoothed cable law and concentrates half of each batch's collocation points near the taut--slack transitions detected in the current solution. Both are specified in Section~S1 of the Supplementary
Material. All three treatments share the same trunk architecture, loss
weights and normalizations, and optimizer settings; none was tuned separately.

\subsection{Adequacy of the nodal member model}
\label{sec:member-pde}
The formulation above is nodal: cables are massless axial force
elements, bars are rigid links, and member motion follows the node motion. This representation is used for all identification results reported in this study. It does, however, discard the member's own dynamics, and near slack this omission introduces a model-form error. A member-resolved alternative treats each member as a one-dimensional continuum coupled to its neighbors at the nodes.

The two descriptions agree except at very small cable tensions. Bending and tension terms balance at
$T^{\star} = E_cI_c(\pi/L_c)^2$, about $5\times10^{-4}$\,N for the cables of
this study, against working tensions of $10$ to $130$\,N. While the cable
carries load, the flexural term is therefore four to five orders of
magnitude below the tension term. The taut-string idealization, however, degenerates exactly at slack: its transverse wave speed $\sqrt{T_c/\rho_cA_c}$ falls to zero, and so does its fundamental frequency. Restoring the flexural term removes the degeneracy and leaves a finite frequency there. The member-resolved model is therefore used only as a diagnostic, to show when the nodal idealization becomes inaccurate; the ratio $T^{\star}/T$ measures by how much.

\subsection{Identifiability analysis}
\label{sec:identifiability}

Although Eq.~\eqref{eq:residual} is linear in the force densities, the map from the parameters to the sensor accelerations is nonlinear, because it passes through the solution of the equations of motion. The sensitivity of this map,
$S(t) = \partial a_{\mathrm{sensors}}(t)/\partial\boldsymbol\theta$, is therefore obtained
by central finite differences of the verified forward model (relative
step $10^{-3}$, floored at $10^{-6}$). The
Fisher information matrix of the parameter set
$\boldsymbol\theta = (\boldsymbol\sigma, c)$ then follows for a
given record, sensor set, and noise level:
$J = \sum_t S(t)^{\top} \Sigma_n^{-1} S(t)$. Here
$\Sigma_n = \operatorname{diag}(s_{n,i}^2)$ collects the per-channel
noise variances, with $s_{n,i}$ the standard deviation of sensor
channel $i$: $2\,\%$ of that channel's root-mean-square. The two
instrumented channels have comparable RMS, so these bounds coincide
with the equal-variance form to the precision reported below. The same
information matrix underlies parameter identifiability and optimal
sensor placement in structural dynamics
\citep{Udwadia1994, Kammer1991}.

The analysis is local to the reference parameters. After removing the
near-null direction spanned by the inert base cable, the identifiable
block of $J$ is only moderately conditioned: its condition number is about
$1.9\times10^{2}$ on the taut record and $5.6\times10^{2}$ on the slack-transition record. Locality does not rule out practical non-uniqueness, in which a run fits the record closely and still returns wrong parameters. This study calls such a run a silent failure; Sections~\ref{sec:res-family} and~\ref{sec:res-bridge} report one on the X2 taut record and one on the bridge.

\subsubsection*{Concentration of prestress information in mechanism modes}

The numerical analysis above measures how much information a record
carries. A modal argument shows where that information comes from, before any data are taken.

Linearized about an equilibrium configuration, the tangent stiffness
separates into a material part and a geometric part,
\begin{equation}
  K(\boldsymbol\sigma, EA) \;=\; K_e(EA) \;+\; K_g(\boldsymbol\sigma),
  \qquad
  K_g(\boldsymbol\sigma) \;=\; \sum_k \sigma_k\, G_k ,
  \label{eq:tangent-split}
\end{equation}
in which the operators $G_k$ depend on the geometry alone, so $K_g$ is
\emph{linear} in the force densities. This is the same linearity that
Eq.~\eqref{eq:residual} exploits, seen in the modal domain. For a mode
$\phi_i$ normalized so that $\phi_i^{\top} M \phi_i = 1$,
\begin{equation}
  \omega_i^2 \;=\; \underbrace{\phi_i^{\top} K_e \phi_i}_{\text{elastic}}
  \;+\; \underbrace{\phi_i^{\top} K_g \phi_i}_{\text{geometric}},
  \qquad
  \frac{\partial \omega_i^2}{\partial \sigma_k}
  \;=\; \phi_i^{\top} G_k\, \phi_i ,
  \label{eq:modal-split}
\end{equation}
the sensitivity following from the standard eigenvalue derivative
together with the linearity of $K_g$.

The two terms of Eq.~\eqref{eq:modal-split} behave differently. An
infinitesimal mechanism is a displacement pattern that the compatibility
operator annihilates. Its members do not extend to first order, so
$\phi_i^{\top} K_e \phi_i \simeq 0$: the frequency is carried entirely by
the geometric term. That term is homogeneous of degree one in
$\boldsymbol\sigma$. Scaling the self-stress by $\lambda$ therefore scales
$\omega_i^2$ by exactly $\lambda$, and the relative sensitivity reaches its
largest possible value,
\begin{equation}
  \frac{\partial \ln \omega_i^2}{\partial \ln \lambda}
  \;=\; \frac{\phi_i^{\top} K_g \phi_i}
             {\phi_i^{\top} K_e \phi_i + \phi_i^{\top} K_g \phi_i}
  \;\longrightarrow\; 1
  \quad\text{as}\quad \phi_i^{\top} K_e \phi_i \to 0 .
  \label{eq:geometric-fraction}
\end{equation}
A mode stiffened elastically is at the other end of the same expression: with $\phi_i^{\top} K_e \phi_i \gg \phi_i^{\top} K_g \phi_i$
the ratio is small and the frequency is nearly insensitive to prestress.

Eq.~\eqref{eq:geometric-fraction} is a property of the geometry and the
mode shape, computable before any record is taken, and it accounts for
observations that otherwise look unrelated. The rotational mode of the
simplex of \citet{Malyszko2020Simplex} is a mechanism, and its squared
frequency is proportional to the self-stress scale in the present rebuilt model over a 12-fold range, the intercept of a linear fit being
$3\times10^{-6}$ of the mean, while the translational modes of the same
structure are flat across all 13 measured prestress levels. Two full-scale structures measured at the same laboratory show the two extremes. The
five-module active tensegrity shifts its lowest frequency by about $29$\,\%
over a $1.7$-fold force change \citep{BelHadjAliSmith2010IJSS}. The
footbridge moves about $1$\,\% while its self-stress quadruples
\citep{BelHadjAli2010}. The ratio in Eq.~\eqref{eq:geometric-fraction} is
near 1 for the first and near 0 for the second. The X-frames of this study, whose frequencies shift about $12$\,\% over a 50-fold prestress change, lie between them.

The identification of Section~\ref{sec:pinn} is a time-domain residual inversion and does not use modes. Eqs.~\eqref{eq:modal-split} and~\eqref{eq:geometric-fraction} therefore do not describe the identification method itself;
they describe the structure of the information the record contains, which
is what the Fisher matrix above then measures numerically. Also, a small geometric fraction limits what the \emph{frequencies} reveal about prestress; a full acceleration record can reveal more. The transient formulation recovers the force densities of the X-frames where a modal method would have little to work with (Section~\ref{sec:res-family}).

\begin{figure*}[!tbp]
  \centering
  \includegraphics[width=0.96\textwidth]{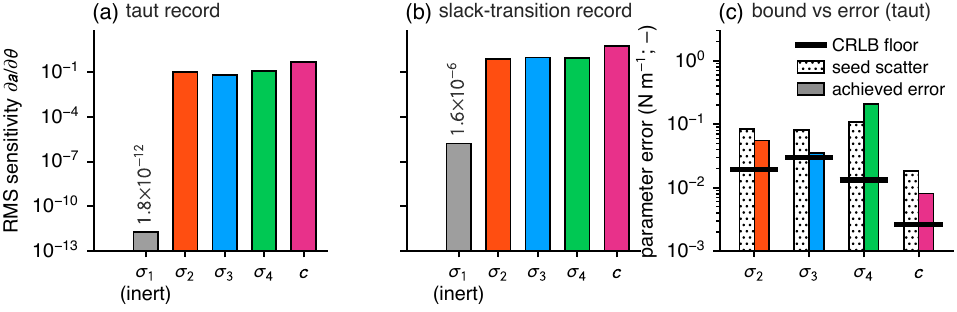}
  \caption{Model-based identifiability analysis of the X1 frame.
  \textbf{(a, b)}~Root-mean-square sensitivity of the sensor
  accelerations to each parameter on the taut and slack-transition records. \textbf{(c)}~Cram\'er--Rao lower bounds at $2$\,\% noise
  (floor lines) against the multi-seed scatter and the single-run
  achieved error on the taut record.}
  \label{fig:ident}
\end{figure*}

The analysis is based on three structural facts: which members the free coordinates cannot detect, how precisely the remaining members can be determined, and how much information is lost during the slack intervals. First, members whose ends are both restrained transmit no
force to the free coordinates. Neither node moves, so the column of $\partial f/\partial\bm\sigma$ belonging to such a member is identically zero. The member's column of the sensitivity matrix $S$, whose rows are the sensor samples and whose columns are the parameters, is then zero as well, and so are its row and column of $J$. These members span the null space of $J$. On the X1 frame the base cable's sensitivity is
about 11 orders of magnitude below the identifiable members' on the taut record, and about 6 orders below on the slack-transition record. $J$
itself has exactly one near-null eigenvalue, matching exactly one
inert member (Figure~\ref{fig:ident}). A taut--slack transition cannot remove this null space. A transition changes which members carry force, thereby redistributing information among the directions that the equilibrium operator determines, but it adds no information to the part of that operator's null space that remains in every configuration.
A doubly restrained member is invisible to the free coordinates
whether taut or slack.

A self-stress combination among members that do move is a weaker case. It produces no free-coordinate force at the reference configuration. The operator, however, depends on the configuration, and the combination leaves the null space as soon as the structure displaces. A record with sufficiently strong excitation therefore provides information about it. The distinction is testable before any data are taken: which members
have both ends restrained is a property of the geometry and the
supports alone. The intervention study of
Section~\ref{sec:res-ladder} confirms this: richer excitation and taut--slack transitions recover the identifiable members, while more sensors and more records
do not.

Applied to the 12 classic tensegrity forms assessed in Section~S3 of the Supplementary Material, this argument determines $164$ of $183$ cables at first order: 17 of the 19 undetermined cables are
individually invisible members spanning two restrained nodes, and the
other two are first-order self-stress combinations on the X2 and X3
frames. Identifiability in these structures depends on the support conditions, not on their size or redundancy, and the remedy differs: an invisible member needs a support layout or an
instrumentation plan, since no record of any length or amplitude will
substitute, while a first-order combination needs only a record with larger excitation.

The same algebra shows how the null direction can be removed. Suppose the members are tied into groups, each group sharing one tension or one force density, so that $\bm\sigma = \bm\Gamma^{\top}\bm g$. Here $\bm g$ holds one value per group, and row $i$ of the grouping matrix $\bm\Gamma$ holds the coefficients that map the value of group $i$ to its members. By the chain rule, the sensitivity to the group values is $S\bm\Gamma^{\top}$, and the Fisher information becomes $\bm\Gamma J\bm\Gamma^{\top}$. This matrix keeps a null direction of $J$ only if that direction lies in the range of $\bm\Gamma^{\top}$, that is, in the row space of $\bm\Gamma$. The members with nonzero entries in the null direction form its support, and the null direction cannot lie in the row space once a member of its support shares a group with a member outside it.

On the X2 frame, grouping can be examined both for static equilibrium and for the dynamic identification, and the two cases differ. For static equilibrium, the equilibrium sensitivity $\partial\bm F/\partial\bm\sigma$ of the seven independent cables gives a Gram matrix with eigenvalues $\{0,1,1,1,2,2,3\}$. The zero eigenvalue makes this parameterization singular: an equal increase of the force densities in the two bottom-chord cables leaves the forces at the free coordinates unchanged. When three cables are tied into one routed group, the smallest eigenvalue rises from 0 to $1.0$, and the parameterization is no longer singular.

The dynamic identification uses a different matrix, the Fisher information $J$, assembled from the acceleration sensitivities over the whole record. Under the taut X2 protocol, $J$ is not singular even for independent cables; its smallest eigenvalue is $76$. The static null direction does not remain in the dynamic problem, because a change of force density alters the geometric stiffness even where it leaves the reference-configuration force balance unchanged. The acceleration sensitivity along that direction is $11\,\%$ of that of the strongest member, which is six orders of magnitude above the finite-difference noise floor.

Grouping still improves the dynamic problem. With the same routed group, the smallest eigenvalue of $J$ rises from $76$ to $2.8\times10^{2}$, and at $2\,\%$ sensor noise the relative Cram\'er--Rao bound of the worst grouped force density is $0.09\,\%$. Both analyses use the same structure and the same record and differ only in the parameterization, so the improvement comes from the parameterization alone. Grouping is the
standard remedy for ill-conditioned updating
\citep{Friswell1998Subset, Shahverdi2009Clustering}. The tensegrity
setting adds that the groups need not be chosen for
numerical convenience: the clustered cables routed through
frictionless pulleys for actuation \citep{Moored2009Clustered} already share a tension by construction. Such a structure does not have the static singularity that the same geometry with independent cables brings to any equilibrium-based identification.

Second, the diagonal of the inverse of the identifiable block gives the
Cram\'er--Rao lower bounds. At 2\,\% sensor noise the best attainable
standard deviations on the X1 frame are $0.013$--$0.030$\,N/m for the
force densities and $0.0026$\,N\,s/m for the damper coefficient. The
identification of Section~\ref{sec:res-1cell} achieves a single-run
damping error within a few percent of that bound, while its force-density errors are within a factor of 20 of theirs. That gap comes from the bias of the optimization and the smoothing. The bound limits the scatter of repeated identifications, not the error of a single run. The
eight-seed study of Section~\ref{sec:res-baseline} finds the achieved dispersion $2.7$ to $8.3$ times above it. The identification is therefore accurate but not statistically efficient.

Third, a slack interval removes the force information of the slack member, while the transition times themselves carry information about rest lengths. On the slack-transition record the bound for the most-loaded cable increases by a factor of about two relative to the taut record. That factor measures the information lost to the slack intervals. On this record, the central-difference sensitivities span transition instants where the acceleration is nonsmooth, and this bound is therefore approximate.

\subsection{Baselines}
\label{sec:baselines}

Two classical methods are applied to the same records, the same noisy data,
and the same initial parameter guesses as the physics-informed
identification. The first is deterministic time-domain model updating
by least-squares shooting: Levenberg--Marquardt
\citep{Levenberg1944, Marquardt1963} over
$(\log\boldsymbol\sigma, \log c)$, with every residual evaluation a full
forward simulation of the verified solver. The second is a joint
state--parameter unscented Kalman filter
\citep{JulierUhlmann1997UT, WanVanDerMerwe2000UKF}
on the augmented state of free
coordinates, velocities, and log-parameters. It propagates that state
by the same equations of motion under a fixed-step Runge--Kutta scheme
with a parameter random walk. Symmetric-eigenvalue matrix square roots,
covariance eigenvalue bounds, and state clipping stabilize the filter
and keep the sigma-point propagation numerically stable. Both use the exact forward model, and the comparison therefore separates the identification strategy from modeling error. Classical modal updating is used where
measured frequency-response data are the target \citep{Prism2025JSV}.

\subsection{Structural assessment of the identified model}
\label{sec:method-assessment}
The recovered force densities describe the prestress state of the structure, and they are the input to a downstream assessment that requires no further
measurement: the tangent stiffness at the identified equilibrium
classifies the self-stressed shape as super-stable, stable, or
unstable, and the member forces yield strength and buckling
utilizations for the cables and struts. The formulation, the classification conditions, the material library, and the energy-balance check that accompanies them are set out in Section~S5 of the Supplementary
Material. The property that matters for identification, how the
assessment responds to identification error, is reported in
Section~\ref{sec:res-assessment}.

\FloatBarrier
\section{Results}
\label{sec:results}

\subsection{Verification of the physics residual}
Before any identification, two checks confirm that the physics residual is implemented correctly. First, the residual must reproduce the equations of motion of the verified simulator, including configurations with slack cables. Second, with the exact parameters, the residual must be zero on the reference trajectories: exactly with the exact cable law, and up to the known smoothing bias with the smoothed cable law. Table~\ref{tab:verification} collects these and the other numerical checks in one place, and the precision figures are not repeated elsewhere. These checks separate errors in the physics from errors of the optimization in all the results that follow.

\begin{table}[!tbp]
  \centering\footnotesize
  \caption{Verification checks of the numerical implementation, as relative figures. The checks of Section~\ref{sec:audits} bound the error of the simulated records themselves.}
  \label{tab:verification}
  \begin{tabular}{@{}lc@{}}
  \toprule
  Check & Level\\
  \midrule
  Residual against simulator state derivatives, slack included & $10^{-15}$\\
  Residual at exact parameters, exact cable law & $10^{-15}$\\
  Smoothed-cable-law peak force-density bias at $\beta=800$ & $<1$\,\%\\
  Simulator against classical-limit truss frequencies & $3\times10^{-9}$\\
  Frame generator against earlier X2 frequencies & $10^{-6}$\\
  Flexible-bar residual against flexible-bar simulator & $10^{-9}$\\
  \bottomrule
  \end{tabular}
\end{table}

\subsection{Demonstration on the X1 frame}
\label{sec:res-1cell}

The inverse experiments begin on the X1 frame at prestress level
$\sigma_{\min} = 25$\,N/m. Its exact force densities are
$\boldsymbol\sigma = (25, 52.2, 25, 25)$\,N/m for the base, top, and
two vertical cables. The damper coefficient is $c = 1$\,N\,s/m, and the axial stiffness is fixed at its exact value. The forcing acts on the vertical coordinate of one top node. The sensors record noisy accelerations, with $2$\,\% noise, at the two coordinates of that node, which are half the free coordinates. Two records are inverted under the smoothed cable law: a taut record ($p = 10$\,N, no taut--slack transitions) and a slack-transition record ($p = 40$\,N, 20 taut--slack transitions).

\begin{figure*}[!tbp]
  \centering
  \includegraphics[width=0.92\textwidth]{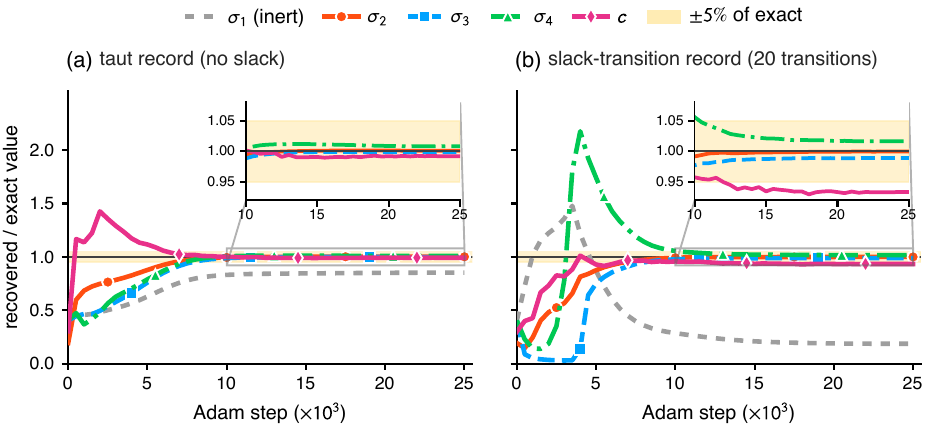}\\[1pt]
  \includegraphics[width=0.96\textwidth]{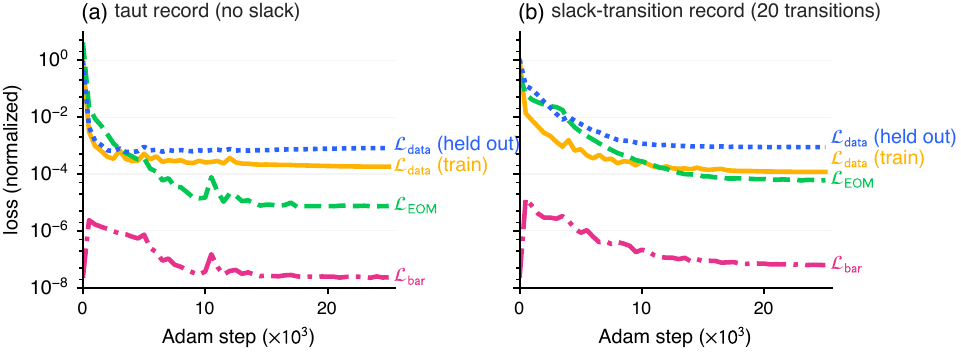}
  \caption{The X1 inversions on (a)~the taut record and (b)~the slack-transition record with 20 taut--slack transitions: parameter convergence
  normalized by the exact values (top, $\pm 5$\,\% band shaded) and
  the training histories (bottom). The inert base cable does not converge to its value, as expected for a member that cannot be identified.}
  \label{fig:conv1}
\end{figure*}

\begin{figure*}[!tbp]
  \centering
  \includegraphics[width=0.78\textwidth]{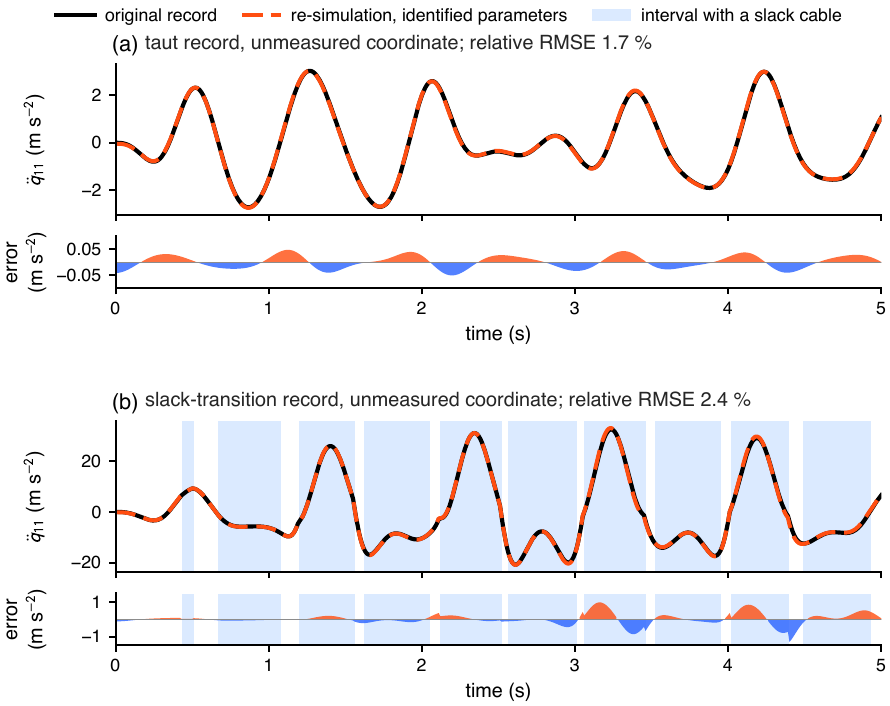}
  \caption{Posterior-predictive validation for (a) the taut record and (b) the slack-transition record: the structure re-simulated with the recovered
  parameters under the same forcing, compared with the original record
  at a coordinate that was never measured. Shading marks the intervals with a slack cable; the strip below each trace is the signed error.}
  \label{fig:val1}
\end{figure*}

In the taut regime, the three identifiable force densities are recovered
to relative errors of $0.33$, $0.37$, and $1.0$\,\%, and the damper
coefficient to $0.27$\,\% (Figure~\ref{fig:conv1}, top).
Reconstruction at the two unmeasured coordinates
errs by $1.3$\,\% of the motion scale. Re-simulation with the
recovered parameters reproduces an unmeasured acceleration to a
relative root-mean-square error of $1.7$\,\% (Figure~\ref{fig:val1}).
The slack-transition record gives similar force-density errors, $0.20$\,\% for the top cable and $1.1$ and $1.3$\,\% for the verticals, through 20 taut--slack transitions with the smoothed cable law alone. Its damping error is larger, $6.2$\,\%. Its reconstruction error
is $0.8$\,\% of the motion scale, and re-simulation reproduces the
unmeasured acceleration to $2.1$\,\% (Figure~\ref{fig:val1}). Training
histories for both records appear in Figure~\ref{fig:conv1} (bottom),
and multi-seed uncertainty statistics in Section~\ref{sec:res-baseline}.

Of the figures just quoted, the reconstruction and re-simulation
errors measure agreement with the data, while the parameter errors
measure agreement with the exact values. Only the parameter agreement
is evidence that the identification worked; a good response fit is
necessary but not sufficient. A silent failure, defined in Section~\ref{sec:identifiability}, is a good fit with wrong parameters; Section~\ref{sec:res-ladder} shows one in which the record is fitted while the force-density error reaches $79.8$\,\%. The misfit cannot separate the two cases. The identifiability analysis of Section~\ref{sec:identifiability} is computed
from the geometry and the sensor set \emph{before} the record is
inverted, and it shows which members can be determined and to what precision; the recovered values are then compared with that prediction. The Cram\'er--Rao bounds of
Section~\ref{sec:res-baseline} give the precision attainable at the stated noise level. A recovery close to the bound is therefore at the limit of what the record allows; a recovery far above it is not, however small the misfit. Finally, the scatter across repeated runs from different training seeds, each a fresh network initialization and noise
realization (Section~\ref{sec:res-baseline}), distinguishes a parameter that the record determines from one at which the optimizer happened to stop. The predictions at unmeasured coordinates quoted above are reported as a consistency check only.

The record inverted in this section was generated by the same forward model that the residual uses. The accuracies above might therefore reflect only self-consistency, a circularity that the inverse-problems literature calls an inverse crime. Section~\ref{sec:res-crossmodel} addresses
it directly, with generators whose governing equations the residual cannot
represent at any parameter value. There, lumping the cable mass at the nodes violates the massless-chord cable law and roughly doubles the taut force-density error, and replacing the rigid bars with elastic ones removes the algebraic constraints and changes the integration scheme. Sections~\ref{sec:res-measured} to~\ref{sec:res-olsztyn} then use measured records only, and Section~\ref{sec:res-wangleg} runs the residual itself on measured records from hardware built by another group.

The inert base cable behaves as theory requires in both regimes.
Initialized at 10\,N/m, it drifts without converging, to $21.5$\,N/m in the taut record and toward zero in the slack-transition record, while every other parameter converges. Its recovered value is therefore not an error of the method; the record does not determine it.

This structure is unfavorable for a modal method. The natural frequencies of the X-frames shift by only about $12$\,\% over a 50-fold change in prestress level, which gives frequency-based updating little information. A full-scale tensegrity footbridge shows the same
insensitivity: its natural frequencies move by about $1$\,\% as its self-stress ratio quadruples \citep{BelHadjAli2010}, because its deformation modes are stiffened mainly by member extension. A structure measured in the same laboratory shows the opposite case. A
five-module active tensegrity whose lowest mode is a prestress-stiffened
mechanism shifts that frequency from about $2.6$ to $3.3$\,Hz across 11 measured prestress levels, while its member forces rise by a
factor of $1.7$ \citep{BelHadjAliSmith2010IJSS}. Although both structures were measured with the same instrument in the same laboratory, their frequencies change by $1$ and by $29$\,\%. The difference lies in whether the observed mode is stiffened geometrically or elastically, as the modal argument of Section~\ref{sec:identifiability} shows. The transient formulation avoids the problem: the force densities enter the equation-of-motion residual linearly, and they are recovered with errors near $1$\,\%. The slack-transition record loses force information during its slack intervals and still identifies the members that remain active, consistent with the information argument of
Section~\ref{sec:identifiability}.

\subsection{Identification across the benchmark structures}
\label{sec:res-family}

The physics-informed identification is applied to the staircase frames and the three-dimensional prism at 2\,\% noise (Table~\ref{tab:scaleup}).
Every single-point row, the X1 rows included, uses 12{,}000 steps; the repeated two-point runs use 25{,}000.
The demonstrations of Section~\ref{sec:res-1cell} ran for 25{,}000 steps, which explains why their errors are below the X1 rows here; Section~\ref{sec:res-treatments} isolates this effect of training length. The
held-out ratio is the ratio of the sensor-data loss on the held-out $30\,\%$ of time samples to that on the training $70\,\%$; a value near 1 means the fit generalizes. Sensors are placed at half the free coordinates on the single-point runs and the repeated X2 runs, and at two-thirds (8 of 12) on the repeated X3 runs. Errors cover the identifiable members only. The base
cable is inert, and excluded, only on the X1 frame
(Section~\ref{sec:identifiability}).

\begin{table}[!tbp]
  \centering
  \footnotesize
  \caption{Identification across the benchmark structures at 2\,\% noise.
  Errors are over the identifiable members; reconstruction is the
  root-mean-square error at unmeasured coordinates; the held-out ratio
  and the Outcome verdicts are defined in
  Section~\ref{sec:res-family}. 1-pt rows use single-point forcing and 2-pt rows are repeated runs with two-point forcing; step counts are given in
  the text.}
  \label{tab:scaleup}
  \setlength{\tabcolsep}{4.5pt}
  \begin{tabular}{llccccccl}
\toprule
Structure & Record & Forcing & Transitions & max $\hat\sigma$ err (\%) & $\hat c$ err (\%) & recon.\ (\%) & held-out ratio & Outcome \\
\midrule
X1 frame & taut & 1-pt & 0 & 3.5 & 1.3 & 4.1 & 2.8 & identified \\
X1 frame & slack-transition & 1-pt & 20 & 18.3 & 3.2 & 11.5 & 9.5 & partial \\
X2 frame & taut & 1-pt & 0 & 79.8 & 7.0 & 5.3 & 4.6 & silent failure \\
X2 frame & taut & 2-pt & 0 & 2.1 & 0.1 & 1.3 & 6.3 & identified \\
X2 frame & slack-transition & 1-pt & 20 & 20.8 & 5.5 & 1.7 & 6.0 & partial \\
X2 frame & slack-transition & 2-pt & 61 & 1.8 & 2.7 & 0.7 & 6.9 & identified \\
X3 frame & taut & 1-pt & 0 & 99.7 & 95.0 & $>$100 & 485.6 & fit failure \\
X3 frame & taut & 2-pt & 0 & $>$100 & 23.8 & 12.1 & 31.8 & fit failure \\
X3 frame & slack-transition & 1-pt & 18 & $>$100 & 30.0 & 36.7 & 92.6 & fit failure \\
X3 frame & slack-transition & 2-pt & 21 & 2.9 & 0.9 & 7.8 & 4.7 & identified \\
Triplex prism & taut & 1-pt & 0 & 0.6 & 3.2 & 0.8 & 2.2 & identified \\
Triplex prism & slack-transition & 1-pt & 14 & 2.0 & 0.5 & 1.0 & 4.4 & identified \\
\bottomrule
\end{tabular}

\end{table}

The three-dimensional prism is identified as accurately as the X1 frame. Its force densities, fixed uniquely by gravity and equilibrium, are recovered from four measured channels of nine free coordinates with errors of $0.6$\,\% on the taut record and $2.0$\,\% on the slack-transition record; the damping errors are $3.2$ and $0.5$\,\%. On the X2 frame, stronger excitation improves the identification. The low-amplitude taut record leaves weakly excited members essentially unidentified, with a maximum error of $80$\,\%, while the slack-transition record, with 20 taut--slack transitions, recovers them several times more accurately. Within the structures and
excitation suites tested, the information gained from the stronger
motion outweighed the information lost while individual cables were
slack, consistent with the Fisher analysis of
Section~\ref{sec:identifiability}. The method fails detectably on the X3 frame at this step count, with parameter errors
approaching or exceeding $100$\,\%.

The failures are of two kinds, and only one is flagged by the fit.
In a fit failure, the network cannot reproduce the measured motion at
all, and the held-out ratio flags it without the exact parameters.
Within the runs of this fixed-length, 2\,\%-noise grid with the smoothed cable law,
the held-out ratio stays below $7$ on every successful run and rises
to $93$ and $486$ on the X3 runs. The threshold depends on the protocol, because the baseline of the ratio follows the noise level: the successful 10\,\%-noise
inversion of Section~\ref{sec:res-robust} has a ratio of $20$. The
ratio flags overfitting of the instrumented channels; a failure that does not fit even those channels appears instead as a large reconstruction error at a low ratio. The two signals are therefore complementary.

The second kind is silent. On the X2 taut record, the motion is fitted well, although not to the noise floor: the reconstruction error is $5.3$\,\%, and the held-out ratio is $4.6$, inside the band of successful runs. The parameter errors still reach $80$\,\%. This case shows most clearly that prediction accuracy and parameter accuracy are different quantities: a model can reproduce the measured response closely while its force densities are badly wrong. For this reason, response fit is reported throughout as a consistency check and never as evidence of identification. The failure is reproducible: across six training seeds, the maximum force-density error is $81\pm1$\,\%. A Fisher analysis of
that record at the exact parameters, following Section~\ref{sec:identifiability}, finds no local lack of information.
The sensitivities of the seven members span only a factor of three, and
every relative Cram\'er--Rao bound is below $0.5$\,\%. The record therefore contains the information, but the finite-length optimization does not extract it: a distant parameter set reproduces the measured channels nearly as well. Neither the held-out ratio nor the local Fisher analysis detects that practical non-uniqueness.

Both kinds of failure respond to richer excitation, but differently on the two frames. On the X2 frame, two-point forcing, at two coordinates and two frequencies with the same sensors and noise, removes both single-point deficits. On the taut record, it removes the silent failure: the maximum error drops from $80$ to $12.4$\,\% in a run of the same 12{,}000 steps (Table~\ref{tab:ladder}), and to $2.1$\,\% after 25{,}000 steps, with a damping error of $0.1$\,\%. On the slack-transition record, it reduces the worst-member error from $21$ to $1.8$\,\%. On the X3 frame, richer excitation succeeds only with the slack-transition record, and only
with the sensor set widened from 6 to 8 of the 12 free coordinates. That record recovers all 10 cables to within
$2.9$\,\% and the damping to $0.9$\,\% through 21 taut--slack transitions at 25{,}000 steps; at 12{,}000 it has not yet converged. The low-amplitude taut record with the same two-point excitation remains a fit failure,
flagged again at a held-out ratio of $32$. At this training length, the richness of the excitation sets the failure boundary more than the size of the structure does, and strong excitation that produces taut--slack transitions is the most effective improvement.

The X3 frame needs a fuller explanation, because the method succeeds on the X1 and X2 frames but not on the X3 frame. The failure is not caused by missing information. The Fisher analysis of the X3 records with multi-point excitation gives worst relative Cram\'er--Rao bounds below $0.7$\,\%, which shows that the records determine the parameters. Nor is it removed by a larger network or longer training alone: a five-layer trunk of
width $160$ trained for $60{,}000$ steps on the gently forced taut
record still overfits the instrumented channels, with a held-out ratio
above $100$ and a median parameter error near $100$\,\%. Stronger excitation removes it, as the convergence with the slack-transition record above shows. At this scale, the boundary is therefore an optimization limit of gently forced records, which richer excitation moves and a larger network does not. The X2 taut record shows the same boundary at a smaller scale.

\subsection{Removal of a silent failure}
\label{sec:res-ladder}
The X2 taut record is used here to test which intervention removes a silent failure. Table~\ref{tab:ladder} starts from that record, with one-point forcing, four sensors, and a maximum force-density error of $79.8$\,\%. The training length is held at $12{,}000$ steps, and one factor is changed at a time. More sensors do not help: enlarging the sensor set from four of the eight free coordinates to all eight changes the force-density error only to $76.7$\,\%. The failure is therefore not caused by a lack of sensors. More data of the same kind do not help either:
three pooled taut records of the same one-point form give $90$\,\%.
Records that excite the same modes carry redundant information about
the non-unique direction.

Excitation that exercises more of the structure removes the failure. Two-point forcing at two frequencies away from the resonances reduces the error to $12.4$\,\%, a one-point slack-transition record reduces it to $20.8$\,\%, and the two together with the full sensor set reduce it to $9.2$\,\%. The damping coefficient, well excited throughout, is
recovered to a few percent by every intervention. A silent failure caused by practical non-uniqueness is therefore removed by richer excitation. More sensors and more copies of the same record do not remove it. This agrees with the
identifiability reading of Section~\ref{sec:identifiability}.

\begin{table}[!tbp]
  \centering
  \footnotesize
  \caption{Removal of the silent failure on the X2 taut record. One factor is
  changed at a time from the baseline record, at a fixed
  $12{,}000$-step run; the first error column is the maximum over
  the identifiable force densities and the second is the
  damping-coefficient error.}
  \label{tab:ladder}
  \begin{tabular}{lcc}
  \toprule
  Intervention & max $\hat\sigma$ err (\%) & $\hat c$ err (\%)\\
  \midrule
  Baseline (one-point taut, four sensors)     & $79.8$ & $7.0$\\
  $+$ sensors (all eight coordinates)         & $76.7$ & $1.9$\\
  $+$ two-point off-resonance forcing         & $12.4$ & $0.4$\\
  $+$ three pooled taut records               & $90.4$ & $1.8$\\
  One-point slack-transition record           & $20.8$ & $5.5$\\
  Combined (transitions, two-point, eight sensors) & $9.2$  & $3.3$\\
  \bottomrule
  \end{tabular}
\end{table}

\subsection{Comparison of transition treatments}
\label{sec:res-treatments}

Table~\ref{tab:treatments} compares the three transition treatments on the same slack-transition records under one protocol, with four seeds at each of two training lengths. This separates the effect of the treatment from that of the training length.

The training length has the stronger influence. Every treatment improves by more than an order of magnitude between $12{,}000$ and $25{,}000$ steps. The median error of the smoothed cable law falls from $18.8$ to $0.77$\,\%, and that of event-aware collocation from $17.9$ to $0.82$\,\%. At the shorter length, none of the three has converged on this record, and the errors there measure progress toward convergence.

Among the treatments, the smoothed cable law and event-aware collocation are indistinguishable. Their medians differ by less than a percentage
point at $12{,}000$ steps and by $0.05$ percentage points at $25{,}000$, and their seed ranges overlap almost completely. Concentrating collocation
points near the transitions therefore neither helps nor hurts once the smoothed cable law is in place. The complementarity treatment performs worst. Its median force-density error is $27.0$\,\% at $12{,}000$ steps and $2.55$\,\% at $25{,}000$, against $18.8$ and $0.77$\,\% for the smoothed cable law. Its damping error, $42.6$ and $24.4$\,\%, is several times that of the other two treatments. Its additional tension head adds unknowns that the record must determine, which is a likely reason for the slower convergence. The same ordering holds on
the X2 frame, where the complementarity run exceeds $100$\,\% while the smoothed cable law reaches $21.1$\,\%.

These results support the smoothed cable law, which matches the best alternative at lower cost and has no auxiliary unknowns. Training long enough to converge matters more than the choice of transition treatment. Both comparisons cover all identifiable cables;
on the X2 frame the largest error of the complementarity run is on the base cable, which is identifiable there. No treatment was tuned separately, and the comparison is reported as measured.

\begin{table}[!tbp]
  \centering
  \footnotesize
  \caption{Comparison of the transition treatments on identical slack-transition records, at both training lengths. Every entry is the median over
  four seeds of the same protocol.}
  \label{tab:treatments}
  \begin{tabular}{llccccc}
\toprule
Structure & Treatment & Steps & max $\hat\sigma$ err (\%) & $\hat c$ err (\%) & recon.\ (\%) & transitions \\
\midrule
X1 & smoothed cable law & 12{,}000 & 18.82 & 3.8 & 11.6 & 20 \\
X1 & Fischer--Burmeister NCP & 12{,}000 & 26.97 & 42.6 & 15.5 & 20 \\
X1 & event-aware collocation & 12{,}000 & 17.90 & 2.4 & 10.9 & 20 \\
\addlinespace[2pt]
X1 & smoothed cable law & 25{,}000 & 0.77 & 6.5 & 0.5 & 20 \\
X1 & Fischer--Burmeister NCP & 25{,}000 & 2.55 & 24.4 & 1.7 & 20 \\
X1 & event-aware collocation & 25{,}000 & 0.82 & 6.3 & 0.6 & 20 \\
\addlinespace[2pt]
X2 & smoothed cable law & 12{,}000 & 21.13 & 5.6 & 1.6 & 20 \\
X2 & Fischer--Burmeister NCP & 12{,}000 & 100.82 & 23.3 & 8.4 & 20 \\
\bottomrule
\end{tabular}

\end{table}

\subsection{Prestress series on the prism}
\label{sec:res-family-prism}

\begin{figure}[!tbp]
  \centering
  \includegraphics[width=0.52\textwidth]{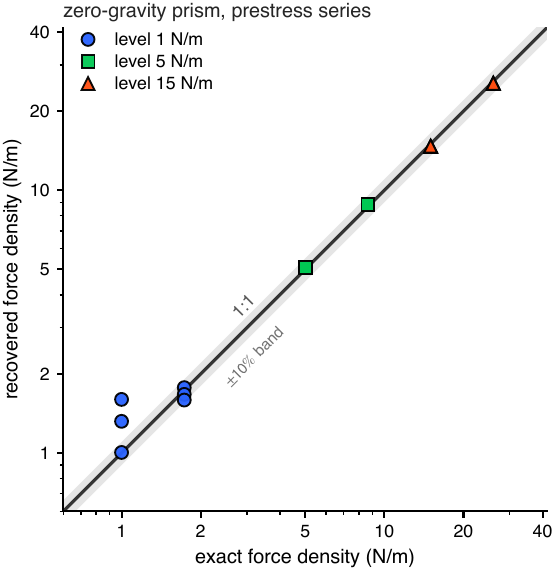}
  \caption{Recovered against exact force densities for the zero-gravity prism prestress series, structurally inert members excluded, with the
  $\pm 10$\,\% band shaded. Levels 5 and 15\,N/m lie on the 1:1 line; the two misses at level 1\,N/m are an optimization limit (Section~\ref{sec:res-family-prism}).}
  \label{fig:family}
\end{figure}

At its pure self-stress twist, the zero-gravity prism has a one-parameter series of prestress states, and the identification follows this series (Figure~\ref{fig:family}). At prestress levels $5$ and $15$\,N/m, the identifiable force densities are recovered to about $2$\,\%, including the ratio of $\sqrt{3}$ between the cross-cable and triangle-cable force densities. At the lowest level
($1$\,N/m), two of the members at the smallest force density are not
recovered, with errors of $32$ and $60$\,\%, while the rest stay inside the $10$\,\% band.

The miss is systematic. Six training seeds at the same $12{,}000$ steps miss the same two members in the same direction, with an average worst-member error of $66.1$\,\%; the run shown in Figure~\ref{fig:family} is the most favorable of the six.

The usual explanations at low prestress are a small taut margin and weak information. The taut margin is small enough that the record contains 45 taut--slack transitions. A Fisher analysis of the same excitation and sensing quantifies the loss of information: the worst relative Cram\'er--Rao bound grows 17-fold, from $0.047$\,\% at $15$\,N/m to $0.804$\,\% at $1$\,N/m, and the conditioning of the information matrix becomes worse.

Neither effect explains the size of the miss. The bound at level 1 is $0.80$\,\%; a method that attains the bound could therefore recover these members with errors well below $1$\,\%. The gap to the bound lies in the
optimization. Doubling the training length to $25{,}000$ steps halves
the error, to $31.8$\,\% on average over the same seeds. That is still about 40 times above the $0.80$\,\% bound at this level. The level-1 result is therefore an
optimization limit, not an identifiability limit. The information does decrease as the prestress approaches zero, and the prestress series quantifies this decrease. At level 1, however, the record still contains the information, and the inversion has not yet extracted it.

\subsection{Extension to a two-span bridge}
\label{sec:res-bridge}

\begin{figure*}[!tbp]
  \centering
  \includegraphics[width=0.92\textwidth]{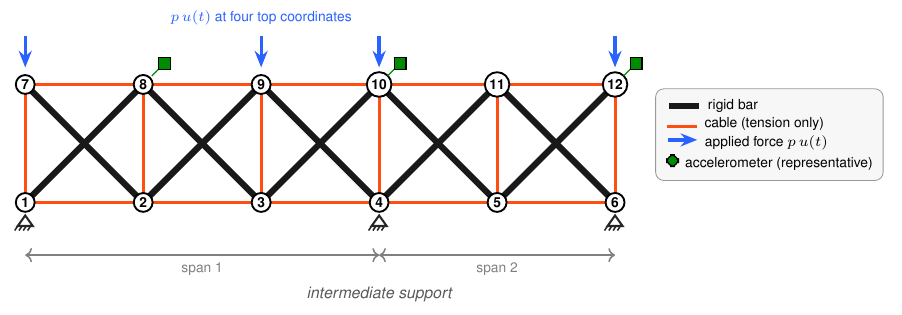}
  \caption{The X5 two-span tensegrity bridge, drawn from the simulator
  geometry, with the two end base nodes and one intermediate base node
  pinned. Blue arrows mark the four driven coordinates; the green
  squares sketch the accelerometer layout.}
  \label{fig:bridge}
\end{figure*}

The largest structure in the study is an X5 planar X-frame with 16 cables and eight mechanical degrees of freedom, against 10 cables and six for the X3 frame. Under gravity, the plain X5 frame is statically unstable, because its long unbraced span has two sway modes with negative stiffness. Pinning one intermediate base node removes the instability and
turns the frame into a two-span tensegrity bridge
(Figure~\ref{fig:bridge}).

The bridge is identified from a taut record and a slack-transition record. Each is forced at four top coordinates and instrumented at nine of the 18 free coordinates. Two of the nine are on the top-chord node
above the interior support. The two bars at that node connect it to pinned base nodes and hold it fixed, which is the nodal counterpart of the invisible cable of Section~\ref{sec:identifiability}. These two channels carry no signal, and seven channels remain active. The Fisher analysis on the seven active channels confirms that both records are informative. No null direction appears among the 16 force densities and the damping
coefficient, and the worst relative Cram\'er--Rao bounds are $0.6$\,\%
on the taut record and $0.2$\,\% on the slack-transition record.

Both central findings recur at this scale. The slack-transition record is the more informative: it recovers the damping to $0.9$\,\% and
the force densities to a median of $4.2$\,\%, with 9 of the 16 cables within $5$\,\%. The taut record fits the sensors equally well, with a held-out ratio of $2.7$ against $3.3$ for the slack-transition record, and the Fisher analysis shows that it is also informative. Even so, it recovers the force densities only to a median
of $16$\,\% and the damping to $5$\,\%. That is a silent failure of the same type as on the X2 frame, at more than twice the number of cables, and the held-out ratio again does not detect it.

The best result also shows the expected effect of size. The worst cable on the slack-transition record reaches $29$\,\%, against errors of a few percent on the smaller structures, most likely because the active sensors are spread across more unknowns. The method therefore extends to a bridge-scale structure, and both central findings hold there.

\subsection{Uncertainty and classical baselines}
\label{sec:res-baseline}

Eight independently seeded repetitions of the taut X1 inversion quantify the scatter of the identification, each with a different noise realization and
network initialization over 12{,}000 steps. The seed-to-seed standard
deviations are $0.085$, $0.081$, and $0.11$\,N/m for the identifiable force densities and $0.018$\,N\,s/m for the damping coefficient. All four are between $2.7$ and $8.3$ times the Cram\'er--Rao bounds of
Section~\ref{sec:identifiability} (Figure~\ref{fig:ident}c). Bias
dominates the force-density error: the seed means are $2$--$4$\,\% from the exact values, each bias roughly an order of magnitude above its own seed standard deviation. This bias is attributed to the finite training length and the smoothed cable law; the measurement noise changes from seed to seed and does not produce a common bias. The damping seed
mean is closer still, $0.7$\,\% from exact, though its scatter of
$1.8$\,\% exceeds its small bias. The held-out ratio averages $2.8$ across seeds, consistent with the
successful-run band.

The classical identification methods extend the comparison. Levenberg--Marquardt shooting is started from the same records, noise, and distant initial guess, with all force densities at $10$\,N/m and the damping at $0.3$\,N\,s/m. It gives different results on the two records. On the low-amplitude taut record, it satisfies its convergence
criteria at a spurious minimum: the identifiable force-density errors
are $61$--$97$\,\%, and the damping is overestimated 66-fold.
The failure is evident without the exact parameters, since the final misfit is orders of magnitude above the noise floor. On the slack-transition record, the same method recovers the identifiable force densities with errors below $0.1$\,\% and the damping within $0.2$\,\%, and the inert member goes to zero.

The computational cost shows the same split by record. Shooting performs one full forward simulation per residual evaluation, and its cost therefore depends on the path of the search. On the taut record, the search moved into high-damping regions where the equations become stiff, and 68 simulations took about three hours. On the slack-transition record, 232 simulations took 83 seconds. The physics-informed identification takes $2819$ and $4643$ seconds on the two X1 records in its $25{,}000$-step runs on the same hardware. On the slack-transition record, where shooting converges, shooting is both faster and more accurate than the physics-informed identification. On the taut record, the physics-informed identification is faster and more accurate. The advantage of the physics-informed identification is therefore its robustness from a distant initialization.

Repeated starts confirm that the taut failure depends on the starting point: of several distant restarts, at least one reaches the exact force densities to within $1$\,\% while others converge to spurious minima.

The two recursive filters show the opposite pattern.
The joint state--parameter unscented Kalman filter uses standard scaled
sigma points on the augmented state and three passes over the record.
It succeeds on the taut record, recovering the identifiable force
densities to $3.1$\,\% and the damping to $1.5$\,\% from the same
distant initialization. On the slack-transition record, its error grows to $183$\,\%, and it overestimates the damping. The likely cause is that the unscented transform propagates sigma points across the taut--slack boundary and averages states on opposite sides of it. A particle filter built around an ensemble Kalman filter represents the parameters by a weighted set of samples instead of one covariance. From the distant initialization, it recovers neither record; its errors are $62$\,\% on the taut record and above $400$\,\% on the slack-transition record. At $2$\,\% noise, the likelihood is so sharply peaked that one particle takes almost all the weight at every step. The effective sample size is then 1 out of 400, and the filter reduces to a greedy search. This may explain why recursive filters in the tensegrity literature are used on linearized models with informed priors \citep{Aswal2021}.

Two of the three classical methods succeed in opposite regimes, and the particle filter fails in both. Shooting converges to a spurious minimum on the taut record and is exact on the slack-transition record; the unscented filter does the reverse. Only the physics-informed identification succeeded on both records from the distant initialization, recovering the identifiable
force densities to $1.0$ and $1.3$\,\%. Classical updating with an
exact forward model remains strongly competitive when the optimization problem is well conditioned, as shooting shows on the slack-transition record. In that record, the taut--slack transitions appear to make the optimization problem easier to solve. The same transitions that cause the unscented filter to fail let an independent gradient-based method reach the exact parameters, two orders of magnitude faster than on the taut record. The benefit of the transitions is therefore not specific to the physics-informed identification.

Table~\ref{tab:methods} compares the four methods in three respects: recovery accuracy, running cost, and how many quantities the user
must choose before the method runs. A hyperparameter is any numeric or
structural choice that is not a property of the structure or the record
and not a convergence tolerance. The shared initialization of all force
densities at $10$\,N/m and damping at $0.3$\,N\,s/m is counted once for every method that uses it. The physics-informed identification has by far the most hyperparameters of the four. Shooting needs two choices and outperforms it on the slack-transition record in both accuracy and wall time.

\begin{table}[!tbp]
\centering
  \footnotesize
\caption{The four identification methods on the X1 frame, from the same records,
the same $2$\,\% noise and the same distant initialization. Accuracy
is the largest relative error over the three identifiable force
densities, with the damping error beside it. Cost is the wall time of
the run that produced the quoted accuracy, on the same hardware, taut
value first; hyperparameter counting is defined in the text.}
\label{tab:methods}
\begin{tabular}{@{}lccccrc@{}}
\toprule
 & \multicolumn{2}{c}{Taut record} & \multicolumn{2}{c}{Slack-transition record}
 & \multicolumn{2}{c}{Cost and tuning}\\
\cmidrule(lr){2-3}\cmidrule(lr){4-5}\cmidrule(lr){6-7}
Method & $\max\hat\sigma$ & $\hat c$ & $\max\hat\sigma$ & $\hat c$
          & wall & hyper-\\
 & err.\ (\%) & err.\ (\%) & err.\ (\%) & err.\ (\%) & (s) & parameters\\
\midrule
Physics-informed identification & $1.0$   & $0.3$  & $1.3$           & $6.2$          & $2819$ / $4643$ & $15$\\
Levenberg--Marquardt shooting & $97.4$  & $6504$ & $\mathbf{0.07}$ & $\mathbf{0.2}$ & $10\,700$ / $\mathbf{83}$ & $\mathbf{2}$\\
Unscented Kalman filter       & $3.1$   & $1.5$  & $182.7$         & $133.5$        & $63$ / $47$ & $11$\\
Particle filter with EnKF     & $61.7$  & $67.1$ & $488.1$         & $80.8$         & $762$ / $751$ & $6$\\
\bottomrule
\end{tabular}
\end{table}

The unscented filter has the lowest cost and is accurate on the taut record. For a structure known to stay taut, it is therefore the appropriate choice. Shooting is the least tuned and the most
accurate of all four on the record it can solve. The particle filter is less accurate and more costly in every comparison. The failure comes from the likelihood: at $2$\,\% noise it is too narrow for 400 particles to represent. The physics-informed identification is robust in both regimes, at the cost of 15 hyperparameters and of wall times one to two orders of magnitude above those of the filters.

Pooling records reduces the noise- and record-specific bias of a single inversion. Three taut records at
forcing amplitudes of $6$, $10$, and $14$\,N are inverted jointly by a
single network conditioned on the record descriptor, with shared
physical parameters. They recover the identifiable force densities to
$0.01$--$0.5$\,\% and the damping to $0.2$\,\%, roughly halving the best single-record force-density errors. Sharing parameters across inexpensive repeated tests is well suited to monitoring practice, where many short records are easier to obtain than one long
one.

\subsection{Robustness to noise, model-form error, and adverse sensing}
\label{sec:res-robust}\label{sec:res-crossmodel}\label{sec:res-harsh}
Three robustness studies on the X1 frame are summarized here; their tables
and full discussion are given in Section~S6 of the Supplementary Material.

Accuracy decreases moderately with noise: a fivefold increase in noise
increases the force-density error by a factor of $1.8$, and removing one of
the two measured channels increases it by less than that. Identifying the
axial stiffness together with the force densities shows the trade-off that
the Fisher analysis predicts. On the taut record, the stiffness is recovered
to $7.4$\,\%, but the force-density errors grow to $35$\,\%; on the
slack-transition record, the joint identification fails, and the held-out
ratio flags the failure.

The simulated records above are generated and inverted with the same forward
model, which is an inverse crime (Section~\ref{sec:res-1cell}). Records from
generators that the residual cannot represent, with the cable mass lumped at
the nodes or with elastic bars integrated by a different solver, at most
double the taut force-density error of the X1 frame, to $7.0$\,\%. A
$5$\,\% perturbation of the mass matrix raises the error on the prism from
$0.6$ to $3.5$\,\%. On the slack-transition record, every generator stays
within the seed scatter of the matched case.

Under adverse sensing, the force-density error grows from $2.8$\,\% in the
baseline case to between $4.7$ and $12.2$\,\% for a single channel, $8$\,\%
noise, or a quarter of the samples in intermittent blocks, and to
$25.0$\,\% when the harmonic forcing is replaced by an impulse. All four
conditions together give no larger error than the impulse alone. On a single
adverse record, the excitation therefore limits the accuracy more than the
quality of the measurement does.

\subsection{Validation on measured data}
\label{sec:res-measured}
No cable slackens in the measured experiments of this and the next two sections. These experiments validate the rebuilt models, the small-oscillation identification, and, on the leg of Section~\ref{sec:res-wangleg}, the time-domain residual on measured records. The treatment of taut--slack transitions remains validated on simulated records only.

A published experiment provides measured frequencies against which the formulation is tested. The public dataset accompanying
\citet{Prism2025JSV} provides full-field motion-capture displacement
histories of all 12 nodes of a hexagonal tensegrity prism (six bars, 18 cables, modeled free--free). It also provides a
driving-point frequency response with four identified peaks at 7.285,
9.688, 10.430, and 11.396\,Hz, and the complete parametric model. The
prism was rebuilt here from the published parameters: rigid bars of
measured mass 42.3\,g, cable stiffness $EA = 379$\,N, and the design
prestress of 10\,N in the two ring-cable groups. The structure carries a single self-stress mode,
and scaling it to the ring value reproduces the published
vertical-cable tensions of 11.88\,N.

The rebuilt model is verified in the modal domain. All eight published
theoretical frequencies (5.39--13.16\,Hz) are reproduced to within
1.0--2.0\,\%, including both degenerate pairs. The small uniform
overprediction is consistent with the neglected string mass (2\,\% of
the total). Matching these frequencies requires the transverse
geometric stiffness of the compressed bars in the linearization: a
cables-only stiffness assembly overestimates this structure's
mechanism-dominated frequencies by tens of percent. The classical-limit check of Section~\ref{sec:formulation} was extended to include this term.

\begin{figure}[!tbp]
  \centering
  \includegraphics[width=0.82\textwidth]{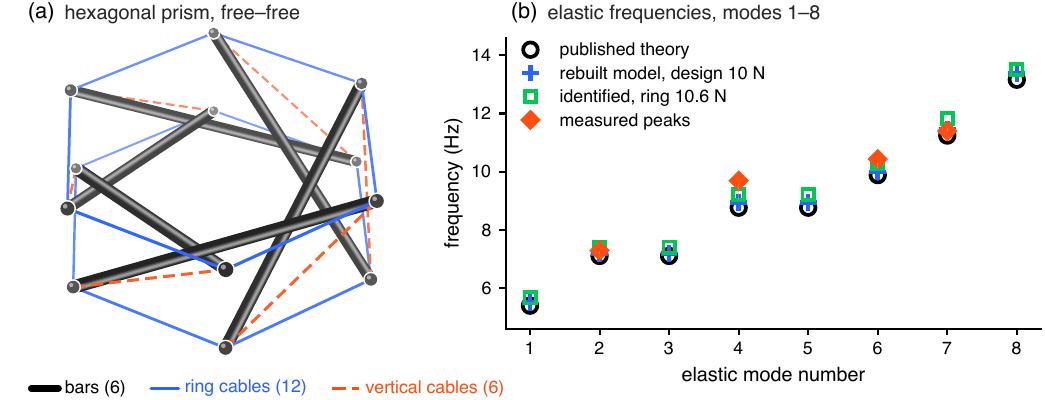}
  \caption{Validation on the hexagonal-prism test of
  \citet{Prism2025JSV}. (a)~The rebuilt model of the structure, from the published parameters. (b)~Elastic frequencies: the published theoretical values and those of the rebuilt model agree to about $2\,\%$
  across all eight modes, and the four measured peaks are best
  explained by a ring tension of 10.6\,N
  (Section~\ref{sec:res-measured}).}
  \label{fig:prismhex}
\end{figure}

Identification proceeds in the modal domain
(Figure~\ref{fig:prismhex}b). A single prestress scale fitted to the
four measured peaks returns a ring tension of 10.6\,N, 6\,\% above the
design value, with a root-mean-square frequency residual of 0.33\,Hz.
The peaks are matched to the model's distinct frequency lines by order;
the dataset's mode shapes are not reconstructed here. One peak is questionable: at 9.688\,Hz it is nearer the mode-6 line than the
mode-4 pair it is assigned to. Even so, the order-preserving
assignment fits all four peaks with the $0.33$\,Hz residual, while pairing 9.688\,Hz with mode~6 gives a residual of $0.92$\,Hz. That alternative still
returns a ring tension of 8.6\,N, near the design value. The conclusion therefore does not depend on the assignment: the prestress can be identified from the frequencies near the design value, while the frequencies are nearly insensitive to $EA$.

No single prestress level makes all four residuals zero: the
measured-to-model frequency ratios at the design prestress span
0.99--1.08. The source publication's own theoretical values show the same inconsistent offsets from its measurements, which points to unmodeled effects such as joint masses and distributed string inertia. Adding the cable stiffness as a
second parameter reduces the residual by only 9\,\% while increasing $EA$ by an implausible factor of 18. The mechanism-dominated modes
the test excites are sensitive to prestress and nearly insensitive to $EA$. This is an experimental example of the flat directions predicted by the Fisher analysis of Section~\ref{sec:identifiability}.

The dataset also distributes the raw motion-capture records, and this
study examined them to decide whether a time-domain inversion was
possible. Three dwell records are provided, each at a single excitation
frequency, sampled at $90$\,Hz over $61$ to $163$\,s. They are not suitable for a time-domain inversion. No force channel is recorded, and the drive point is not documented. A residual inversion would therefore have the amplitude, the phase, and the position of the load as further unknowns. The motion is also too small to slacken a cable. The largest change
in any inter-node distance across the three records is $7.6$\,mm, against
$14.4$\,mm of prestretch on the shortest cable at the design prestress. This comparison does not require the markers to be registered to the model. In addition, forming accelerations by differentiating the displacements twice reduces the signal-to-noise ratio in the excitation band from between $8.7$ and $42$ to between $0.19$ and $2.4$, with only $8.6$ to $12.4$ samples per cycle; the ratio is below 1 in two of the three records. Narrowband filtering about the drive frequency restores the ratio, but it also linearizes the record. The modal comparison above is therefore the comparison that this dataset supports.

\subsection{Identification against an independently measured spring rate}
\label{sec:res-wangleg}
The prism test of Section~\ref{sec:res-measured} compares an identified prestress with a design value. A second published experiment provides a measured reference, and its raw records allow the time-domain residual of Section~\ref{sec:pinn}
to be run on measured data, which no other experiment in this study
permits.

\citet{Wang2026Leg} report a planar biomimetic leg: a fixed pelvis, a
femur on a cable-constrained hip that acts as a revolute joint, and a
tibia carried on an antiparallelogram knee, with compliance from four
extension springs. Five spring configurations are tested. Every
quantity used here is published: node coordinates and body properties,
spring rates measured on a dedicated spring rig, and frequencies
identified from chirp tests. The 18 raw chirp records behind those tables were also supplied by the authors, who granted their use
for this study with citation and acknowledgment.

The rebuilt model is checked before it is inverted. Built from the published tables alone, with no fitted constant, it reproduces the authors' own
computed frequencies to $0.015$\,\% root-mean-square over 10 values,
and it reproduces the eight frequencies of their gravity-ablation
table, a physical regime not used to build the model, to
$0.094$\,\%. Against their measurements, the same model differs by $3.2$\,\%, which this study cannot reduce, because the authors' own computation differs by $3.19$\,\% from the same measurements.

The knee spring rate is the identification target. This choice is made before any inversion, by the identifiability analysis of Section~\ref{sec:identifiability} applied to the springless Case~E. Gravity supplies most of the restoring stiffness of the first mode: the hip springs contribute only $17$ to $23$\,\% of it, while the knee springs contribute $56$ to $70$\,\% of the restoring stiffness of the second mode. A frequency error enters an identified spring rate amplified by $2/s$, with $s$ the share of the restoring stiffness that the spring supplies. The amplification is about 3 at the knee and about 10 at the hip. At the hip, the source publication's own computed-to-measured frequency gap of $3.25$ to $4.89$\,\% already corresponds to spring-rate errors of $37$ to $47$\,\%. The hip rate therefore cannot be identified reliably from this experiment. It is included below only as a nuisance parameter, with its statistics given in Section~S4, and the knee provides the test.

\begin{figure*}[!tbp]
  \centering
  \includegraphics[width=\textwidth]{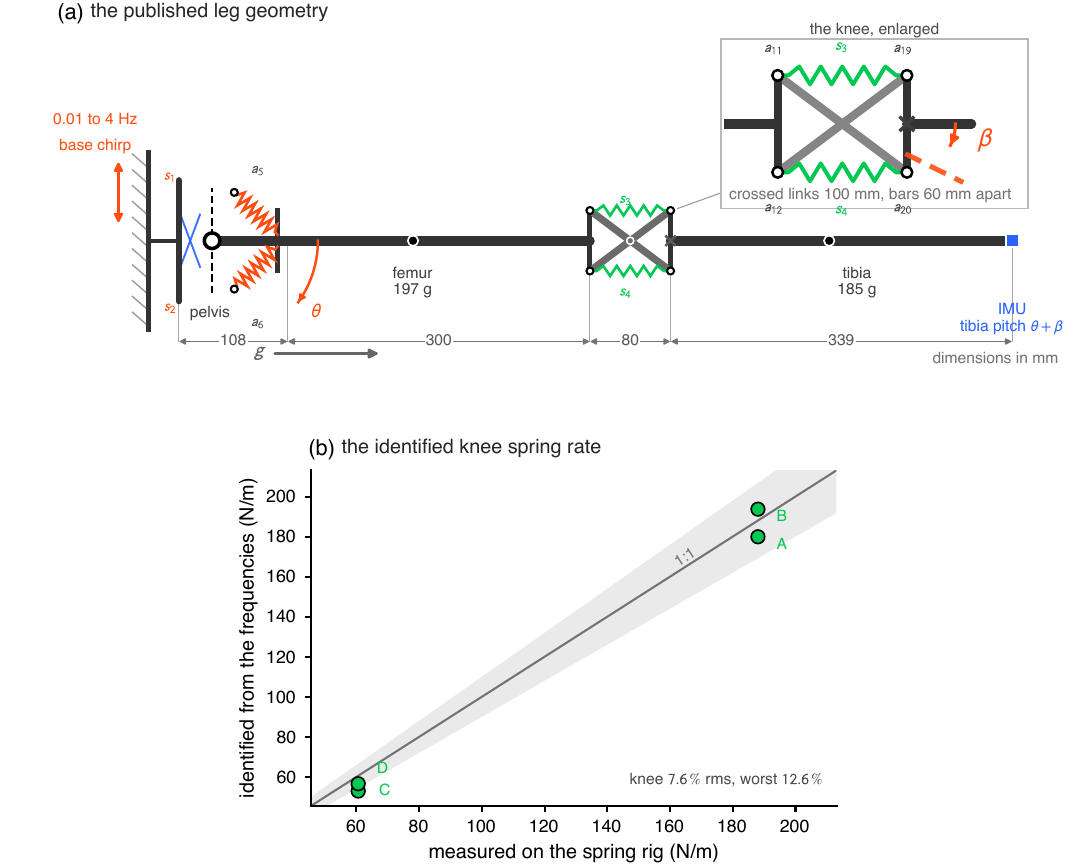}
  \caption{Identification on the biomimetic leg of \citet{Wang2026Leg},
  from their published tables and records: \textbf{(a)}~the leg drawn
  to scale from their published dimensions, with the antiparallelogram
  knee enlarged; \textbf{(b)}~the knee spring identified from each
  case's measured frequency pair against the rig-measured rate, with
  the $\pm 10$\,\% band shaded.}
  \label{fig:wangleg}
\end{figure*}

When identified from the measured frequency pair of each case, the knee scale agrees with the rig-measured rate to $7.6$\,\% root-mean-square, with a worst case of $12.6$\,\% (Figure~\ref{fig:wangleg}b, Table~\ref{tab:wangleg}). The rig rate is measured independently of this identification. When identified record by record instead of from the published pairs, the knee scale is within $8$\,\% in three of the four cases,
with record-to-record spreads of $4$ to $6$ percentage points.

\begin{table}[!tbp]
  \centering\footnotesize
  \caption{The knee spring identified from the published measured
  frequencies of \citet{Wang2026Leg}, against the rate measured
  independently on their spring rig; a scale of $1.000$ reproduces the
  rig value exactly. The identification target is selected in advance
  by the identifiability screening described in the text; Case~E
  carries no springs and supplies that screening instead.}
  \label{tab:wangleg}
  \begin{tabular}{@{}lccc@{}}
  \toprule
  case & rig rate (N/m) & identified scale & error (\%)\\
  \midrule
  A & $188.0$ & $0.957$ & $-4.3$\\
  B & $188.0$ & $1.030$ & $+3.0$\\
  C & $60.6$  & $0.874$ & $-12.6$\\
  D & $60.6$  & $0.934$ & $-6.6$\\
  \midrule
  \multicolumn{3}{@{}l}{root-mean-square error} & $7.6$\\
  \bottomrule
  \end{tabular}
\end{table}

The records also allow the time-domain residual of
Section~\ref{sec:pinn} to be run on measured data: the drive is known, the response is measured, and the stiffness and damping are unknown. The
fit is taken on the smoothed response envelope rather than point by
point, because a small frequency error accumulates into a large phase
error across a resonance passage. Six unknowns are identified per
record: two stiffness scales, two joint dampers, the chirp start, and
one drive amplitude; the hip scale is among them only so that it is set by the record. The procedure is first tested on a simulated record with known parameters and $2$\,\% noise, where it recovers all six unknowns to within $5$\,\% at an
envelope error of $0.0026$ (details in Section~S4).

On the measured records, the knee is identified to about the precision of the rig measurement. Pooled over the 15 records with springs, the knee scale is $1.001$, with a standard deviation of $0.183$. The mean agrees with the rig-measured rate to $0.1$\,\%, and no case departs by more than $8$\,\%. The nuisance parameter behaves as the screening predicts; Section~S4 reports it in full. The envelope errors range from $0.045$ to $0.093$, against $0.0026$ on the simulated test, which shows that model form limits these fits.

The leg is not self-stressed, and its springs never slacken. It therefore tests the small-oscillation formulation and the time-domain inverse against a measured reference, but not the treatment of taut--slack transitions. The experiment shows that the method can recover an identifiable member property from measured records, to $0.1$\,\% in the mean and $8$\,\% at worst, with the identification target selected by the screening before any inversion.

\subsection{Prestress identification against measured cable forces}
\label{sec:res-olsztyn}
The prism test of Section~\ref{sec:res-measured} compares an identified prestress with a design value. A second experiment allows a stronger comparison. The structure is self-stressed, its prestress is measured directly by force transducers built into the cables, and it is tested at 13 independent prestress levels.

\begin{figure*}[!tbp]
  \centering
  \includegraphics[width=\textwidth]{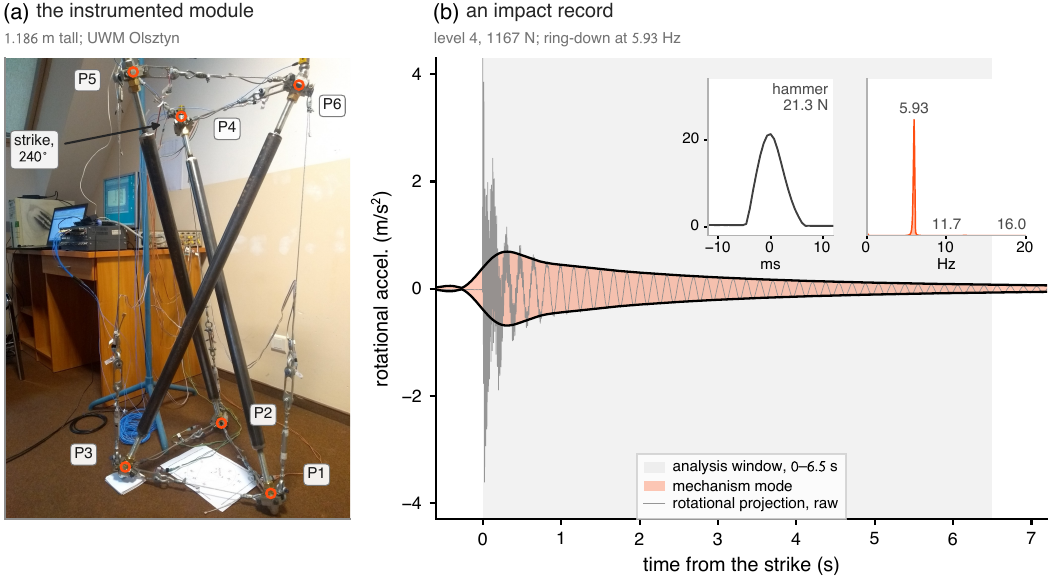}
  \caption{The tensegrity simplex of \citet{Malyszko2020Simplex}.
  \textbf{(a)}~The instrumented module, photographed by L.~Ma{\l}yszko and A.~Rutkiewicz,
  with the nodes marked: the struts P1--P4, P2--P5 and P3--P6 on their
  threaded-rod ends, the horizontal cables of the two triangles, the cross
  cables with their turnbuckles, a force transducer in each of the nine
  cables, triaxial accelerometers at the three free nodes, and the hammer
  strike at node~4, azimuth $240^\circ$.
  \textbf{(b)}~A single impact record, at level 4: the rotational projection, the envelope
  of the mechanism mode, and the time window used to measure the frequency.}
  \label{fig:olzexp}
\end{figure*}

\citet{Malyszko2020Simplex} report on a full-scale steel tensegrity simplex that is \SI{1.186}{\metre} tall and consists of three struts and nine cables, with the lower triangle fixed and the upper triangle free (Figure~\ref{fig:olzexp}). A force transducer was built into each of the nine cables; the transducers were individually calibrated and re-zeroed at the beginning of each prestress level by slackening the turnbuckles until the module carried no internal force. To obtain the dynamic response, the structure was struck with an instrumented hammer, and the response was recorded by three triaxial accelerometers mounted at the free nodes. The experimental data used in this study were collected by co-authors Ma{\l}yszko and Rutkiewicz. The dataset comprises one strike at each of the 13 prestress levels, all at the same node and azimuth, so that the prestress level is the only variable across the set.

Prior to inversion, the rebuilt model is verified against the results of the source publication. Table~3 of \citet{Malyszko2020Simplex} reports the base-cable force twice, once as measured and once as calculated from its Equation~(5). The calculated column is an equilibrium statement about the published geometry, which this study can reproduce independently. Evaluating the single self-stress mode of
the simplex on the published coordinates gives a base-to-cross force ratio of $0.2091$. The calculated column gives $0.2079$: this is the mean, over the 13 levels, of the ratio of the calculated base-cable force to the measured cross-cable force, and the per-level values range from $0.2070$ to $0.2091$. The two ratios differ by $0.6$\,\%. The two formulations give the same static equilibrium before any measured record is used.

\begin{figure*}[!tbp]
  \centering
  \includegraphics[width=\textwidth]{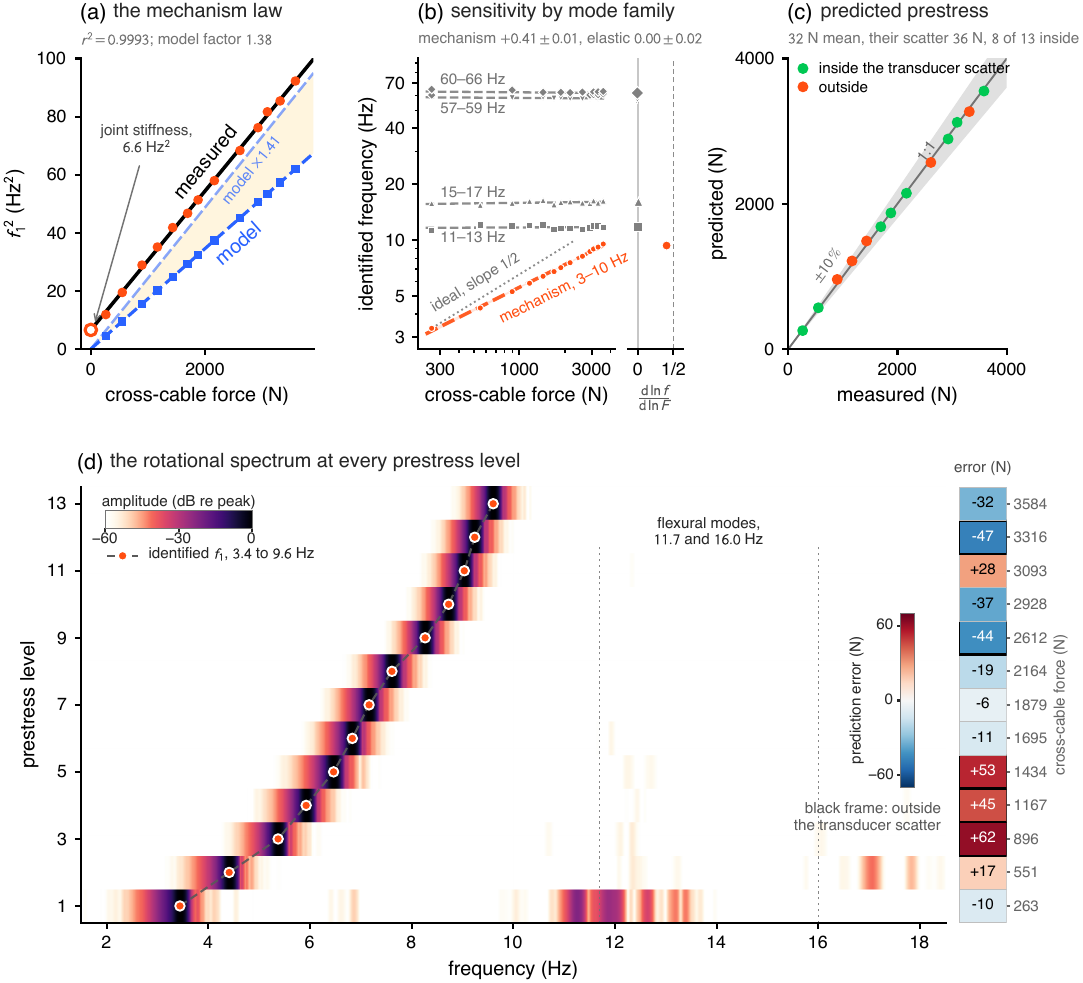}
  \caption{Identification on the simplex of Figure~\ref{fig:olzexp},
  from 13 impact records at 13 prestress levels:
  \textbf{(a)}~the mechanism law: the measurement, the rebuilt model at
  the transducer reading, and the same model at the prestress obtained from
  static equilibrium in the 2020 publication, the shaded band between the
  two being that factor and the open marker the joint-supplied intercept;
  \textbf{(b)}~frequency
  against force by mode family, log--log, the mechanism family in
  color with its $95$\,\% confidence band and the dotted ideal slope
  of $1/2$ anchored at the first point; the side strip shows each
  family's exponent $\mathrm{d}\ln f/\mathrm{d}\ln F$, with its
  $95$\,\% error bar, against the zero and one-half references;
  \textbf{(c)}~prestress predicted against the transducer reading, the
  joint stiffness alone fitted leave-one-out ($\pm 10$\,\% band);
  \textbf{(d)}~the rotational spectrum at every level, each record
  normalized to its own peak, with the per-level prediction error as
  the color-graded column on the right, framed in black where it
  exceeds the transducer scatter.}
  \label{fig:olz}
\end{figure*}

The rotational mode is isolated by projecting the three nodal accelerations
onto the tangential directions of the upper triangle, that is, onto its rigid-body rotation about the vertical axis. This projection removes the flexural modes by construction and does not rely on their separation in frequency. This matters at the lowest prestress levels, where the rotational mode falls to \SI{3.4}{\hertz}. With this projection (Figure~\ref{fig:olzexp}b, Figure~\ref{fig:olz}d), the 13 records give the first frequency within $1.19$\,\% root-mean-square of the values published in Table~4 of the 2020 study, with a worst case of $2.47$\,\%. This study therefore extracts the same frequencies from the records as the 2020 study, and any later disagreement concerns the model.

The identifiability argument presented in Section~\ref{sec:identifiability} can now be tested directly. Eq.~\eqref{eq:geometric-fraction} indicates that for a mode carrying no elastic strain energy, the squared frequency is proportional to the prestress, which yields a log-slope of $\mathrm{d}\ln f/\mathrm{d}\ln F = 1/2$; for an elastically stiffened mode, this log-slope approaches zero. The 2020 study identified every mode it could resolve at every level. Grouping these modes by family gives exponents of $+0.41$ for the rotational mechanism mode and $+0.00$ and $+0.01$ for the two flexural families below
\SI{20}{\hertz} (Figure~\ref{fig:olz}b). Under the same excitation and the same 14-fold change of force, one mode of the structure is more than 100 times as sensitive to prestress as another. The rotational family near \SI{62}{\hertz} is also flat, at $-0.01$. It turns the upper triangle but also stretches members, and Eq.~\eqref{eq:geometric-fraction} classifies a mode by the member stretching it produces.
A second strike series at azimuth $330^\circ$, recorded by the same authors at all 13 levels and processed directly in this study, gives the same result. It reproduces the first frequency of the first series to $0.5$\,\% on average, and its flexural peaks have exponents of $+0.003$ and $+0.014$, against $+0.39$ for the mechanism mode.

The mechanism law itself follows directly from the records, without a model. Fitting the squared first frequency against the transducer force across the 13 levels gives
\begin{equation}
  f_1^2 \;=\; 0.0239\,F \;+\; 6.56 ,
  \qquad r^2 = 0.9993 ,
  \label{eq:olzlaw}
\end{equation}
with a residual scatter of $1.2$\,\% of the mean. Figure~\ref{fig:olz}a compares this line with the rebuilt model, and the two differ in two separate ways. First, the model line passes through the origin, as it must for an ideal pin-jointed simplex, while the measurement keeps a rotational frequency of \SI{2.56}{\hertz} at zero prestress. This frequency comes from a stiffness that the joints supply and the idealization omits. The Discussion of the 2020 publication states the same expectation, that the first frequency should be zero without prestress. The discrepancy therefore stems from the shared idealization failing to capture the physical hardware, not from a conflict between the two numerical models. The intercept also keeps the sensitivity in Figure~\ref{fig:olz}b below the ideal value. The fitted mechanism exponent is $+0.405 \pm 0.012$ at the $95$\,\% level, significantly below the $1/2$ of Eq.~\eqref{eq:geometric-fraction}. Eq.~\eqref{eq:olzlaw} explains this shortfall: with $A = 0.0239$ and $B = 6.56$ its two constants, the local slope $\mathrm{d}\ln f/\mathrm{d}\ln F = AF/2(AF+B)$ rises from $0.245$ at level 1 to $0.464$ at level 13.

Second, the slopes differ. This is the more informative of the two differences, and the source publication supplies what is needed to explain it. In Table~2 of \citet{Malyszko2020Simplex}, the cable mass is reported as \SI{0.516}{\kilo\gram\per\metre} and categorized as non-physical: it is a reduced value for a consistent mass matrix, and it already includes the transducers and turnbuckles, distributed along each cable. The authors of the 2020 study confirm how it was formed: the weighed total
less six joint masses of \SI{0.897}{\kilo\gram} and three struts at the
published linear density, spread over the total cable length. Taken at face value, this mass determines the share of the cable mass that moves, with nothing to fit. The three lower horizontal cables join fixed nodes and contribute nothing. The three cross cables run from a fixed end to a free one, and a consistent mass matrix puts a third of each at the free end. The upper horizontal cables rotate with the top triangle, but along a chord whose mean squared radius is exactly half that of the nodes; half of their mass therefore counts when referred to a node. With the \SI{25}{\gram} accelerometer mass stated in the 2020 publication, this gives a calculated mass of \SI{0.342}{\kilo\gram} at each free node.

The convention is no longer the only option, because the authors of the 2020 study have since weighed every element the module contains: rod, sensor, nuts, pads,
plate, thimbles, clamps and shackles, together with the stated composition
order of every cable. Only the turnbuckles, which were lost after the experiment, were never weighed directly. However, the difference between the weighed total of \SI{16.205}{\kilo\gram} and the remaining components gives \SI{234}{\gram} for each turnbuckle, compared with the \SI{260}{\gram} recalled by the authors of the 2020 study.

The struts need one further statement, because their mass is not uniform
along the rod. The published \SI{1.990}{\kilo\gram\per\metre} is the linear density of the tube and of the threaded end sections alike. Each weighed strut is \SI{224}{\gram} heavier than this density implies, and the surplus is carried at the two welded connections between the tube and the end sections. The threaded sections are \SI{16}{\centi\metre} long; each connection is therefore \SI{13.75}{\centi\metre} from its node and carries \SI{112}{\gram}, and the model places these masses at those points.
Placing every weighed part where the photographs and the composition orders
put it gives \SI{1.166}{\kilo\gram} moving with each free node.

With the weighed masses, the model requires the prestress acting in the structure to be $1.384$ times the value that the cross-cable transducers report. With the distributed-mass convention, it requires $1.373$. The two mass models, built from different information, agree to $0.8$\,\%. Table~3 of the 2020 publication
gives $1.410 \pm 0.027$ for the same quantity, from two columns that need no
model at all: the measured horizontal-cable force against the value its
Equation~(5) requires. The dynamic and static estimates differ by
$1.9$\,\%, which is one standard deviation of that measurement's
scatter. The two estimates are independent, and neither was fitted to the other. They share only the measured force. The remaining difference therefore concerns the transducer calibration or the force distribution. Figure~\ref{fig:olz}a shows that comparison directly: evaluated at the prestress obtained from static
equilibrium, rather than at the transducer reading, the model line lies just below the measured line across the whole range, separated by $6.6$\,Hz$^2$ at zero
force and $4.9$\,Hz$^2$ at the highest level. This remaining separation is the joint stiffness that the pin-jointed idealization omits, and it is the only constant that the identification fits.

The identification is then compared with the transducers. The slope comes from the rebuilt model
carrying the hardware mass just described, with nothing fitted. The force
factor comes from Table~3 of \citet{Malyszko2020Simplex} by static
equilibrium alone.
Only the joint stiffness is taken from the frequencies, because a pin-jointed model cannot supply it; it is fitted leave-one-out, so that no level informs its own prediction, and all other quantities come from the model (Table~\ref{tab:ident}, Figure~\ref{fig:olz}c,\,d).

The mean absolute error is \SI{32}{\newton}, which is inside the \SI{36}{\newton} transducer scatter reported in the 2020 publication. The worst error is \SI{62}{\newton}, and the root-mean-square error is $2.9$\,\%. For comparison, fitting both constants to the frequencies, without using the model, gives \SI{28}{\newton} and $6.5$\,\%. This is better in newtons but has twice the relative error, because the free fit misses the lowest level by $22$\,\%, against $4$\,\% for the model-based prediction. Fitting no constant, which leaves the structure without joint stiffness, gives \SI{214}{\newton} and
$29$\,\%.

The fitted constant has a physical meaning: it is the rotational stiffness of the joints. Without it, the model line passes through the origin, and the 2020 publication states in its Discussion that an unprestressed simplex should have no first frequency at all. The force factor is taken from the source publication, not derived here. The comparison therefore inherits whatever lies behind it, including the effect on the force distribution of the \SI{32.5}{\milli\metre} eccentricity between the strut axis and the cross-cable attachment, which this study does not resolve. Finally, the excitation is a light impact about a taut equilibrium. No cable slackens, and this experiment does not test the treatment of taut--slack transitions (Section~\ref{sec:switch}).

The records also contain the damping of the rotational mode, which the
source publication does not report. This study examined it but does not report values. The log-decrement value depends on the chosen window, because the decay rate drifts by about a quarter within a single ring-down. The values are also between $0.56$ and $0.82$ times those obtained in the 2020 study from the same experiment, a disagreement larger than any effect that could be claimed. The limitation of the viscous damping law
is argued below from modal strain energy, which does not depend on
these estimates.

\begin{table}[!tbp]
  \centering\footnotesize
  \caption{Prestress identified from the vibration record at each of the 13 levels, against the force the cable transducers measured there.
  Bold marks the levels predicted inside their own transducer uncertainty.}
  \label{tab:ident}
  \begin{tabular}{@{}rrrrrr@{}}
\toprule
level & measured (N) & $\pm$ & predicted (N) & error (N) & error (\%)\\
\midrule
1 & 263 & 27 & 253 & $\mathbf{-10}$ & $-3.76$\\
2 & 551 & 37 & 568 & $\mathbf{+17}$ & $+3.02$\\
3 & 896 & 40 & 958 & $+62$ & $+6.89$\\
4 & 1167 & 38 & 1212 & $+45$ & $+3.86$\\
5 & 1434 & 35 & 1487 & $+53$ & $+3.72$\\
6 & 1695 & 31 & 1684 & $\mathbf{-11}$ & $-0.66$\\
7 & 1879 & 29 & 1873 & $\mathbf{-6}$ & $-0.30$\\
8 & 2164 & 28 & 2145 & $\mathbf{-19}$ & $-0.88$\\
9 & 2612 & 35 & 2568 & $-44$ & $-1.67$\\
10 & 2928 & 42 & 2891 & $\mathbf{-37}$ & $-1.25$\\
11 & 3093 & 43 & 3121 & $\mathbf{+28}$ & $+0.92$\\
12 & 3316 & 45 & 3269 & $-47$ & $-1.42$\\
13 & 3584 & 40 & 3552 & $\mathbf{-32}$ & $-0.90$\\
\bottomrule
\end{tabular}
\end{table}

\subsubsection*{An identifiability limit on damping}
The same structure shows a limit of the damping model, which follows from the geometry alone. Damping enters the formulation of
Section~\ref{sec:pinn} as an axial viscous coefficient in each cable. Its effect on a mode therefore scales with how much that mode extends the cables. Per unit
of modal kinetic energy, the rotational mode of this structure changes the
cable lengths about $2900$ times less than the modes above it. It is an
infinitesimal mechanism, stiffened by prestress rather than by extension, and
axial damping is nearly insensitive to it. A single cable coefficient fitted to a damping ratio of a few percent in the rotational mode would predict damping ratios of order $10^{6}$ in the modes above it. A structure whose modal damping is of the same order across these modes therefore cannot be represented by axial dashpots with any coefficient. Such damping points instead to hysteretic or joint-friction dissipation.

This is a limitation of the damping model, not of the identification, and it supports the identifiability argument of
Section~\ref{sec:identifiability}. The property that makes the
prestress-carrying mode informative about prestress, namely its lack of
elastic strain energy, is the same property that makes it uninformative
about any extension-based damping. Prestress and damping are recoverable
from different parts of the same measurement, and should not be quoted with
a shared confidence. On records simulated with the axial law, the coefficient is recovered correctly; only a real structure reveals the limitation.

In addition, the simplex identification reads one frequency from each record and does not run the time-domain residual of Section~\ref{sec:pinn}. It therefore tests the small-oscillation identification against a measured reference; the residual itself is tested on measured records with the leg of Section~\ref{sec:res-wangleg}.

\subsection{Stability, strength, and energy of the identified structure}
\label{sec:res-assessment}
The assessment turns a set of force densities into a stability margin
and member utilizations. With the exact force densities, it classifies every free-standing polyhedral tensegrity among the classic tensegrity forms as super-stable at its own self-stress, and the planar benchmark frames as merely stable under gravity. Under excitation, it tracks the instantaneous stability class of the prism through the same taut--slack transitions that the identification uses. The full demonstration, including the member utilization curves, the energy balance through taut--slack transitions, and the
bar-stiffness sensitivity study, is given in Section~S5 of the
Supplementary Material.

This section asks how the assessment responds to identification error. Table~\ref{tab:assessment} answers this by substituting the identified force densities for the exact ones. The
outcome follows the identifiability structure of
Section~\ref{sec:identifiability}. The large force-density errors concentrate in the directions that the record cannot determine. These directions, such as the cable joining the two pinned nodes, carry no free-coordinate stiffness, and they therefore hardly change the stability margin. On the X1 frame, an $87$\,\% maximum force-density error leaves the
margin off by $0.04$\,\%. The identified model therefore predicts stability well even where individual force densities are poorly recovered. The strength utilizations scale with the governing member force and follow the error on that member, which lies in the well-identified directions.

\begin{table}[!tbp]
\centering
  \footnotesize
\caption{Propagation of identification error to the structural assessment.}
\label{tab:assessment}
\begin{tabular}{lccc}
\toprule
Recovered model & max.\ $\sigma$ error & margin error & $U_c$ error\\
\midrule
X1, taut          & $14$\,\% & $0.12$\,\% & $0.3$\,\%\\
X1, slack-transition & $87$\,\% & $0.04$\,\% & $0.2$\,\%\\
X2, multi-record   & $90$\,\% & $6.0$\,\% & $22$\,\%\\
\bottomrule
\end{tabular}
\end{table}

\section{Discussion}
\label{sec:discussion}
This study provides the inverse counterpart to the forward tensegrity networks of \citet{Xu2025NAECON} and
\citet{QinChen2026SymPINN}. Here, the residual that those studies solve for the motion is used with the member parameters as unknowns. The constrained Cartesian formulation makes this identification tractable, because its mass matrix is constant and the force densities enter it linearly. Compared with the differentiable physics engines used in robotics \citep{Wang2023Real2Sim, Chen2024GNN}, the present formulation aims at the structural interpretation of the identified quantities. It adds an
identifiability analysis with Cram\'er--Rao bounds, and it works at the sensor counts and noise levels of structural monitoring instead of motion capture.

Within the structures and excitation suites tested, taut--slack transitions acted less as an obstacle to identification than as a source of information. Records with repeated taut--slack transitions were inverted with errors of a few percent on the X1 frame and the prism. Where low-amplitude forcing leaves members weakly
excited, as on the X2 frame, the strong slack-transition record recovered those members several times more accurately, to a worst-member error of $21$\,\%.
The gain from strong excitation outweighed the information lost while individual cables carried no force. On the already well-excited X1 frame and
prism, the two records give comparable force-density accuracy, with neither record better for every parameter.

The classical baseline and the X3 frame show the same pattern. Shooting converges to a spurious minimum on the low-amplitude taut record but recovers the slack-transition record almost exactly (Section~\ref{sec:res-baseline}). Two-point excitation with taut--slack transitions succeeds on the X3 frame; the low-amplitude taut record with the same two-point excitation still fails (Table~\ref{tab:scaleup}). Within the range tested, the excitation sets the limit of the single-record formulation. The prestress level sets a different limit. At the low-prestress end of the prism series, the information decreases 17-fold, and the identification reaches an optimization limit before the information runs out (Section~\ref{sec:res-family-prism}).

The simplest transition treatment was the most reliable of the three: the smoothed cable law was at least as accurate as the alternatives on the X1 and X2 comparisons. The complementarity treatment slowed convergence, event-aware collocation neither helped nor hurt, and neither overtook the smoothed cable law on the structures tested. No treatment was tuned separately.

The failure diagnostics have known limits. Fit failures raise the held-out ratio one to two orders of magnitude above the successful-run band. The silent failure of the gently forced X2 record is well fitted yet wrong by up to $80$\,\%, and neither the held-out ratio nor a local Fisher analysis at the exact parameters detects it; the Cram\'er--Rao bounds even show that the record is informative. Detecting such
practical non-uniqueness without the exact parameters, through multi-seed
scatter for instance, remains open. Richer excitation removes the failure
itself.

The Cram\'er--Rao bounds give the scatter that a record permits, and the eight-seed study gives the scatter achieved, which is $2.7$ to $8.3$ times that floor. With eight seeds, this ratio is only indicative. Neither measure captures the dominant error. The seed means of the force densities are $2$ to $4$\,\% from the exact values, an order of magnitude more than the seed scatter. The error therefore comes mainly from bias. An interval built from the spread of repeated fits, whether through dropout, an ensemble, or a variational posterior over the weights, would not capture this bias and could be narrow where the estimate is least accurate. This study reports bias and scatter separately for
that reason.

All evidence for identification through taut--slack transitions is simulated, on small to moderate structures with two to eight mechanical degrees of freedom and up to 16 cables. Damping is modeled as viscous and assigned to the cables, while measured tensegrity decay is dominated by joint friction \citep{OppenheimWilliams2001}. Axial stiffness is fixed except in the joint identification of Section~\ref{sec:res-robust}, where it traded error against the force densities as the Fisher analysis predicts. The
rigid-bar model with lumped inertia also omits the member internal
displacements that shape the high-frequency response \citep{Yuan2023CSD}.
Identification is thus confined to the low-frequency, mechanism-dominated
modes the benchmarks emphasize.

The most consequential of these limits is the massless-chord cable
model, in which a cable carries force along its chord and has no mass
of its own. A continuum cable supports one axial and two transverse
fields; the error of neglecting them depends on how far those modes lie from the mechanism modes. On the benchmarks the cable axial fundamentals lie near
$47$\,Hz and the transverse fundamentals between $10$ and $36$\,Hz, one to
two orders above mechanism modes near $1.4$\,Hz. The bar axial modes stand
three orders above. Transverse cable motion is therefore the first neglected effect that a wider frequency band would capture. Restoring it takes more than
the taut-string model. Its transverse wave speed $\sqrt{T_k/\rho A}$ vanishes with the tension, and the taut-string model therefore degenerates in the slack regime that this study addresses. The flexural rigidity, which the taut-string idealization discards, remains at zero tension. With it, the member behaves as a beam at zero tension, a well-posed model with finite frequencies. The two models differ only
below a crossover tension $T^{*}=EI(\pi/L)^2$ of order $10^{-3}$\,N, against
working tensions of tens of newtons. A transverse model that remains valid at slack would therefore add no error while the cable is taut, and it would be a natural basis for identifying cable mass and bending together with the prestress.

Section~\ref{sec:res-measured} is modal and taut: it validates the structural model and the prestress against a physical test, but not the time-domain residual inversion, because that dataset contains no forcing record. Sections~\ref{sec:res-wangleg} and~\ref{sec:res-olsztyn} add two
independent experiments, a biomimetic leg from another group with springs
measured on a separate rig, and the simplex of Ma{\l}yszko and Rutkiewicz with load cells in its cables. The leg experiment provides the most direct evidence for the time-domain residual, because its raw records were made available and they allow
the time-domain residual itself to be run on measured data, with the
drive known and the response measured. It recovers the knee spring to $0.1$\,\% in the mean over 15 records, on an identification
target the identifiability diagnostic selects in advance. The simplex provides the second measured validation. From 13 impact records at 13 measured prestress levels, the prestress is identified with a mean absolute error of \SI{32}{\newton}, inside the \SI{36}{\newton} scatter of the transducers. The hardware mass in this model comes from the elements weighed by the authors of the 2020 study, and a single joint-stiffness constant is fitted. None of the three experiments tests the treatment of taut--slack transitions, because each is excited about a taut prestressed equilibrium and no cable slackens. That validation requires a tensegrity test in which a cable slackens under measured excitation while the cable forces are recorded, and it is the next experimental step of this work.

The damping law is the one place where the measured data exceed what the formulation can represent. This study keeps the viscous law. The whole formulation and every simulated result are built on the axial coefficient of Section~\ref{sec:pinn}, and on records generated by that law it is recovered correctly. The simplex shows that it cannot represent the dissipation of a real
self-stressed structure. Section~\ref{sec:res-olsztyn} gives the structural
reason. The mechanism mode stiffened by the prestress barely extends any member, and extension-based damping is therefore nearly insensitive to it. Yet the measured
damping is near-uniform across modes whose strain energies differ by three
orders of magnitude. A hysteretic or joint-friction law would be the natural extension. It changes only the member force law and leaves the inverse formulation as it is, with the constant mass matrix and the linear force-density structure. Making that change here would
alter every result in Section~\ref{sec:results} without benefit for the taut--slack question that this study addresses; the change is therefore left to future work. The finding itself is reported in full because it bears on
what any identification of damping in a mechanism-dominated tensegrity
can claim.

\FloatBarrier
\section{Conclusions}
\label{sec:conclusions}
This study formulated physics-informed inverse identification for
tensegrity structural dynamics and examined it on benchmark structures, against classical identification methods, and on measured records from
laboratory structures. The residual is the constrained Cartesian equation of motion. Its constant mass matrix and the linear entry of the force densities make the identification trainable and the residual sensitivities closed-form. The tension-only cable law is retained through a smoothed cable law, and a Fisher-information analysis shows, before inversion, which member parameters a record can determine and to what precision.

Four findings summarize what was learned. First, within the cases tested, the excitation, more than the structure size or the sensor count, set the limit of a single-record identification. Strong records with taut--slack transitions recovered members that low-amplitude taut records left undetermined, adverse sensing increased the error less than weak excitation did, and the failures that occurred were removed by richer excitation, not by more sensors, more data of the same kind, or larger networks. Within the structures and excitation suites tested, the taut--slack transitions supplied more information than the slack intervals removed.

Second, a good response fit is not evidence of identification. One
benchmark record was fitted closely while its parameter errors approached $80$\,\%, and the same silent failure recurred at bridge scale. The held-out ratio flags fit failures; neither it nor a local Fisher analysis detects silent failures, and their detection without the exact parameters remains open.
Identification claims in this study are therefore based on the identifiability analysis and the parameter comparisons, never on fit quality.

Third, the benefit of taut--slack transitions is not specific to the physics-informed identification. Classical methods given the same records succeeded in opposite regimes: shooting converged to a spurious minimum on the taut record and was nearly exact on the slack-transition record, and the unscented filter did the reverse. The physics-informed identification succeeded on both from the same distant initialization, at a higher computational and tuning cost.

Fourth, the limits of a record can be predicted in advance. A
member restrained at both ends is invisible at any amplitude, taut or
slack; a first-order self-stress combination is recovered by a record with larger excitation; and grouping members that share a tension can remove the remaining degeneracy. The recovered prestress supports a stability and strength assessment whose margins are far less sensitive
to identification error than the individual force densities, because
the poorly recovered directions are the ones carrying no
free-coordinate stiffness.

All evidence for identification through taut--slack transitions is simulated, on structures of two to eight mechanical
degrees of freedom. The measured validations are taut: one modal, one time-domain, and one at 13 prestress states with a single fitted
hardware constant. Damping is modeled as viscous and axial. The measured data from a self-stressed structure show that this law cannot represent joint-dominated dissipation, and prestress and damping should therefore not be quoted with a shared confidence. Cables are modeled as massless force elements; the error of this model was measured by deliberate model-form mismatch, and a refinement for the slack regime is outlined in Section~\ref{sec:member-pde}. Extension to measured records with taut--slack transitions, to a
hysteretic or friction damping law, and to larger structures is
ongoing work.

\section*{Data Availability}
The simulated datasets and the code that generates them are available from the
authors. The biomimetic-leg measurements analyzed in
Section~\ref{sec:res-wangleg}, comprising 18 inertial-measurement-unit
chirp records together with the measured spring and component properties, were
provided by Y.~Wang and colleagues and are used here with their permission;
requests for those records should be directed to them. The simplex
measurements analyzed in Section~\ref{sec:res-olsztyn} combine the
published tables of \citet{Malyszko2020Simplex}, which is open access, with
impact-hammer records, element masses and joint geometry held by the
co-authors L.~Ma{\l}yszko and A.~Rutkiewicz; requests for those records
should be directed to them.

\section*{Acknowledgments}
The authors thank Yihang Wang for supplying the biomimetic-leg records and for
answering questions about the excitation.

\bibliographystyle{unsrtnat}
{\renewcommand{\bibfont}{\footnotesize\linespread{0.95}\selectfont}\bibliography{references}}

\begin{thebibliography}{76}
\providecommand{\natexlab}[1]{#1}
\providecommand{\url}[1]{\texttt{#1}}
\expandafter\ifx\csname urlstyle\endcsname\relax
  \providecommand{\doi}[1]{doi: #1}\else
  \providecommand{\doi}{doi: \begingroup \urlstyle{rm}\Url}\fi

\bibitem[Snelson(2012)]{Snelson2012}
Kenneth Snelson.
\newblock The art of tensegrity.
\newblock \emph{International Journal of Space Structures}, 27\penalty0
  (2-3):\penalty0 71--80, 2012.
\newblock \doi{10.1260/0266-3511.27.2-3.71}.

\bibitem[Motro(2003)]{Motro2003}
Ren\'e Motro.
\newblock \emph{Tensegrity: Structural Systems for the Future}.
\newblock Kogan Page Science, London, 2003.
\newblock ISBN 9781903996379.

\bibitem[Skelton and de~Oliveira(2009)]{Skelton2009book}
Robert~E. Skelton and Mauricio~C. de~Oliveira.
\newblock \emph{Tensegrity Systems}.
\newblock Springer, 2009.
\newblock \doi{10.1007/978-0-387-74242-7}.

\bibitem[Calladine(1978)]{Calladine1978}
C.~R. Calladine.
\newblock {Buckminster Fuller's ``Tensegrity'' structures and Clerk Maxwell's
  rules for the construction of stiff frames}.
\newblock \emph{International Journal of Solids and Structures}, 14\penalty0
  (2):\penalty0 161--172, 1978.
\newblock \doi{10.1016/0020-7683(78)90052-5}.

\bibitem[Pellegrino and Calladine(1986)]{PellegrinoCalladine1986}
S.~Pellegrino and C.~R. Calladine.
\newblock Matrix analysis of statically and kinematically indeterminate
  frameworks.
\newblock \emph{International Journal of Solids and Structures}, 22\penalty0
  (4):\penalty0 409--428, 1986.
\newblock \doi{10.1016/0020-7683(86)90014-4}.

\bibitem[Ashwear and Eriksson(2014)]{AshwearEriksson2014}
Nasseradeen Ashwear and Anders Eriksson.
\newblock Natural frequencies describe the pre-stress in tensegrity structures.
\newblock \emph{Computers \& Structures}, 138:\penalty0 162--171, 2014.
\newblock \doi{10.1016/j.compstruc.2014.01.020}.

\bibitem[Sychterz and Smith(2018)]{SychterzSmith2018}
Ann~C. Sychterz and Ian F.~C. Smith.
\newblock Using dynamic measurements to detect and locate ruptured cables on a
  tensegrity structure.
\newblock \emph{Engineering Structures}, 173:\penalty0 631--642, 2018.
\newblock \doi{10.1016/j.engstruct.2018.06.083}.

\bibitem[Tibert and Pellegrino(2003)]{TibertPellegrino2003}
A.~G. Tibert and S.~Pellegrino.
\newblock Review of form-finding methods for tensegrity structures.
\newblock \emph{International Journal of Space Structures}, 18\penalty0
  (4):\penalty0 209--223, 2003.
\newblock \doi{10.1260/026635103322987940}.

\bibitem[Murakami(2001)]{Murakami2001Part1}
Hidenori Murakami.
\newblock Static and dynamic analyses of tensegrity structures. part 1.
  nonlinear equations of motion.
\newblock \emph{International Journal of Solids and Structures}, 38\penalty0
  (20):\penalty0 3599--3613, 2001.
\newblock \doi{10.1016/S0020-7683(00)00232-8}.

\bibitem[Goyal and Skelton(2019)]{GoyalSkelton2019}
Raman Goyal and Robert~E. Skelton.
\newblock Tensegrity system dynamics with rigid bars and massive strings.
\newblock \emph{Multibody System Dynamics}, 46\penalty0 (3):\penalty0 203--228,
  2019.
\newblock \doi{10.1007/s11044-019-09666-4}.

\bibitem[Ljung(1999)]{Ljung1999}
Lennart Ljung.
\newblock \emph{System Identification: Theory for the User}.
\newblock Prentice Hall Information and System Sciences Series. Prentice Hall,
  Englewood Cliffs, NJ, 2nd edition, 1999.
\newblock ISBN 978-0-13-656695-3.

\bibitem[Friswell and Mottershead(1995)]{FriswellMottershead1995}
Michael~I. Friswell and John~E. Mottershead.
\newblock \emph{Finite Element Model Updating in Structural Dynamics},
  volume~38 of \emph{Solid Mechanics and Its Applications}.
\newblock Springer, Dordrecht, 1995.
\newblock ISBN 978-0-7923-3431-6.
\newblock \doi{10.1007/978-94-015-8508-8}.

\bibitem[Farrar and Worden(2013)]{FarrarWorden2013}
Charles~R. Farrar and Keith Worden.
\newblock \emph{Structural Health Monitoring: A Machine Learning Perspective}.
\newblock John Wiley \& Sons, Chichester, UK, 2013.
\newblock ISBN 978-1-119-99433-6.
\newblock \doi{10.1002/9781118443118}.

\bibitem[Dub{\'e} and Angellier(2013)]{DubeAngellier2013}
Jean~Fran{\c c}ois Dub{\'e} and Nicolas Angellier.
\newblock Using of tensegrity grid dynamic behavior to identify its self-stress
  level.
\newblock \emph{Journal of Constructional Steel Research}, 90:\penalty0
  133--139, 2013.
\newblock \doi{10.1016/j.jcsr.2013.07.026}.

\bibitem[Ma{\l}yszko and Rutkiewicz(2020)]{Malyszko2020Simplex}
Leszek Ma{\l}yszko and Andrzej Rutkiewicz.
\newblock Response of a tensegrity simplex in experimental tests of a modal
  hammer at different self-stress levels.
\newblock \emph{Applied Sciences}, 10\penalty0 (23):\penalty0 8733, 2020.
\newblock \doi{10.3390/app10238733}.

\bibitem[Bel Hadj~Ali and Smith(2010)]{BelHadjAliSmith2010IJSS}
Nizar Bel Hadj~Ali and Ian F.~C. Smith.
\newblock Dynamic behavior and vibration control of a tensegrity structure.
\newblock \emph{International Journal of Solids and Structures}, 47\penalty0
  (9):\penalty0 1285--1296, 2010.
\newblock \doi{10.1016/j.ijsolstr.2010.01.012}.

\bibitem[Adam and Smith(2007)]{AdamSmith2007}
Bernard Adam and Ian F.~C. Smith.
\newblock Self-diagnosis and self-repair of an active tensegrity structure.
\newblock \emph{Journal of Structural Engineering}, 133\penalty0 (12):\penalty0
  1752--1761, 2007.
\newblock \doi{10.1061/(ASCE)0733-9445(2007)133:12(1752)}.

\bibitem[Aloui et~al.(2019)Aloui, Lin, and
  Rhode-Barbarigos]{Aloui2019SensorPlacement}
Omar Aloui, Jan Lin, and Landolf Rhode-Barbarigos.
\newblock A theoretical framework for sensor placement, structural
  identification and damage detection in tensegrity structures.
\newblock \emph{Smart Materials and Structures}, 28\penalty0 (12):\penalty0
  125004, 2019.
\newblock \doi{10.1088/1361-665X/ab3d21}.

\bibitem[Tian et~al.(2024)Tian, Zhang, Spencer, and
  Tao]{Tian2024GlobalCableForce}
Xiang Tian, Yuxin Zhang, Billie~F. Spencer, and Chengcheng Tao.
\newblock A study of global cable force identification of tensegrity
  structures.
\newblock \emph{Structures}, 65:\penalty0 106610, 2024.
\newblock \doi{10.1016/j.istruc.2024.106610}.

\bibitem[Huang et~al.(2025)Huang, Zhao, Lin, Shi, Ye, Luo, Wu, Jiang, and
  Wang]{Prism2025JSV}
Ke~Huang, Ziheng Zhao, Chengrong Lin, Meiqin Shi, Dexin Ye, Jiahui Luo, Zhigang
  Wu, Jianping Jiang, and Xing Wang.
\newblock Towards validated dynamic models for tensegrity structures:
  Parametric modelling, vibration testing, and model updating of a hexagonal
  prism.
\newblock \emph{Journal of Sound and Vibration}, 618:\penalty0 119350, 2025.
\newblock \doi{10.1016/j.jsv.2025.119350}.

\bibitem[Aswal et~al.(2021)Aswal, Sen, and Mevel]{Aswal2021}
Neha Aswal, Subhamoy Sen, and Laurent Mevel.
\newblock Estimation of local failure in tensegrity using interacting
  particle-ensemble {K}alman filter.
\newblock \emph{Mechanical Systems and Signal Processing}, 160:\penalty0
  107824, 2021.
\newblock \doi{10.1016/j.ymssp.2021.107824}.

\bibitem[Chatzi and Smyth(2009)]{ChatziSmyth2009UKF}
Eleni~N. Chatzi and Andrew~W. Smyth.
\newblock The unscented {K}alman filter and particle filter methods for
  nonlinear structural system identification with non-collocated heterogeneous
  sensing.
\newblock \emph{Structural Control and Health Monitoring}, 16\penalty0
  (1):\penalty0 99--123, 2009.
\newblock \doi{10.1002/stc.290}.

\bibitem[Bel Hadj~Ali et~al.(2010)Bel Hadj~Ali, Rhode-Barbarigos, Pascual~Albi,
  and Smith]{BelHadjAli2010}
Nizar Bel Hadj~Ali, Landolf Rhode-Barbarigos, Alberto~A. Pascual~Albi, and Ian
  F.~C. Smith.
\newblock Design optimization and dynamic analysis of a tensegrity-based
  footbridge.
\newblock \emph{Engineering Structures}, 32\penalty0 (11):\penalty0 3650--3659,
  2010.
\newblock \doi{10.1016/j.engstruct.2010.08.009}.

\bibitem[Oppenheim and Williams(2001)]{OppenheimWilliams2001}
Irving~J. Oppenheim and William~O. Williams.
\newblock Vibration and damping in three-bar tensegrity structure.
\newblock \emph{Journal of Aerospace Engineering}, 14\penalty0 (3):\penalty0
  85--91, 2001.
\newblock \doi{10.1061/(ASCE)0893-1321(2001)14:3(85)}.

\bibitem[Lagaris et~al.(1998)Lagaris, Likas, and Fotiadis]{Lagaris1998}
Isaac~E. Lagaris, Aristidis Likas, and Dimitrios~I. Fotiadis.
\newblock Artificial neural networks for solving ordinary and partial
  differential equations.
\newblock \emph{IEEE Transactions on Neural Networks}, 9\penalty0 (5):\penalty0
  987--1000, 1998.
\newblock \doi{10.1109/72.712178}.

\bibitem[Raissi et~al.(2019)Raissi, Perdikaris, and Karniadakis]{Raissi2019}
Maziar Raissi, Paris Perdikaris, and George~E. Karniadakis.
\newblock Physics-informed neural networks: a deep learning framework for
  solving forward and inverse problems involving nonlinear partial differential
  equations.
\newblock \emph{Journal of Computational Physics}, 378:\penalty0 686--707,
  2019.
\newblock \doi{10.1016/j.jcp.2018.10.045}.

\bibitem[Karniadakis et~al.(2021)Karniadakis, Kevrekidis, Lu, Perdikaris, Wang,
  and Yang]{Karniadakis2021}
George~Em Karniadakis, Ioannis~G. Kevrekidis, Lu~Lu, Paris Perdikaris, Sifan
  Wang, and Liu Yang.
\newblock Physics-informed machine learning.
\newblock \emph{Nature Reviews Physics}, 3\penalty0 (6):\penalty0 422--440,
  2021.
\newblock \doi{10.1038/s42254-021-00314-5}.

\bibitem[Haghighat et~al.(2021)Haghighat, Raissi, Moure, Gomez, and
  Juanes]{Haghighat2021SolidPINN}
Ehsan Haghighat, Maziar Raissi, Adrian Moure, Hector Gomez, and Ruben Juanes.
\newblock A physics-informed deep learning framework for inversion and
  surrogate modeling in solid mechanics.
\newblock \emph{Computer Methods in Applied Mechanics and Engineering},
  379:\penalty0 113741, 2021.
\newblock \doi{10.1016/j.cma.2021.113741}.

\bibitem[Guo and Fang(2023)]{Guo2023PINNStructural}
Xin-Yu Guo and Sheng-En Fang.
\newblock Structural parameter identification using physics-informed neural
  networks.
\newblock \emph{Measurement}, 220:\penalty0 113334, 2023.
\newblock \doi{10.1016/j.measurement.2023.113334}.

\bibitem[Liu and Meidani(2023)]{LiuMeidani2023JEM}
Tong Liu and Hadi Meidani.
\newblock Physics-informed neural networks for system identification of
  structural systems with a multiphysics damping model.
\newblock \emph{Journal of Engineering Mechanics}, 149\penalty0 (10):\penalty0
  04023079, 2023.
\newblock \doi{10.1061/JENMDT.EMENG-7060}.

\bibitem[Al-Adly and Kripakaran(2024)]{AlAdlyKripakaran2024}
Anmar I.~F. Al-Adly and Prakash Kripakaran.
\newblock Physics-informed neural networks for structural health monitoring: a
  case study for {K}irchhoff--{L}ove plates.
\newblock \emph{Data-Centric Engineering}, 5, 2024.
\newblock \doi{10.1017/dce.2024.4}.

\bibitem[Yang et~al.(2021)Yang, Meng, and Karniadakis]{Yang2021BPINN}
Liu Yang, Xuhui Meng, and George~Em Karniadakis.
\newblock {B-PINNs}: Bayesian physics-informed neural networks for forward and
  inverse {PDE} problems with noisy data.
\newblock \emph{Journal of Computational Physics}, 425:\penalty0 109913, 2021.
\newblock \doi{10.1016/j.jcp.2020.109913}.

\bibitem[Mahar et~al.(2025)Mahar, Sen, and Mevel]{Mahar2025MSSP}
Nikhil Mahar, Subhamoy Sen, and Laurent Mevel.
\newblock Physics informed long short-term memory neural network for dual
  state-parameter estimation of linear dynamical systems robust to input
  forces.
\newblock \emph{Mechanical Systems and Signal Processing}, 236:\penalty0
  113012, 2025.
\newblock \doi{10.1016/j.ymssp.2025.113012}.

\bibitem[Trinh et~al.(2022)Trinh, Lee, Kang, and Lee]{Trinh2022}
Duy T.~N. Trinh, Seunghye Lee, Joowon Kang, and Jaehong Lee.
\newblock Force density-informed neural network for prestress design of
  tensegrity structures with multiple self-stress modes.
\newblock \emph{European Journal of Mechanics -- A/Solids}, 94:\penalty0
  104584, 2022.
\newblock \doi{10.1016/j.euromechsol.2022.104584}.

\bibitem[Xu et~al.(2025)Xu, Aono, Chen, and Qin]{Xu2025NAECON}
Youyun Xu, Haruku Aono, Muhao Chen, and Jing Qin.
\newblock Predicting dynamics of tensegrity structures using physics-informed
  neural networks.
\newblock In \emph{NAECON 2025 -- IEEE National Aerospace and Electronics
  Conference}, pages 1--6, 2025.
\newblock \doi{10.1109/NAECON65708.2025.11235432}.

\bibitem[Wang et~al.(2025)Wang, Zhu, and Miao]{WangZhuMiao2025}
Jin Wang, Mingliang Zhu, and Zhiwei Miao.
\newblock Two-stage physics-informed deep neural networks framework for
  form-finding of tensegrity structures.
\newblock \emph{Computer-Aided Design}, 186--187:\penalty0 103898, 2025.
\newblock \doi{10.1016/j.cad.2025.103898}.

\bibitem[Qin and Chen(2026)]{QinChen2026SymPINN}
Jing Qin and Muhao Chen.
\newblock Symmetry-reduced physics-informed learning of tensegrity dynamics.
\newblock arXiv:2603.17824, 2026.

\bibitem[Song et~al.(2025)Song, Wang, Wang, and Peng]{Song2025NLDyn}
Ningning Song, Mingji Wang, Xinwei Wang, and Haijun Peng.
\newblock A novel machine learning method for real-time dynamic analysis of
  tensegrity flexible multibody systems.
\newblock \emph{Nonlinear Dynamics}, 113\penalty0 (15):\penalty0 19047--19074,
  2025.
\newblock \doi{10.1007/s11071-025-11152-w}.

\bibitem[Wang et~al.(2023)Wang, Johnson, Lu, Huang, Booth, Kramer-Bottiglio,
  Aanjaneya, and Bekris]{Wang2023Real2Sim}
Kun Wang, William~R. Johnson, III, Shiyang Lu, Xiaonan Huang, Joran Booth,
  Rebecca Kramer-Bottiglio, Mridul Aanjaneya, and Kostas Bekris.
\newblock {Real2Sim2Real} transfer for control of cable-driven robots via a
  differentiable physics engine.
\newblock In \emph{2023 IEEE/RSJ International Conference on Intelligent Robots
  and Systems (IROS)}, pages 2534--2541, 2023.
\newblock \doi{10.1109/IROS55552.2023.10341811}.

\bibitem[Chen et~al.(2025)Chen, Wang, Johnson, Kramer-Bottiglio, Bekris, and
  Aanjaneya]{Chen2024GNN}
Nelson Chen, Kun Wang, William~R. Johnson, III, Rebecca Kramer-Bottiglio,
  Kostas Bekris, and Mridul Aanjaneya.
\newblock Learning differentiable tensegrity dynamics using graph neural
  networks.
\newblock In \emph{Proceedings of the 8th Conference on Robot Learning (CoRL
  2024)}, volume 270 of \emph{Proceedings of Machine Learning Research}, pages
  5134--5149, 2025.

\bibitem[Chatzis and Chatzi(2017)]{Chatzis2017DUKF}
Manolis~N. Chatzis and Eleni~N. Chatzi.
\newblock A discontinuous unscented {Kalman} filter for non-smooth dynamic
  problems.
\newblock \emph{Frontiers in Built Environment}, 3:\penalty0 56, 2017.
\newblock \doi{10.3389/fbuil.2017.00056}.

\bibitem[Kong et~al.(2024)Kong, Payne, Zhu, and Johnson]{Kong2024Saltation}
Nathan~J. Kong, J.~Joe Payne, James Zhu, and Aaron~M. Johnson.
\newblock Saltation matrices: The essential tool for linearizing hybrid
  dynamical systems.
\newblock \emph{Proceedings of the IEEE}, 112\penalty0 (6):\penalty0 585--608,
  2024.
\newblock \doi{10.1109/JPROC.2024.3440211}.

\bibitem[Sahin et~al.(2024)Sahin, von Danwitz, and Popp]{Sahin2024Contact}
Tarik Sahin, Max von Danwitz, and Alexander Popp.
\newblock Solving forward and inverse problems of contact mechanics using
  physics-informed neural networks.
\newblock \emph{Advanced Modeling and Simulation in Engineering Sciences},
  11\penalty0 (1):\penalty0 11, 2024.
\newblock \doi{10.1186/s40323-024-00265-3}.

\bibitem[Li et~al.(2024)Li, Bai, Ouyang, Martelli, Tang, Yang, Wei, Liu, Wei,
  and Gu]{Li2024Friction}
Zilin Li, Jinshuai Bai, Huajiang Ouyang, Saulo Martelli, Ming Tang, Yang Yang,
  Hongtao Wei, Pan Liu, Ronghan Wei, and Yuantong Gu.
\newblock Physics-informed neural networks for friction-involved nonsmooth
  dynamics problems.
\newblock \emph{Nonlinear Dynamics}, 112\penalty0 (9):\penalty0 7159--7183,
  2024.
\newblock \doi{10.1007/s11071-024-09350-z}.
\newblock arXiv:2303.02542.

\bibitem[Friswell et~al.(1998)Friswell, Mottershead, and
  Ahmadian]{Friswell1998Subset}
Michael~I. Friswell, John~E. Mottershead, and Hamid Ahmadian.
\newblock Combining subset selection and parameter constraints in model
  updating.
\newblock \emph{Journal of Vibration and Acoustics}, 120\penalty0 (4):\penalty0
  854--859, 1998.
\newblock \doi{10.1115/1.2893911}.

\bibitem[Moored and Bart-Smith(2009)]{Moored2009Clustered}
Keith~W. Moored and Hilary Bart-Smith.
\newblock Investigation of clustered actuation in tensegrity structures.
\newblock \emph{International Journal of Solids and Structures}, 46\penalty0
  (17):\penalty0 3272--3281, 2009.
\newblock \doi{10.1016/j.ijsolstr.2009.04.026}.

\bibitem[Tadiparthi et~al.(2019)Tadiparthi, Hsu, and
  Bhattacharya]{Tadikonda2019STEDY}
Vaishnav Tadiparthi, Shao-Chen Hsu, and Raktim Bhattacharya.
\newblock {STEDY}: Software for {TE}nsegrity {DY}namics.
\newblock \emph{Journal of Open Source Software}, 4\penalty0 (33):\penalty0
  1042, 2019.
\newblock \doi{10.21105/joss.01042}.

\bibitem[Oliveto and Sivaselvan(2011)]{OlivetoSivaselvan2011}
Nicholas~D. Oliveto and Mettupalayam~V. Sivaselvan.
\newblock Dynamic analysis of tensegrity structures using a complementarity
  framework.
\newblock \emph{Computers \& Structures}, 89\penalty0 (23--24):\penalty0
  2471--2483, 2011.
\newblock \doi{10.1016/j.compstruc.2011.06.003}.

\bibitem[Kan et~al.(2018)Kan, Peng, and Chen]{Kan2018}
Ziyun Kan, Haijun Peng, and Biaosong Chen.
\newblock Complementarity framework for nonlinear analysis of tensegrity
  structures with slack cables.
\newblock \emph{AIAA Journal}, 56\penalty0 (12):\penalty0 5013--5027, 2018.
\newblock \doi{10.2514/1.J057149}.

\bibitem[Rao et~al.(2021)Rao, Sun, and Liu]{RaoSunLiu2021Elastodynamics}
Chengping Rao, Hao Sun, and Yang Liu.
\newblock Physics-informed deep learning for computational elastodynamics
  without labeled data.
\newblock \emph{Journal of Engineering Mechanics}, 147\penalty0 (8):\penalty0
  04021043, 2021.
\newblock \doi{10.1061/(ASCE)EM.1943-7889.0001947}.

\bibitem[Lu et~al.(2021)Lu, Pestourie, Yao, Wang, Verdugo, and
  Johnson]{Lu2021HardConstraintPINN}
Lu~Lu, Raphael Pestourie, Wenjie Yao, Zhicheng Wang, Francesc Verdugo, and
  Steven~G. Johnson.
\newblock Physics-informed neural networks with hard constraints for inverse
  design.
\newblock \emph{SIAM Journal on Scientific Computing}, 43\penalty0
  (6):\penalty0 B1105--B1132, 2021.
\newblock \doi{10.1137/21M1397908}.

\bibitem[Kingma and Ba(2015)]{KingmaBa2015}
Diederik~P. Kingma and Jimmy Ba.
\newblock Adam: A method for stochastic optimization.
\newblock In \emph{3rd International Conference on Learning Representations
  (ICLR 2015)}, San Diego, CA, USA, 2015.
\newblock \doi{10.48550/arXiv.1412.6980}.
\newblock arXiv:1412.6980.

\bibitem[Loshchilov and Hutter(2017)]{LoshchilovHutter2017}
Ilya Loshchilov and Frank Hutter.
\newblock {SGDR}: Stochastic gradient descent with warm restarts.
\newblock In \emph{International Conference on Learning Representations
  (ICLR)}, 2017.

\bibitem[Udwadia(1994)]{Udwadia1994}
Firdaus~E. Udwadia.
\newblock Methodology for optimum sensor locations for parameter identification
  in dynamic systems.
\newblock \emph{Journal of Engineering Mechanics}, 120\penalty0 (2):\penalty0
  368--390, 1994.
\newblock \doi{10.1061/(ASCE)0733-9399(1994)120:2(368)}.

\bibitem[Kammer(1991)]{Kammer1991}
Daniel~C. Kammer.
\newblock Sensor placement for on-orbit modal identification and correlation of
  large space structures.
\newblock \emph{Journal of Guidance, Control, and Dynamics}, 14\penalty0
  (2):\penalty0 251--259, 1991.
\newblock \doi{10.2514/3.20635}.

\bibitem[Shahverdi et~al.(2009)Shahverdi, Mares, Wang, and
  Mottershead]{Shahverdi2009Clustering}
Hossein Shahverdi, Cristinel Mares, Weizhuo Wang, and John~E. Mottershead.
\newblock Clustering of parameter sensitivities: Examples from a helicopter
  airframe model updating exercise.
\newblock \emph{Shock and Vibration}, 16\penalty0 (1):\penalty0 75--87, 2009.
\newblock \doi{10.1155/2009/982439}.

\bibitem[Levenberg(1944)]{Levenberg1944}
Kenneth Levenberg.
\newblock A method for the solution of certain non-linear problems in least
  squares.
\newblock \emph{Quarterly of Applied Mathematics}, 2\penalty0 (2):\penalty0
  164--168, 1944.
\newblock \doi{10.1090/qam/10666}.

\bibitem[Marquardt(1963)]{Marquardt1963}
Donald~W. Marquardt.
\newblock An algorithm for least-squares estimation of nonlinear parameters.
\newblock \emph{Journal of the Society for Industrial and Applied Mathematics},
  11\penalty0 (2):\penalty0 431--441, 1963.
\newblock \doi{10.1137/0111030}.

\bibitem[Julier and Uhlmann(1997)]{JulierUhlmann1997UT}
Simon~J. Julier and Jeffrey~K. Uhlmann.
\newblock A new extension of the {K}alman filter to nonlinear systems.
\newblock In \emph{Signal Processing, Sensor Fusion, and Target Recognition
  VI}, volume 3068 of \emph{Proceedings of SPIE}, pages 182--193. SPIE, 1997.
\newblock \doi{10.1117/12.280797}.
\newblock AeroSense '97, Orlando, FL.

\bibitem[Wan and Van Der~Merwe(2000)]{WanVanDerMerwe2000UKF}
Eric~A. Wan and Rudolph Van Der~Merwe.
\newblock The unscented {K}alman filter for nonlinear estimation.
\newblock In \emph{Proceedings of the IEEE 2000 Adaptive Systems for Signal
  Processing, Communications, and Control Symposium (AS-SPCC)}, pages 153--158.
  IEEE, 2000.
\newblock \doi{10.1109/ASSPCC.2000.882463}.

\bibitem[Wang et~al.(2026)Wang, He, Li, Gao, Jiang, and Zhang]{Wang2026Leg}
Yihang Wang, Jiawei He, Xin Li, Yang Gao, Hongzhou Jiang, and Hui Zhang.
\newblock Linearized vibration analysis of tensegrity structures in rigid-body
  space.
\newblock \emph{International Journal of Mechanical Sciences}, 311:\penalty0
  111178, 2026.
\newblock \doi{10.1016/j.ijmecsci.2026.111178}.

\bibitem[Yuan and Zhu(2023)]{Yuan2023CSD}
Sichen Yuan and Weidong Zhu.
\newblock A {C}artesian spatial discretization method for nonlinear dynamic
  modeling and vibration analysis of tensegrity structures.
\newblock \emph{International Journal of Solids and Structures}, 270:\penalty0
  112179, 2023.
\newblock \doi{10.1016/j.ijsolstr.2023.112179}.

\bibitem[Ma et~al.(2022)Ma, Chen, and Skelton]{MaChen2022TsgFEM}
Shuo Ma, Muhao Chen, and Robert~E. Skelton.
\newblock {TsgFEM}: Tensegrity finite element method.
\newblock \emph{Journal of Open Source Software}, 7\penalty0 (75):\penalty0
  3390, 2022.
\newblock \doi{10.21105/joss.03390}.

\bibitem[{European Committee for Standardization}(2008)]{EN12385}
{European Committee for Standardization}.
\newblock {EN 12385-10}: Steel wire ropes --- safety --- part 10: Spiral ropes
  for general structural applications.
\newblock Standard, CEN, Brussels, 2008.

\bibitem[{ASTM International}(2018)]{ASTMA416}
{ASTM International}.
\newblock {ASTM A416/A416M-18}: Standard specification for low-relaxation,
  seven-wire steel strand for prestressed concrete.
\newblock Standard, ASTM International, West Conshohocken, PA, 2018.

\bibitem[Carpentieri et~al.(2014)Carpentieri, Skelton, and
  Fraternali]{Carpentieri2014Parametric}
Gerardo Carpentieri, Robert~E. Skelton, and Fernando Fraternali.
\newblock Parametric design of minimal mass tensegrity bridges under yielding
  and buckling constraints, 2014.

\bibitem[Connelly and Back(1998)]{Connelly1998}
Robert Connelly and Allen Back.
\newblock Mathematics and tensegrity.
\newblock \emph{American Scientist}, 86\penalty0 (2):\penalty0 142--151, 1998.

\bibitem[Fern{\'a}ndez-Ruiz et~al.(2022)Fern{\'a}ndez-Ruiz,
  Hern{\'a}ndez-Montes, and Gil-Mart{\'i}n]{FernandezRuiz2022}
J.~F. Fern{\'a}ndez-Ruiz, E.~Hern{\'a}ndez-Montes, and L.~M. Gil-Mart{\'i}n.
\newblock Topological design of the octahedron tensegrity family.
\newblock \emph{Engineering Structures}, 265:\penalty0 114490, 2022.

\bibitem[Zhang and Ohsaki(2012)]{ZhangOhsakiJMPS2012}
Jing~Yao Zhang and Makoto Ohsaki.
\newblock Self-equilibrium and stability of regular truncated tetrahedral
  tensegrity structures.
\newblock \emph{Journal of the Mechanics and Physics of Solids}, 60\penalty0
  (10):\penalty0 1757--1770, 2012.

\bibitem[Zhang and Ohsaki(2015)]{ZhangOhsaki2015}
Jing~Yao Zhang and Makoto Ohsaki.
\newblock \emph{Tensegrity Structures: Form, Stability, and Symmetry}, volume~6
  of \emph{Mathematics for Industry}.
\newblock Springer Japan, Tokyo, 2015.
\newblock ISBN 9784431548126.
\newblock \doi{10.1007/978-4-431-54813-3}.

\bibitem[Papadimitriou(2004)]{Papadimitriou2004}
Costas Papadimitriou.
\newblock Optimal sensor placement methodology for parametric identification of
  structural systems.
\newblock \emph{Journal of Sound and Vibration}, 278\penalty0 (4--5):\penalty0
  923--947, 2004.
\newblock \doi{10.1016/j.jsv.2003.10.063}.

\bibitem[Moaveni et~al.(2009)Moaveni, Conte, and Hemez]{Moaveni2009Uncertainty}
Babak Moaveni, Joel~P. Conte, and Fran{\c{c}}ois~M. Hemez.
\newblock Uncertainty and sensitivity analysis of damage identification results
  obtained using finite element model updating.
\newblock \emph{Computer-Aided Civil and Infrastructure Engineering},
  24\penalty0 (5):\penalty0 320--334, 2009.
\newblock \doi{10.1111/j.1467-8667.2008.00589.x}.

\bibitem[Bao et~al.(2013)Bao, Li, Sun, Yu, and Ou]{Bao2013Compressive}
Yuequan Bao, Hui Li, Xiaodan Sun, Yan Yu, and Jinping Ou.
\newblock Compressive sampling-based data loss recovery for wireless sensor
  networks used in civil structural health monitoring.
\newblock \emph{Structural Health Monitoring}, 12\penalty0 (1):\penalty0
  78--95, 2013.
\newblock \doi{10.1177/1475921712462936}.

\bibitem[Zhang et~al.(2024)Zhang, Huang, Wan, Deng, He, and
  Luo]{Zhang2024Missing}
Jianwei Zhang, Minshui Huang, Neng Wan, Zhihang Deng, Zhongao He, and Jin Luo.
\newblock Missing measurement data recovery methods in structural health
  monitoring: The state, challenges and case study.
\newblock \emph{Measurement}, 231:\penalty0 114528, 2024.
\newblock \doi{10.1016/j.measurement.2024.114528}.

\bibitem[Ewins(2000)]{Ewins2000Modal}
David~J. Ewins.
\newblock \emph{Modal Testing: Theory, Practice and Application}.
\newblock Research Studies Press, Baldock, Hertfordshire, England, 2nd edition,
  2000.
\newblock ISBN 9780863802188.

\bibitem[Chen et~al.(2021)Chen, Liu, and Sun]{Chen2021Scarce}
Zhao Chen, Yang Liu, and Hao Sun.
\newblock Physics-informed learning of governing equations from scarce data.
\newblock \emph{Nature Communications}, 12:\penalty0 6136, 2021.
\newblock \doi{10.1038/s41467-021-26434-1}.

\end{thebibliography}

\clearpage
\pdfbookmark[0]{Supplementary Material}{supplement}
\begin{center}
{\Large\bfseries Supplementary Material}
\end{center}

This supplement contains the implementation details and extended analyses
that support the main text: the alternative treatments of the taut--slack
transition, the optimizer and loss-weight details with the full parameter
table, the assessment of the classic tensegrity forms, the exclusion
analysis and hip statistics behind the biomimetic-leg reconstruction, the
structural-assessment formulation with its demonstration on exact force
densities, and the robustness studies on noise, model-form error, and
adverse sensing. Section, table, and equation numbers of the form S1, S2
refer to this supplement; unprefixed numbers refer to the main text.

\setcounter{section}{0}
\setcounter{figure}{0}
\setcounter{table}{0}
\setcounter{equation}{0}
\renewcommand{\thesection}{S\arabic{section}}
\renewcommand{\thefigure}{S\arabic{figure}}
\renewcommand{\thetable}{S\arabic{table}}
\renewcommand{\theequation}{S\arabic{equation}}
\renewcommand{\theHsection}{S\arabic{section}}
\renewcommand{\theHfigure}{S\arabic{figure}}
\renewcommand{\theHtable}{S\arabic{table}}
\renewcommand{\theHequation}{S\arabic{equation}}

\section{Alternative treatments of the taut--slack transition}
\label{sm:switch}

The main text trains with the smoothed cable law and compares two alternatives with it on identical records. Their construction is as follows.

The complementarity treatment keeps the exact law and moves the
nonsmoothness out of the residual, adapting the static-contact
construction of \citet{Sahin2024Contact} to the dynamic setting. An
auxiliary tension head outputs a non-negative tension $T_k(\tau)$ for every cable, and a softplus output layer enforces the positivity. A nominal tension scales this output. It is the $10$\,N/m initialization force density times the mean member length of the reference geometry, and it therefore carries no information about the prestress being identified. The cable
term of the equation-of-motion residual uses the force density
$T_k/\lVert s_k\rVert$. A Fischer--Burmeister residual then ties the
tension to the elastic law,
\begin{equation}
  r_{\mathrm{NCP},k} \;=\; \phi_{\mathrm{FB}}\bigl(T_k,\, g_k\bigr),
  \qquad
  \phi_{\mathrm{FB}}(a,b) \;=\; a + b - \sqrt{a^2 + b^2},
  \label{eq:ncp}
\end{equation}
with $g_k$ the elastic slackness defined in the main text. Tension and slackness are normalized by member scales before they enter the loss, and the loss gains the term $10\,\overline{\smash{r_{\mathrm{NCP}}^{\,2}}}$.
The zero set of $\phi_{\mathrm{FB}}$ is exactly the tension-only cable law: $r_{\mathrm{NCP},k} = 0$ if and only if $T_k \ge 0$, $g_k \ge 0$, and $T_k g_k = 0$. The cable is then either taut and elastic or slack and force-free. The function $\phi_{\mathrm{FB}}$ is smooth everywhere except at $T_k = g_k = 0$, where the cable is just taut and carries no force. This reduces the nonsmooth set from the whole taut--slack boundary to one point per cable.

The event-aware treatment keeps the smoothed cable law and concentrates the collocation points where its bias occurs. Every 500 steps, the current solution is checked for taut--slack transitions. Half of each batch's collocation
points are then drawn from narrow Gaussian windows around the detected transition times; the other half remain uniform. Apart from the tension
head and the complementarity term, the three treatments share the same
trunk architecture, loss weights and normalizations, and optimizer
settings.

\section{Optimizer, loss weights, and run archival}
\label{sm:optimizer}

The larger learning rate on the physical parameter group can be used because those parameters are represented in logarithmic form. The size of an Adam step is set by the normalized gradient, not by the gradient magnitude. A step in $\log\sigma_k$ is therefore multiplicative in $\sigma_k$, and one relative step size serves force densities, an
axial stiffness, and damper coefficients whose numerical values differ
by orders of magnitude.

Table~\ref{tab:weights} varies the loss weights on the X1 taut record.
Below $w_b \simeq 10$, the bar-length constraint is not enforced and
every quantity degrades together, reaching $29.5$\,\% at $w_b = 1$.
Above this value, the recovery is flat to within about 2 percentage points across three further decades, and the value used throughout, $w_b = 100$, lies in the middle of this plateau. Between $w_{\mathrm{eom}} = 0.3$ and $3$, the mean force-density error changes by about 1 percentage point; at $w_{\mathrm{eom}} = 10$, where the sensor data are effectively ignored, the error increases sharply.

The third block of the table carries a record with a model-form error
the residual cannot represent. Down-weighting the residual to
$w_{\mathrm{eom}} = 0.1$ makes the identification worse, $6.0$ against
$5.8$\,\%, and the best value is at or slightly above equal weighting. The residual remains informative about the parameters even when it contains a model-form error, because this error is structured by the geometry and is not random. Reducing the weight of the residual discards this structure together with the error. The penalty for keeping
the equal weighting used throughout is $0.06$ percentage points.

\begin{table}[!tbp]
  \centering
  \caption{Sensitivity of the identification to the loss weights, on the
  X1 taut record. Errors are over the three identifiable force densities;
  the inert base cable is excluded, being unidentifiable by construction.
  A dagger marks the value used throughout.}
  \label{tab:weights}
  \footnotesize
  \begin{tabular}{@{}rccc@{}}
\toprule
& \multicolumn{2}{c}{force density (\%)} & damping (\%)\\
\cmidrule(lr){2-3}
\multicolumn{4}{@{}l}{\textit{bar-constraint weight, matched generator}}\\
$w_b$ & max & mean & \\
\midrule
1 & 29.50 & 23.11 & 16.59\\
10 & 7.42 & 5.99 & 0.62\\
100$^{\dagger}$ & 3.25 & 2.69 & 1.45\\
1000 & 3.86 & 3.18 & 2.57\\
10000 & 2.14 & 1.69 & 3.49\\
\midrule
\multicolumn{4}{@{}l}{\textit{data against physics, matched generator}}\\
$w_{\mathrm{eom}}$ & max & mean & \\
0.1 & 4.74 & 3.54 & 8.43\\
0.3 & 4.77 & 3.95 & 1.59\\
1$^{\dagger}$ & 3.25 & 2.69 & 1.45\\
3 & 3.13 & 2.59 & 1.53\\
10 & 9.79 & 8.04 & 2.50\\
\midrule
\multicolumn{4}{@{}l}{\textit{the same, with a model-form error present}}\\
$w_{\mathrm{eom}}$ & max & mean & \\
0.1 & 7.71 & 6.01 & 8.90\\
0.3 & 7.73 & 6.38 & 6.45\\
1$^{\dagger}$ & 6.84 & 5.78 & 2.73\\
3 & 6.85 & 5.72 & 2.31\\
10 & 11.87 & 9.86 & 1.30\\
\bottomrule
\end{tabular}

\end{table}

Every identification run through the experiment engine has an
archived per-run configuration file that reproduces it exactly. The
robustness, baseline, and multi-record studies come from dedicated
scripts that fix their protocols in code and record the varied factors
with each result. Table~\ref{tab:params} lists the simulation, sensing, and training parameters used throughout the main text.

\begin{table}[!tbp]
  \centering
  \caption{Parameters used throughout the main text, with the origin of
  each value. Section numbers refer to the main text; material
  values carry their published source. Per-run amplitudes and step
  counts are given in the main-text sections that use them.}
  \label{tab:params}
  \footnotesize
  \setlength{\tabcolsep}{3pt}
  \begin{tabular}{@{}p{0.34\linewidth}p{0.38\linewidth}p{0.22\linewidth}@{}}
\toprule
Parameter & Value & Source\\
\midrule
\multicolumn{3}{@{}l}{\textit{Members, synthetic benchmarks}}\\
\quad Cable radius, modulus       & $1$\,mm, $70$\,MPa ($EA = 220$\,N)      & benchmark definition\\
\quad Cable damping $c_k$         & $1.0$\,N\,s/m, taut cables only          & benchmark definition\\
\quad Bar radius, density         & $5$\,cm, $500$\,kg/m$^3$                 & benchmark definition\\
\quad Bar treatment               & rigid (length constraint)                & \S2.1\\
\quad Hexagonal prism             & $EA = 379$\,N, bar mass $42.3$\,g        & \citet{Prism2025JSV}\\
\midrule
\multicolumn{3}{@{}l}{\textit{Prestress}}\\
\quad Planar frames, bridge       & LP, $25$\,N/m minimum force density      & \S2.3\\
\quad Triplex prism, gravity      & uniquely fixed by equilibrium            & \S2.3\\
\quad Prism series, zero gravity  & self-stress scale $1$--$15$\,N/m         & \S4.6\\
\quad Free-standing polyhedra     & own self-stress mode, sign-checked       & \S4.13\\
\midrule
\multicolumn{3}{@{}l}{\textit{Members, structural assessment}}\\
\quad Bar modulus, yield          & $206$\,GPa, $300$\,MPa (Q345 steel)      & \citet{MaChen2022TsgFEM}\\
\quad Cable effective modulus     & $76$\,GPa (steel rope)                   & \citet{MaChen2022TsgFEM}\\
\quad Cable allowable stress      & $600$\,MPa, about half rope strength     & \citet{EN12385, ASTMA416}\\
\quad Section, buckling factor    & circular $I = \pi R^4/4$, $K = 1$        & \citet{Carpentieri2014Parametric}\\
\quad Rank tolerance              & $10^{-6}$ relative singular value        & \S4.13\\
\midrule
\multicolumn{3}{@{}l}{\textit{Excitation and records}}\\
\quad Classes                     & harmonic, half-sine impulse, broadband   & \S2.4\\
\quad Fixed protocol              & $1.0$\,Hz, one coordinate (two, prism)   & \S2.4\\
\quad Multi-point excitation                  & two coordinates, $1.0$ and $1.7$\,Hz     & \S4.4\\
\quad Amplitude range             & $3$--$60$\,N, spanning both regimes      & \S2.4\\
\quad Record length, output step  & $5$\,s, $0.01$\,s                        & \S2.4\\
\quad Reference integration       & \textsc{rk45}, tolerance $10^{-10}$      & \S2.5\\
\midrule
\multicolumn{3}{@{}l}{\textit{Sensing}}\\
\quad Measured signal             & nodal accelerations                      & \S2.4\\
\quad Instrumented coordinates    & half to two-thirds of free coordinates; one channel in the adverse-sensing study & \S4.9\\
\quad Noise model                 & zero-mean Gaussian, scaled to per-channel RMS & \S2.4\\
\quad Noise level                 & $2$\,\% nominal; $5$, $8$, $10$\,\% in the robustness studies & \S4.9\\
\quad Sparse, intermittent record & $25$\,\% of samples, three dropout windows & \S4.9\\
\midrule
\multicolumn{3}{@{}l}{\textit{Network}}\\
\quad Trunk                       & 4 layers $\times$ 64 units, $\tanh$      & \S3.1\\
\quad Input features              & 8 Fourier frequencies in scaled time     & \S3.1\\
\quad Heads                       & displacement, constraint multipliers     & \S3.1\\
\quad Hard initial condition      & $s(\tau) = (1 - e^{-\tau/0.08})^2$       & \S3.1\\
\quad Displacement scale          & $0.1$\,m                                 & \S3.1\\
\quad Smoothing sharpness $\beta$    & 800                                      & \S3.1.3\\
\midrule
\multicolumn{3}{@{}l}{\textit{Optimization}}\\
\quad Optimizer                       & Adam, cosine-annealed                    & \S3.1\\
\quad Learning rates              & $2\times10^{-3}$ network, $10^{-2}$ parameters & \S3.1\\
\quad Loss                        & $\mathcal{L}_{\mathrm{data}} + \mathcal{L}_{\mathrm{EOM}} + 100\,\mathcal{L}_{\mathrm{bar}}$ & \S3.1\\
\quad Collocation points          & 256 fresh points per step                & \S3.1\\
\quad Training length             & 25{,}000 steps; 12{,}000 in the grid studies   & \S3.1\\
\quad Data split                  & 70 train / 30 held out                   & \S3.1\\
\quad Parameter initialization    & $\sigma_k = 10$\,N/m, $c = 0.3$\,N\,s/m  & \S3.1\\
\quad Seeds                       & 0--8, network and noise realization      & \S4.8\\
\bottomrule
\end{tabular}

\end{table}

\section{Assessment across the classic tensegrity forms}
\label{sm:families}
The stability classification of the main text is based on the same rank count that governs identifiability. This section applies it beyond the
identification benchmarks, to forms that are assessed with their exact force densities and are never identified.

\begin{figure}[!htbp]
\centering
\includegraphics[width=0.97\linewidth]{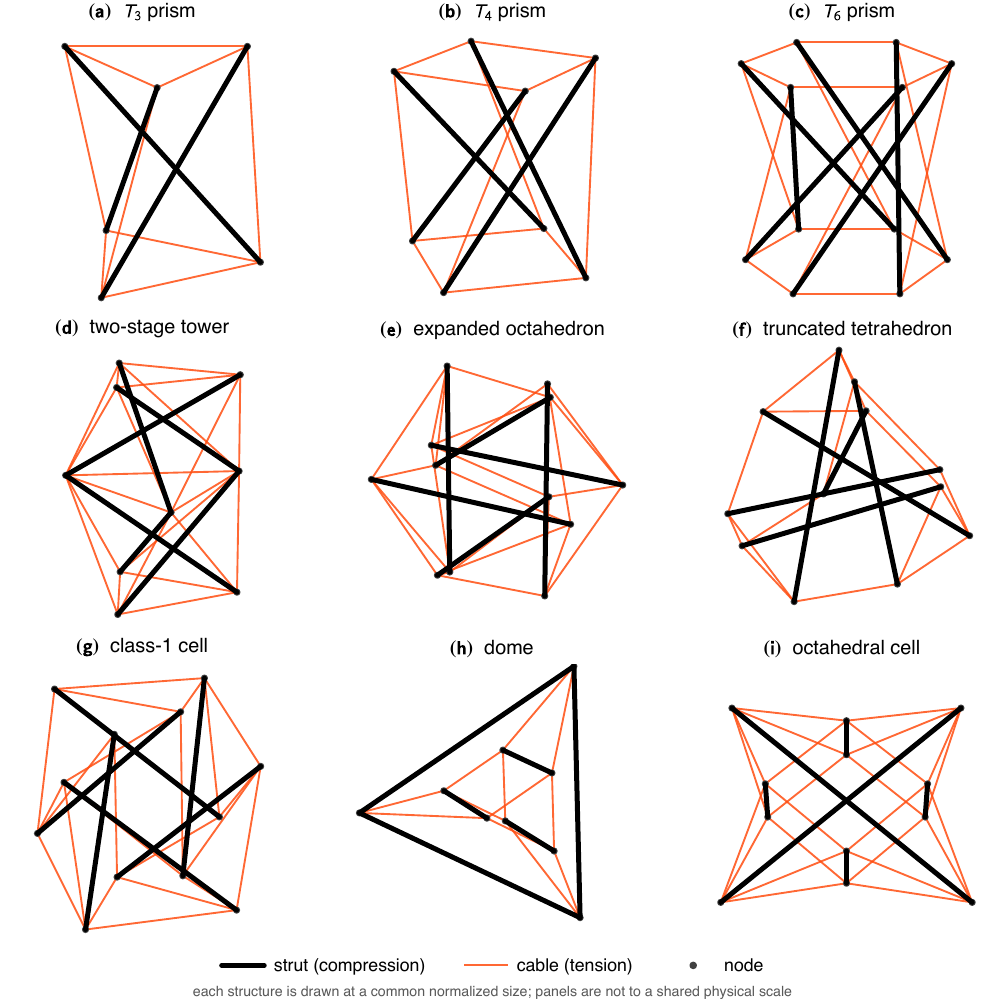}
\caption{The classic tensegrity forms assessed alongside the
identification benchmarks, struts heavy and cables light. None of
these is identified; they extend the structural analysis only.}
\label{fig:zoo}
\end{figure}

\begin{figure}[!htbp]
\centering
\includegraphics[width=0.82\linewidth]{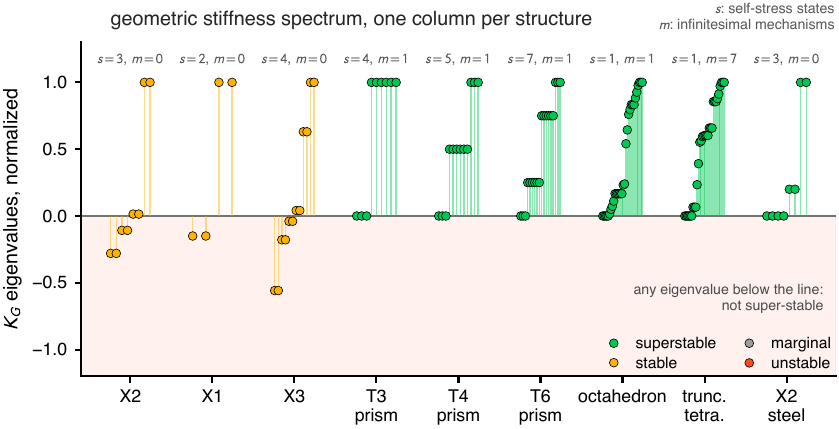}
\caption{Normalized geometric-stiffness spectrum of each structure, green
for super-stable spectra (all eigenvalues non-negative) and orange for
merely stable ones.}
\label{fig:spectra}
\end{figure}

The rank of the equilibrium matrix $\bm A$ fixes the
number of self-stress states $s=n_{\mathrm{mem}}-\operatorname{rank}\bm A$
and infinitesimal mechanisms $m=n_{\mathrm{free}}-\operatorname{rank}\bm A$,
related by Maxwell's rule $m-s=n_{\mathrm{free}}-n_{\mathrm{mem}}$
\citep{Calladine1978, PellegrinoCalladine1986}. Table~\ref{tab:structural}
collects these counts and the class for the planar identification
benchmarks. Ranks use a relative singular-value tolerance of
$10^{-6}$, which accommodates the six-figure coordinates of the
published polyhedral geometries. The octahedral forms satisfy their
self-stress to machine precision and the rest to about $10^{-7}$. The table also covers three classic forms assessed with the same formulation (Figure~\ref{fig:zoo}): the tensegrity prisms $T_3$, $T_4$, $T_6$, the expanded octahedron (the six-strut tensegrity
icosahedron) \citep{Connelly1998,
FernandezRuiz2022}, and the truncated-tetrahedron tensegrity
\citep{ZhangOhsakiJMPS2012}. These forms are free-standing and are therefore minimally grounded here. Their prestress is set from their own self-stress mode. Their counts
come from the equilibrium matrix on all coordinates, since no reaction
absorbs equilibrium at a grounding node. The supported frames and
prisms are counted on their free coordinates. The self-stress count is
precisely the dimension of the identifiable prestress space, and the
mechanisms are what that prestress must brace. The rank deficiency
that admits a self-stress thus also makes the prestress identifiable
and stability-critical. The planar frames carry no mechanism and are merely stable under gravity, because their stress matrices are indefinite (Figure~\ref{fig:spectra}). Every free-standing polyhedral
tensegrity in the table, by contrast, is super-stable at its own
self-stress. Its margin falls as the number of struts grows ($28.2$ for $T_3$ to
$8.4$ for $T_6$) and as the mechanism count rises. The truncated
tetrahedron is the extreme case: seven mechanisms braced by a single
self-stress state. Gravity load moves the frames off the
super-stability boundary.

The column of determined cables gives the identifiability result of the main text. Of the $183$ cables of the 12 structures, the free coordinates determine $164$ at first order. Of the 19 undetermined cables, 17 are individually invisible, each spanning two restrained nodes.
The remaining two are combinations of members: the X2 and X3
frames each carry a self-stress direction that produces no
free-coordinate force at the reference configuration; no single cable
of either is invisible, but one combination is undetermined at first order. The pattern is exact on the tensegrity prisms. There, the $n$ cables of the pinned base polygon join restrained nodes only and are undetermined, while every other member is determined; 3, 4, and 6 of the 9, 12, and 18 cables of $T_3$, $T_4$, and $T_6$ are therefore undetermined. In the free-standing polyhedra every cable is determined, since minimal
grounding leaves no pair of fully restrained nodes for a cable to
span.

\begin{table}[!tbp]
\centering
  \footnotesize
\caption{Free coordinates, members, rank of the equilibrium matrix,
self-stress states $s$, mechanisms $m$, cables determined at first order,
stability class, and margin $\lambda_{\min}$ of the reduced tangent
stiffness; rank tolerances are given in the text.}
\label{tab:structural}
\begin{tabular}{lcccccccc}
\toprule
Structure & Free & Mem. & Rank $\bm A$ & $s$ & $m$ & Resolvable & Class & $\lambda_{\min}$\\
\midrule
X1 frame & 4  & 6  & 4  & 2 & 0 & 3/4 & stable & 148.6\\
X2 frame & 8  & 11 & 8  & 3 & 0 & 6/7 & stable & 211.6\\
X3 frame & 12 & 16 & 12 & 4 & 0 & 9/10 & stable & 45.4\\
$T_3$ prism & 9  & 12 & 8  & 4 & 1 & 6/9 & super-stable & 28.2\\
$T_4$ prism & 12 & 16 & 11 & 5 & 1 & 8/12 & super-stable & 17.7\\
$T_6$ prism & 18 & 24 & 17 & 7 & 1 & 12/18 & super-stable & 8.4\\
expanded octahedron & 30 & 30 & 29 & 1 & 1 & 24/24 & super-stable & 2.0\\
octahedral cell & 30 & 30 & 28 & 2 & 2 & 24/24 & super-stable & 2.5\\
dome & 21 & 18 & 17 & 1 & 4 & 12/12 & super-stable & 17.4\\
truncated tetrahedron & 30 & 24 & 23 & 1 & 7 & 18/18 & super-stable & 1.3\\
class-1 cell & 30 & 30 & 29 & 1 & 1 & 24/24 & super-stable & 24.0\\
two-stage tower & 18 & 27 & 18 & 9 & 0 & 18/21 & super-stable & 22.3\\
\bottomrule
\end{tabular}
\end{table}

\section{Biomimetic-leg reconstruction details}
\label{sm:leg}

Three candidate origins of the model-to-measurement frequency gap of
the leg of \citet{Wang2026Leg} are excluded quantitatively. First, the gap is not measurement scatter. The springless case, repeated three times, has a standard deviation of \SI{0.0006}{\hertz} on $f_1$, and its \SI{0.0155}{\hertz} distance from the computed value is 26 standard errors. Second, the gap is not an artifact of the chirp, which would bias a peak in the direction of the frequency change. Driving the leg's own mass and stiffness matrices with their protocol, a linear chirp from $0.01$ to \SI{4}{\hertz} at \SI{0.0327}{\hertz\per\second}, and picking peaks from the response spectrum as they do displaces the peak by at most $0.8$\,\% for modal damping between $1$ and $5$\,\%. Third, the gap is not an error in the reported masses or dimensions. No entry of their Tables~D.5 or~D.6 has enough influence: every log-sensitivity of $f_1$ is below $0.21$ in magnitude. The most influential entry would need a $10$\,\% error to move $f_1$ by $2$\,\%, and it would increase the error in $f_2$.

The missing restoring stiffness has two parts: one that remains with no springs installed, measured by Case~E, and one that grows with the installed hip springs. When the cases are ordered by the stiffness installed at the hip, the $f_1$ gap is $2.3$, $3.4$, $3.7$, $5.1$, and $5.1$\,\% for Cases~E, A, D, B, and C. Fitting a residual
rotational stiffness at the two joints to Case~E alone returns
\SI{0.107}{\newton\metre\per\radian} at the hip and
\SI{0.011}{\newton\metre\per\radian} at the knee, about $10$\,\% of the gravitational restoring stiffness of the first mode. If a single multiplicative factor on the hip rate is also allowed, all eight frequencies of Cases~A to~D are reproduced within $1.2$\,\% with two constants, and the factor is $1.27$. This result is recorded as a diagnosis and not adopted as a correction, because it is fitted to the frequencies that it then reproduces.

The hip statistics behind the exclusion are as follows. When both scales are identified from the published frequency pairs, contrary to the screening, the hip rate exceeds the rig rate by $49$ to $68$\,\% across Cases~A to~D, with a root-mean-square of $56.5$\,\%. Multiplying each case's frequency gap by its amplification $2/s$ predicts $37$ to $47$\,\%, and the prediction and the outcome correlate at $0.89$ across the cases. The overshoot is therefore the amplified model-form gap and not an artifact of the optimizer. Identified record by record, the hip scale is $+29 \pm 38$, $+62 \pm 24$, $+185 \pm 141$ and
$+70 \pm 8$\,\% for Cases~A to~D, and in the time-domain envelope
fits it pools to $0.931 \pm 0.336$ over the 15 records with springs. This value is centered on the rig value, but its coefficient of variation is $36$\,\%, against $18$\,\% for the knee, which is too imprecise to report as a result. The screening in the main text summarizes these statistics, and they are reported here in full.

The simulated test of the time-domain envelope identification, a record with known parameters and $2$\,\% noise, returns the hip
scale to $-4.4$\,\%, the knee scale to $+0.1$\,\%, both dampers to
within $0.5$\,\%, and the drive amplitude to $+0.6$\,\%, at an
envelope error of $0.0026$.

The measured records also give two quantities that the source publication does not report. The damping ratio of the second mode is $0.64 \pm 0.24$\,\% of critical across all 18 records, measured on the free ring-down after the chirp has passed the resonance; that of the first mode is near $0.4$\,\%, with much wider scatter. Every record also requires a drive amplitude larger than the commanded stroke, by a pooled factor of $1.183 \pm 0.137$, and an independent comparison of swing amplitudes gives $1.21$. Because two independent routes require the same increase of about $20$\,\%, its likely source is the test stage.

\section{Structural assessment of the identified model}
\label{sm:assessment}
This section contains the structural-assessment formulation of the main text and its demonstration with exact force densities. The propagation
of identification error to the assessment remains in the main text,
since it is a property of the identification.

The recovered force densities describe the prestress state of the structure. This prestress determines two safety-critical properties: the geometric stability of the self-stressed shape and the strength demand on its members. Both follow from the identified prestress without further measurement. The time-domain
identification keeps the rigid-bar model; the assessment adds the
members' finite axial rigidity and their strength limits. The two agree
on the mechanism-dominated response that sets the margin, as verified below.

Let $\bm A$ be the equilibrium matrix mapping member force densities to
nodal forces, $\bm D=\operatorname{diag}(E_mA_m/\ell_m)$ the member axial
stiffnesses, and $\bm S=\sum_m \sigma_m\,\bm C_m$ the stress matrix
assembled from the force densities $\sigma_m$. The tangent stiffness is the
Hessian of the total potential energy,
\begin{equation}
  \bm K_t \;=\; \underbrace{\bm A\,\bm D\,\bm A^{\top}}_{\bm K_M}
  \;+\; \underbrace{\vphantom{\bm A^{\top}}\bm S}_{\bm K_G},
  \label{eq:tangent}
\end{equation}
the sum of a material stiffness $\bm K_M$ and the geometric prestress
stiffness $\bm K_G$; this is the standard tangent stiffness of the
force-density method \citep{ZhangOhsaki2015}. For the rigid bars of this study, the geometric stiffness is the
constraint-Hessian stress matrix rather than the elastic-truss
softening, and the transverse softening that governs a stocky member
enters the buckling check below instead. Restricted to the free
coordinates, the configuration is stable when the reduced $\bm K_t$ is
positive definite, and its smallest eigenvalue $\lambda_{\min}$
measures the margin.

Where the structure carries infinitesimal mechanisms,
displacements $\bm u_m$ in the null space of $\bm A^{\top}$, stability depends on the prestress alone. The configuration is then
prestress-stable when $\bm u_m^{\top}\bm S\,\bm u_m>0$ for every
mechanism, equivalently when the reduced stress matrix
$\bm U_m^{\top}\bm S\,\bm U_m$ is positive definite
\citep{PellegrinoCalladine1986, Calladine1978}. A configuration is super-stable when the stress matrix $\bm S$ is
itself positive semidefinite on the free coordinates
\citep{ZhangOhsaki2015}. It is then stable for any material stiffness and any positive scaling of the prestress. Without super-stability, stability also depends on the material stiffness. A third, weaker sufficient condition tests only the
directions the material stiffness cannot resist. With $\bm U_0$ a basis
for the null space of $\bm K_M$, the configuration is prestress-stable
when
\begin{equation}
  \bm U_0^{\top}\bm K_G\,\bm U_0 \succ 0,
  \qquad \bm K_M\bm U_0 = \bm 0,
  \label{eq:ohsaki}
\end{equation}
a test projected onto the mechanisms, which is not needed when $\bm K_M$ is already definite \citep{ZhangOhsaki2015}. The three classes, super-stable,
stable, and unstable, follow from the signs of the two margins
$\lambda_{\min}(\bm K_t)$ and $\lambda_{\min}(\bm K_G)$,
\begin{equation}
  \text{unstable if } \lambda_{\min}(\bm K_t) < 0,\quad
  \text{super-stable if } \lambda_{\min}(\bm K_G) \geq 0,\quad
  \text{stable otherwise,}
  \label{eq:classes}
\end{equation}
where each sign is taken against a tolerance scaled by $\lVert\bm K_G\rVert$, because this norm sets the size of the mechanism-mode eigenvalues on which the classification depends.

Member strength is checked against the same force densities. A cable
carries $T_m=\sigma_m\ell_m$ and a strut the balancing compression
$C_m$. Each is measured against its capacity by a utilization,
\begin{equation}
  U_c = \frac{T_m}{A_m\,\sigma_{\mathrm{all}}},
  \qquad
  U_b = \frac{C_m}{P_{\mathrm{cr}}},
  \qquad
  P_{\mathrm{cr}} = \frac{\pi^2 E_m I_m}{(K\ell_m)^2},
  \label{eq:utilization}
\end{equation}
with a circular section $I_m=\pi R_m^4/4$ and the pinned-pinned factor
$K=1$ appropriate to cable-terminated nodes
\citep{Carpentieri2014Parametric}. A member reaches
its limit at unit utilization. Geometric instability, indicated by $\lambda_{\min}$, and material failure, indicated by $U_c$ and $U_b$, are thus separated and reported together.

Member properties follow the tensegrity material library of
\citet{MaChen2022TsgFEM}: Q345 structural-steel bars with $E_b=206$\,GPa, and
wire-rope cables with an effective $E_c=76$\,GPa. The bar yield is taken as
$300$\,MPa, a design value about $13$\,\% below the $345$\,MPa nominal
that names the grade. The cable modulus is that of multi-strand wire rope.
It is well below solid steel because the helical lay of the strands adds a
geometric compliance in series with the wire material. It should not
be read as the modulus of a spiral strand, which retains a larger fraction
of the solid-steel value. The cable is
checked against an allowable $\sigma_{\mathrm{all}}=600$\,MPa, about half
the $1200$\,MPa breaking strength assumed by the source library. That allowable stress is well inside the range of standard products: spiral ropes
for structural applications are specified in EN 12385-10
\citep{EN12385}, and seven-wire prestressing strand under ASTM A416
carries a specified minimum tensile strength of $1860$\,MPa in grade 270
\citep{ASTMA416}. The utilization therefore already carries a factor of at
least two on the material.

The model also tracks the structure's energy through each excitation:
the kinetic energy $T$, the gravitational potential $V_g$, and the cable
strain energy $V_s$ are evaluated at every step. Their sum obeys the exact
balance $\tfrac{\mathrm d}{\mathrm dt}(T+V_g+V_s)=(f_{\mathrm{ext}}+
f_{\mathrm{damp}})\cdot\dot{\bm q}$. The balance also serves as an integration check and shows the strain energy stored and released at the taut--slack transitions.

The assessment of this section turns a set of
force densities into a stability margin and member utilizations. It is used in two different ways. The formulation, and the behavior of the classes across structures and configurations, are demonstrated with the exact force densities; they therefore show properties of the structure, not of one inversion. Only
the propagation table of the main text substitutes the identified force densities
for the exact ones, and it shows how much the identification error changes the assessment.

The lowest natural frequency from the tangent stiffness of
Eq.~\eqref{eq:tangent} agrees with the constrained modal analysis of
the rigid-bar dynamics to within $0.2$\,\%: $1.410$ against
$1.413$\,Hz on the X2 frame. The elastic-member assessment and the
rigid-bar model therefore share the mechanism-dominated response. The
assessment also admits a rigid or a flexible bar model. On the steel members, the two give stability margins within $0.5$\,\% of each other. The rigid-bar assumption behind the identification therefore carries into the assessment with negligible error. The strength utilizations depend only
on the member forces and are identical between the two.

The triplex prism has one infinitesimal mechanism, and its reduced stress matrix is positive, with a margin of $28.2$. The prestress alone therefore braces the mechanism, which is the defining prestress-stability property of a tensegrity \citep{ZhangOhsaki2015}. The mechanism-projected condition
agrees. The geometric stiffness restricted to the null space of the
material stiffness has smallest eigenvalue $28.2$ on the prism and
stays positive on every free-standing structure assessed. On the planar
frames the material stiffness is already definite and the test is not needed.

\begin{figure}[!tbp]
\centering
\includegraphics[width=0.97\linewidth]{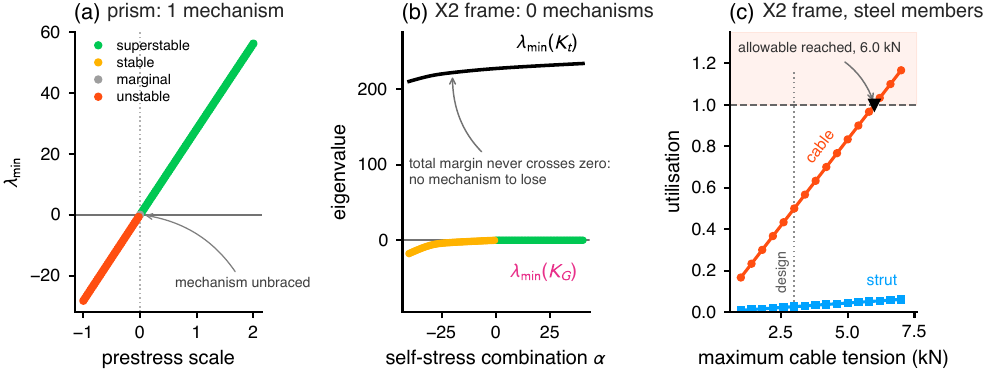}\\[2pt]
\includegraphics[width=0.90\linewidth]{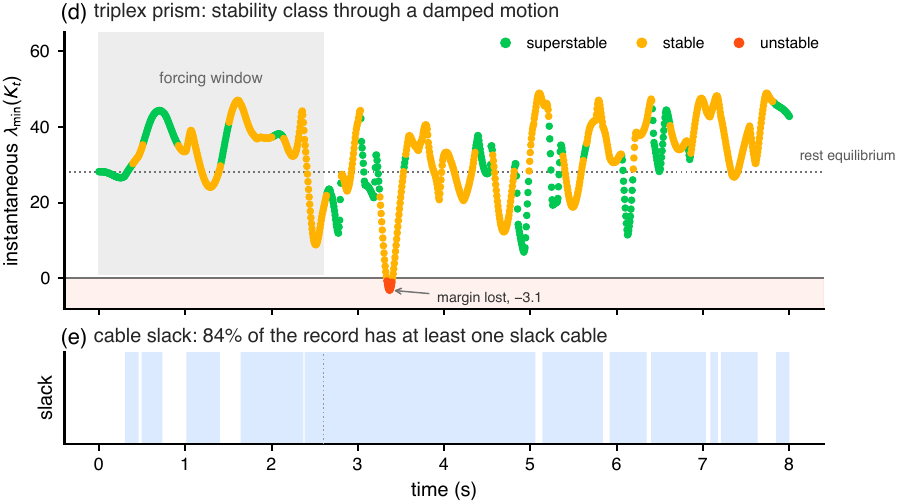}
\caption{Stability from the identified parameters: (a)~the prism over a range of prestress; (b)~the X2 frame over a range of the self-stress combination; (c)~member utilizations of the steel X2 frame
against prestress; (d)~instantaneous stability class of the prism
through a forced, damped motion (green super-stable, orange merely
stable, red momentarily unstable), with (e)~the cable-slack
intervals.}
\label{fig:stabclasses}
\end{figure}

Classifying the reduced stiffness and stress matrix places each configuration in one of the three classes (Figure~\ref{fig:stabclasses}). The benchmarks fall into two of them. The triplex prism is the only benchmark with an infinitesimal mechanism, and its stability therefore depends on the prestress alone; it is also the only one that becomes unstable when the prestress is reversed.
The planar frames carry no mechanism, and their material stiffness holds them whatever the prestress does; only the definiteness of the geometric stiffness changes, and it marks the super-stable boundary.

The class depends on the configuration as well as the topology. The
triplex prism at its identified self-stress is super-stable in the
sense of this section, and so is the X2 frame at
the steel design prestress used below. Under gravity load, the frames instead carry an indefinite stress matrix and are merely stable.
Scaling and reversing the prism prestress takes it through these classes: the mechanism loses its bracing as the prestress approaches zero and becomes unstable once the prestress reverses. On the X2 frame, a non-optimal self-stress makes the stress matrix indefinite, leaving a merely stable
configuration held by the material stiffness. The identified
load-balancing self-stress lies on the super-stability boundary.

The classification uses the same rank count as the identifiability analysis. Section~\ref{sm:families} applies it to the classic tensegrity forms and gives the counts, classes, margins, and determined cables of all 12 structures.

The class is not only a property of the resting shape. The triplex prism is used for this demonstration for the same reason it is used for the prestress series: a frame without mechanisms cannot show it. Repeating
the experiment on the X2 frame leaves its margin above $43$ throughout, because the material stiffness holds the shape whatever the taut--slack state of the cables. Its class therefore does not change. Under excitation, the instantaneous configuration of the prism changes class as cables change between taut and slack (Figure~\ref{fig:stabclasses}, bottom). A slack
cable drops out of the stress matrix, and the frozen shape passes from
super-stable through merely stable to momentarily unstable once enough
bracing is lost. Damping restores the super-stable equilibrium as the
motion decays. The stability class thus tracks the same taut--slack transitions that the identification uses.

The X2 frame is rebuilt with the steel members of the material library
at its design prestress, a maximum cable tension of $3$\,kN. There the
reduced tangent stiffness is positive definite and the self-stressed
shape is super-stable. The strongest cable carries half its allowable
($U_c=0.50$), and the most compressed strut a small fraction of its
buckling load ($U_b=0.03$). Raising the prestress traces the
utilization curves of Figure~\ref{fig:stabclasses}(c). The cable
reaches its allowable ($U_c=1$) at a maximum tension of $6$\,kN while
the stability margin stays large and the strut stays far from
buckling. On this frame, material strength, not geometric instability, governs the safe prestress.

\begin{figure}[!tbp]
\centering
\includegraphics[width=0.84\linewidth]{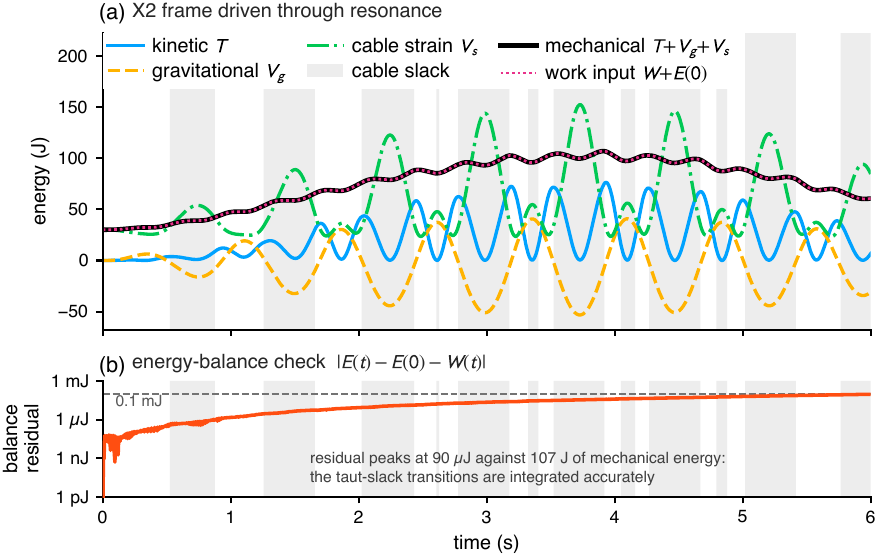}
\caption{Energy components of the X2 frame under resonant excitation
(top) and the energy-balance residual (bottom); shaded bands mark slack
intervals.}
\label{fig:energy}
\end{figure}

Figure~\ref{fig:energy} traces the energy balance of
this section on the X2 frame under resonant excitation that slackens the cables. The balance is a property of the constrained integration, and it is therefore reported on the reference planar benchmark only. The check is evaluated on every record behind the results above. The strain energy rises and falls as
cables load and unload, and the mechanical energy tracks the work
input. The balance residual stays below $10^{-4}$\,J through the taut--slack transitions.

The flexible-bar model also lets the identification carry the bar
axial stiffness as an unknown. With the residual extended to elastic
bars and $EA_{\mathrm{bar}}$ added to the trainable parameters, the
extended residual is verified against the flexible-bar simulator
(the verification table of the main text). A sensitivity analysis nonetheless shows that records of this bandwidth carry little information about the bar stiffness. Bar axial deformation is a high-frequency mode, decoupled from the mechanism motion whose frequencies the prestress sets. The low-frequency, mechanism-dominated record therefore carries little information about $EA_{\mathrm{bar}}$.
Band-limiting the response below the bar axial modes removes about
$70$\,\% of its sensitivity at the soft bar stiffness tested. The
remainder, the bar's quasi-static compliance, scales as
$1/EA_{\mathrm{bar}}$. The stiff limit itself is too stiff to
integrate and was not computed. Identifying the bar stiffness would
require exciting the bar axial modes directly. The bar force, by
contrast, is fixed by the self-stress and is identified alongside the
cable force densities. It is the same separation of an identifiable
force from an unidentifiable stiffness seen in the cable
stiffness--prestress trade-off.

\section{Robustness studies}
\label{sm:robustness}
This section gives the three robustness studies on the X1 frame that Section~4.9 of the main text summarizes.

\subsection{Noise, sensor count, and joint stiffness identification}
\label{sm:robust-noise}

Accuracy decreases moderately with noise: a fivefold increase in noise increases the force-density error by a factor of $1.8$. Table~\ref{tab:robust} varies sensor noise
and the sensor count, one factor at a time, on the X1 taut record.
Every entry is a mean over seeds of the same $12{,}000$-step protocol, and the spread is the scatter over seeds. The damping error is the more sensitive of the two and by far the more scattered; at the higher noise levels, its standard deviation approaches its mean. Removing one of the two measured channels leaves a single accelerometer coordinate for a two-degree-of-freedom structure, and the inversion still identifies the force densities and the damping to within a few percent. Halving the instrumentation therefore increases the error less than raising the noise fivefold.

\begin{table}[!tbp]
  \centering
  \footnotesize
  \caption{Robustness of the X1 inversion on the taut record. Each
  entry is the mean over seeds of the matched $12{,}000$-step
  protocol, with one standard deviation.}
  \label{tab:robust}
  \begin{tabular}{lccc}
\toprule
Condition & Seeds & max $\hat\sigma$ err (\%) & $\hat c$ err (\%) \\
\midrule
$2$\,\% noise, 2 sensors & 9 & $3.60\pm0.28$ & $1.58\pm0.93$ \\
$5$\,\% noise, 2 sensors & 4 & $4.81\pm0.76$ & $4.10\pm2.10$ \\
$10$\,\% noise, 2 sensors & 4 & $6.60\pm0.79$ & $4.28\pm3.53$ \\
$2$\,\% noise, 1 sensor & 4 & $4.26\pm0.67$ & $3.57\pm2.95$ \\
\bottomrule
\end{tabular}

\end{table}

Identifying the axial stiffness as well shows the trade-off that the Fisher analysis predicts. With $EA$, $\boldsymbol\sigma$, and $c$ identified jointly
on the taut record, the stiffness is recovered to $7.4$\,\% but the force-density errors grow to $35$\,\%: the two parameter groups exchange
error along their correlated direction. On the slack-transition record, the joint identification fails with an $83$\,\% stiffness error, and the held-out ratio again flags the failure. Reliable stiffness identification
therefore needs either records rich enough to separate the groups or an
independent stiffness measurement. The joint formulation makes this ambiguity visible.

\subsection{Model-form error}
\label{sm:robust-model}

The simulated records of the main text are generated and inverted with the same forward model, which is an inverse crime (Section~4.2 of the main text). Four generators test whether the accuracies hold under model error, with the identification left unchanged throughout. The first perturbs the mass matrix by up to $5$\,\% per coordinate, which represents unmodeled joint masses and fittings but leaves the model form intact. The second and third change the model form. The second lumps half of each cable's mass at its end nodes, which makes the massless-chord cable law of the residual wrong at every parameter value. The third replaces the rigid bars with elastic ones, which removes the bar-length constraints, and it is integrated by an implicit multistep solver instead of the explicit scheme used for the matched records. Neither the equations nor their discretization are
shared. The fourth tightens the generator
tolerance from $10^{-10}$ to $10^{-13}$, which changes the discretization only.

The mass perturbation leaves the X1 taut recovery essentially unchanged, at $3.0$\,\% against $3.5$\,\% for the matched case. On the prism, it raises the force-density error from $0.6$ to $3.5$\,\% and the damping error from $3.2$ to $4.2$\,\%.
Table~\ref{tab:mismatch} reports the rest. The lumped cable mass is only $1.2$\,\% of the structural mass, but it changes the acceleration record by $8.6$\,\%, four times the sensor noise, and doubles the taut force-density error to $7.0$\,\%. Elastic bars increase the error by a factor of $1.5$. Lumping the cable mass at the nodes is the more severe of the two standard choices. A consistent distributed-chord mass matrix moves a third of that mass into coupling between the end nodes; it changes the record by $5.7$\,\% instead of $8.6$\,\% and gives an error of $5.9$\,\% instead of $7.0$\,\%. Both increases are small, and the errors stay below $10$\,\%. On the slack-transition record, every generator stays within the seed scatter of the matched case, and the model-form errors tested therefore do not limit that inversion. The tolerance row is unchanged to two significant figures, which shows that the integration accuracy does not affect the results. Within these tests, the accuracies reported in the main text therefore also hold for generators that the residual cannot represent.

\begin{table}[!tbp]
\centering
  \footnotesize
\caption{Model-form mismatch on the X1 frame, as percentage errors. Each
record is produced by a generator the residual cannot represent at any
parameter value, and the inversion is left unchanged. The $\sigma$ column is
the maximum over the identifiable force densities, mean and standard
deviation over three seeds; $c$ is the damping coefficient.}
\label{tab:mismatch}
\begin{tabular}{lcccc}
\toprule
& \multicolumn{2}{c}{Taut record} & \multicolumn{2}{c}{Slack-transition record}\\
\cmidrule(lr){2-3}\cmidrule(lr){4-5}
Generator & $\sigma$ & $c$ & $\sigma$ & $c$\\
\midrule
matched & $3.5\pm0.2$ & $1.1$ & $19.7\pm2.1$ & $4.6$\\
cable mass & $7.0\pm0.2$ & $2.1$ & $21.0\pm2.8$ & $5.7$\\
flexible bars & $5.3\pm0.4$ & $1.9$ & $17.0\pm1.9$ & $6.0$\\
tolerance $10^{-13}$ & $3.5\pm0.2$ & $1.1$ & $19.3\pm1.6$ & $5.6$\\
\bottomrule
\end{tabular}

\end{table}

\subsection{Adverse sensing}
\label{sm:robust-sensing}
The identifications of the main text use favorable conditions: two acceleration channels, $2$\,\% noise, harmonic forcing, and a dense record. Field monitoring rarely provides all four. The X1 record was therefore regenerated and inverted under four adverse conditions drawn from practice, first one at a time and then all together. The first is a single acceleration channel, the
sparse sensor placement forced when a full structure cannot be
instrumented \citep{Kammer1991, Papadimitriou2004}. The second is
$8$\,\% measurement noise, the upper end of what vibration studies
report \citep{Moaveni2009Uncertainty}. The third keeps
only a quarter of the time samples, in intermittent blocks with three
sensor-off dropout windows; aging and telemetry faults make such data loss common \citep{Bao2013Compressive,
Zhang2024Missing}. The fourth replaces the harmonic forcing with an impulse (impact) excitation \citep{Ewins2000Modal}. Physics-informed methods have been used for identification from such sparse and noisy records \citep{Raissi2019, Chen2021Scarce}.

Turning each factor on alone separates their contributions
(Table~\ref{tab:harsh}). The baseline case recovers the identifiable force densities to $2.8$\,\%. Quadrupling the noise and halving the sensor count increase the error only by factors of $1.6$ and $1.7$. Discarding three quarters of the samples in intermittent blocks increases it by a factor of $4.3$. Replacing the harmonic forcing with an impulse increases it by a factor of $8.8$, the largest of the four. An impulse supplies energy only once, after which the structure vibrates freely. The cables therefore spend most of the record in free decay instead of being driven through their working range.

With all four together, the error is no worse than with the impulse alone, $24.9$\,\% against $25.0$\,\%. The fit remains close throughout, which indicates that the lost accuracy comes from reduced information and not from a failure to converge. The damping is the exception. Across three seeds of the combined condition, its error is $101$\,\% with a scatter of $\pm73$\,\%, consistent with monitoring experience that damping estimates degrade first
\citep{Moaveni2009Uncertainty}. On a single adverse record, the excitation is the limiting factor: the attainable accuracy depends more on what the excitation makes observable than on how well the response is measured.

\begin{table}[!tbp]
\centering
  \footnotesize
\caption{Adverse-sensing ablation on the X1 frame: each condition
turns on one adverse factor against the baseline case, and the last turns on all
four. Errors are the maximum over the identifiable force densities.}
\label{tab:harsh}
\begin{tabular}{lccccc}
\toprule
Condition & Sensors & Noise & Samples & Forcing & $\sigma$ error\\
\midrule
baseline          & 2 & $2$\,\% & 500 & harmonic & $2.8$\,\%\\
high noise        & 2 & $8$\,\% & 500 & harmonic & $4.7$\,\%\\
single sensor     & 1 & $2$\,\% & 500 & harmonic & $4.9$\,\%\\
sparse, dropout   & 2 & $2$\,\% & 106 & harmonic & $12.2$\,\%\\
impulse           & 2 & $2$\,\% & 500 & impulse  & $25.0$\,\%\\
all four          & 1 & $8$\,\% & 120 & impulse  & $24.9\pm4.2$\,\%\\
\bottomrule
\end{tabular}
\end{table}

\end{document}